\pdfoutput=1                        %
\documentclass[twoside,11pt]{article}
\usepackage[preprint]{jmlr2e}       %
\usepackage[utf8]{inputenc}
\usepackage[T1]{fontenc}
\usepackage{lmodern}                %
\usepackage{microtype}
\usepackage{subcaption}
\usepackage{float}
\usepackage{booktabs}
\usepackage{array}
\usepackage{longtable}
\usepackage{algorithm}
\usepackage{algorithmic}
\usepackage{xcolor}
\usepackage{xurl}                   %
\hypersetup{hidelinks}              %
\hypersetup{pdftitle={Whose Effect Is It? How Objectives Inflate Learned Subspace Interventions},
  pdfauthor={Toygar Tanyel and Serkan Ayvaz},
  pdfkeywords={interpretability, causal abstraction, interchange interventions, distributed alignment search, activation patching}}
\renewcommand{\_}{\textunderscore\allowbreak}   %
\usepackage{amsmath}
\usepackage{mathtools}
\let\proof\relax
\ifdefined\newcounteralias\RemoveFromHook{package/amsthm/after}[firstaid/aliascounter]\fi
\usepackage{amsthm}
\usepackage{thmtools}
\usepackage[capitalize,noabbrev]{cleveref}
\usepackage{floatpages}
\usepackage{placeins}
\usepackage{etoc}                   %
\usepackage{etoolbox}
\makeatletter               %
\patchcmd{\l@section}{\setlength\@tempdima{1.5em}}{\setlength\@tempdima{1.9em}}{}{\errmessage{contents: the number width patch failed}}
\makeatother
\usepackage{needspace}
\renewcommand{\floatpagefraction}{0.85}   %

\floatstyle{plain}\restylefloat{table}
\makeatletter
\DeclareCaptionTextFormat{upright}{{\let\textbf\@firstofone #1}}
\@for\@env:=theorem,proposition,lemma,corollary,definition,remark,example\do{%
  \@for\@cs:=,end,c@,the,p@,cl@,theH\do{\expandafter\let\csname\@cs\@env\endcsname\relax}}
\makeatother

\theoremstyle{plain}
\newtheorem{theorem}{Theorem}
\newtheorem{proposition}[theorem]{Proposition}

\theoremstyle{definition}
\newtheorem{definition}[theorem]{Definition}

\newtheorem{example}{Example}
\theoremstyle{remark}
\newtheorem{remark}[theorem]{Remark}

\newcommand{\E}{\mathbb{E}}

\newcommand{\roTasks}{1168}

\newcommand{\roOppositeNum}{880}
\newcommand{\roOppositeDen}{882}
\newcommand{\roSlopeHigh}{$-0.49$}
\newcommand{\roSlopeLow}{$-0.60$}

\newcommand{\roTraceDevMid}{0.05--3.8}
\newcommand{\roRoom}{1124}
\newcommand{\roCeMissNpi}{27}
\newcommand{\roClsSeedsOne}{1.47, 1.38 and 0.15}
\newcommand{\roClsTenT}{2.10}
\newcommand{\npiCases}{112}
\newcommand{\npiMissTop}{2.4}
\newcommand{\npiHitTop}{3.5}
\newcommand{\npiRoomP}{0.001}
\newcommand{\npiPbP}{0.35}
\newcommand{\roCeAbove}{1096}
\newcommand{\roCeMiss}{28}
\newcommand{\roEffAbove}{1124}
\newcommand{\roDistinctTasks}{90}
\newcommand{\roCeAboveAll}{1103}
\newcommand{\roCeAboveTen}{1102}
\newcommand{\roDupRavel}{121}
\newcommand{\roDupSst}{14}
\newcommand{\roRoomConstr}{994}
\newcommand{\roCeAboveConstr}{966}
\newcommand{\roCeMissConstr}{28}
\newcommand{\roExcl}{19}
\newcommand{\roExclFew}{14}
\newcommand{\roExclFlat}{5}
\newcommand{\roExclFewClusters}{10--18}
\newcommand{\roExclFewCe}{1.2--3.4}
\newcommand{\gmIoiRawT}{$-15.2$ and $-14.6$}
\newcommand{\gmIoiSotT}{10.7--12.9}
\newcommand{\gmRavelPop}{24}
\newcommand{\gmRavelRaw}{8}
\newcommand{\gmRavelSot}{23}
\newcommand{\roCeAboveBH}{1093}
\newcommand{\roEffAboveBH}{1121}
\newcommand{\gmCeilingTasksBH}{25}
\newcommand{\roCeMedian}{1.64--3.00}

\newcommand{\roTopExactOfLam}{0.96--1.00}
\newcommand{\roGradShareMin}{0.9996}
\newcommand{\roThetaMed}{79.7}
\newcommand{\roThetaQone}{73.1}
\newcommand{\roThetaQthree}{83.2}

\newcommand{\roOneQFO}{798}
\newcommand{\roNoRoomTasks}{27}
\newcommand{\roCeOfTop}{0.76--0.97}

\newcommand{\figOnePct}{45}
\newcommand{\figOneMedian}{2.68}
\newcommand{\figOneAboveTText}{all 100}
\newcommand{\figOneRatio}{0.766}
\newcommand{\figOneRatioPct}{24}
\newcommand{\figOneRatioMed}{0.857}
\newcommand{\figOneRatioQone}{0.773}
\newcommand{\figOneRatioQthree}{0.941}
\newcommand{\figOneTen}{2.75}
\newcommand{\figOneTopOos}{3.01}
\newcommand{\roCeOfTopTen}{0.81--0.98}
\newcommand{\roSqMedianDev}{3.0}
\newcommand{\roSqDevQthree}{7.2}
\newcommand{\roCeAboveSq}{1102}
\newcommand{\roSqTCloser}{4--43}
\newcommand{\roSqOptCloser}{5--30}
\newcommand{\roSqOptCloserSum}{236}
\newcommand{\roSqTCloserSum}{170}
\newcommand{\roSqDiffer}{839}
\newcommand{\roSqTCloserGemma}{43}
\newcommand{\roSqOptCos}{0.50}
\newcommand{\roSqMedianDevTen}{2.4}
\newcommand{\roSqDevQthreeTen}{5.0}
\newcommand{\roSqSentOne}{8.4}
\newcommand{\roSqSentTen}{4.4}
\newcommand{\roJsdMedianTen}{0.95--1.08}
\newcommand{\roSentSteps}{25}
\newcommand{\roOtherSteps}{100}
\newcommand{\gmTraceOpposite}{51 of 52}
\newcommand{\kuOverlapDet}{8}
\newcommand{\kuOverlapOnly}{6}
\newcommand{\kuOverlapSimple}{5}
\newcommand{\kuOverlapN}{50}
\newcommand{\noOverlapTasks}{3}
\newcommand{\noOverlapLo}{2}
\newcommand{\noOverlapHi}{4}
\newcommand{\noOverlapN}{100}
\newcommand{\kuSame}{25}
\newcommand{\kuCases}{28}

\newcommand{\hoIoiN}{13}
\newcommand{\hoRavelRoom}{224}
\newcommand{\hoRavelAbove}{223}
\newcommand{\hoKumonN}{210}
\newcommand{\hoKumonRoom}{97}
\newcommand{\hoKumonAbove}{90}
\newcommand{\hoKumonCe}{0.45}
\newcommand{\hoDisjointRoom}{196}
\newcommand{\hoDisjointAbove}{172}
\newcommand{\hoSentN}{121}
\newcommand{\hoSentRoom}{60}
\newcommand{\hoSentAbove}{54}
\newcommand{\hoSentCe}{0.83}
\newcommand{\hoGemmaRavel}{23}
\newcommand{\hoGemmaRavelAbove}{22}
\newcommand{\hoGemmaIoiCe}{1.53}
\newcommand{\roJsdMedian}{0.91--1.06}
\newcommand{\roBhatMedian}{1.000}
\newcommand{\roBhatQone}{0.997}
\newcommand{\roBhatQthree}{1.003}
\newcommand{\roAngleCeSpearman}{0.71--0.98}
\newcommand{\roAngleTopErr}{2.1--5.6}
\newcommand{\roAngleWeightMedian}{0.03}
\newcommand{\roAngleWeightMax}{0.95}
\newcommand{\roAngleSignedMax}{89.0}
\newcommand{\roCosVOne}{0.89--0.98}
\newcommand{\roIoiTwentyLp}{43--44}
\newcommand{\roJsdMin}{$-0.15$}
\newcommand{\roJsdMax}{1.95}
\newcommand{\gmTasks}{71}
\newcommand{\gmTopAboveLam}{47}
\newcommand{\gmTopAboveLamUncapped}{0}
\newcommand{\gmOtherTopOfLam}{0.96--1.00}
\newcommand{\gmOtherTopOfLamMax}{1.02}
\newcommand{\gmTopOfLam}{1.26}
\newcommand{\gmAngleTopErr}{28}
\newcommand{\gmCeilingTasks}{34}
\newcommand{\gmCapRank}{0.57}
\newcommand{\gmCapRankN}{57}
\newcommand{\gmDeltaPos}{53}
\newcommand{\gmDepth}{1.4}
\newcommand{\gmBend}{0.042}
\newcommand{\gmCapShare}{0.127}
\newcommand{\gmCapShareRaw}{0.136}
\newcommand{\gmCapShareStraight}{0.030}
\newcommand{\gmCapShareRho}{0.996}
\newcommand{\gmDepthRho}{0.88}
\newcommand{\gmDeltaMedian}{0.26}
\newcommand{\gmCGRho}{0.72}
\newcommand{\gmCGN}{29}
\newcommand{\gmNonCGRho}{0.38}
\newcommand{\gmNonCGN}{28}
\newcommand{\gmMove}{0.33}
\newcommand{\gmSourceP}{0.07}
\newcommand{\gmTanhExact}{1.2}
\newcommand{\gmTanhLead}{1.2}
\newcommand{\gmCondMedian}{96}
\newcommand{\gmFamilies}{33}
\newcommand{\gmFamilyRho}{0.65}
\newcommand{\gmWithinFamilyP}{0.14}
\newcommand{\gmCeilAllAbove}{40}
\newcommand{\gmCeilAllTasks}{190}
\newcommand{\gmDevCap}{0.061}
\newcommand{\gmDevUnc}{0.033}
\newcommand{\gmDevOthers}{0.025--0.039}
\newcommand{\gmDevClosed}{100}
\newcommand{\gmCeilingCap}{0.21}
\newcommand{\gmCeilingUnc}{0.00}
\newcommand{\gmCeilingOthers}{0.00}
\newcommand{\gmCeilingClosed}{100}
\newcommand{\gmAngleCap}{0.073}
\newcommand{\gmAngleUnc}{0.056}
\newcommand{\gmAngleOthers}{0.046--0.050}
\newcommand{\gmAngleClosed}{74.5}
\newcommand{\gmGradCap}{0.98049}
\newcommand{\gmGradUnc}{0.99988}
\newcommand{\gmGradOthers}{0.99995--0.99996}
\newcommand{\gmGradClosed}{99.6}
\newcommand{\gmBhatCap}{0.061}
\newcommand{\gmBhatUnc}{0.032}
\newcommand{\gmBhatOthers}{0.027--0.038}
\newcommand{\gmBhatClosed}{100}
\newcommand{\roAccCe}{0.96--1.00}
\newcommand{\roAccOwn}{0.93--0.97}
\newcommand{\roAccSq}{0.86--0.94}
\newcommand{\roAccUpCe}{555}
\newcommand{\roAccDownCe}{13}
\newcommand{\roAccSatCe}{317}
\newcommand{\noCe}{1.20--2.14}
\newcommand{\noCeParent}{1.17--2.09}
\newcommand{\noAbove}{84}
\newcommand{\noTasks}{84}
\newcommand{\noLemma}{0.00}

\newcommand{\npiTopShare}{0.79--0.95}
\newcommand{\npiCeMinusTop}{0.56--0.86}

\newcommand{\npiPb}{0.003--0.006}
\newcommand{\npiMirRest}{0.50--0.51}
\newcommand{\npiMirGemmaTop}{0.72}
\newcommand{\npiMirGemmaCe}{0.68}
\newcommand{\npiTGemma}{8.7}
\newcommand{\npiTOtherLarge}{5.1--5.8}
\newcommand{\npiTopAbsGemma}{44}
\newcommand{\npiTopAbsOtherLarge}{20--21}
\newcommand{\npiTopLowerGemma}{41}
\newcommand{\npiTopLowerOtherLarge}{17}
\newcommand{\npiCeGainLarge}{7.6--8.1}
\newcommand{\cvStartCosTen}{0.995}
\newcommand{\cvStartCosOne}{0.979}
\newcommand{\sbTasks}{1168}
\newcommand{\sbDefined}{1029}

\newcommand{\sbCeGained}{3}
\newcommand{\sbCeSqGained}{3}
\newcommand{\sbSqChangedText}{12 classifications of where the squared error ends up change (GPT-2 small: \texttt{npi\_ever\_subj-relc}, \texttt{ravel\_city\_country}, \texttt{ravel\_city\_language\_json}, \texttt{ravel\_nobel\_prize\_winner\_gender}; Pythia-160m: \texttt{ravel\_nobel\_prize\_winner\_country\_of\_birth\_json}, \texttt{ravel\_physical\_object\_texture}; Pythia-410m: \texttt{cleft}; Gemma-2-9B: \texttt{ravel\_physical\_object\_category\_json}; Qwen2.5-7B: \texttt{NPISimple}; Pythia-6.9b: \texttt{ravel\_occupation\_work\_location\_json}; Qwen2.5-1.5B: \texttt{NPISimple}; Qwen2.5-32B: \texttt{cleft})}
\newcommand{\cvOptCos}{0.986--0.998}
\newcommand{\cvOptGain}{0.3--2.2}
\newcommand{\cvOptGainNinetyLarge}{3--13}
\newcommand{\cvOptMaxLarge}{24}
\newcommand{\ssQwenRho}{30}
\newcommand{\ssQwenR}{18}
\newcommand{\ssQwenSigma}{9}
\newcommand{\ssQwenSigmaFalls}{79}
\newcommand{\ssQwenTasks}{84}
\newcommand{\ssLlamaRho}{41}
\newcommand{\ssLlamaR}{29}
\newcommand{\ssLlamaSigma}{9}
\newcommand{\ssLlamaSigmaFalls}{77}
\newcommand{\ssLlamaTasks}{84}
\newcommand{\ssPythiaRho}{24}
\newcommand{\ssPythiaR}{16}
\newcommand{\ssPythiaSigma}{6}
\newcommand{\ssPythiaSigmaFalls}{74}
\newcommand{\ssPythiaTasks}{84}
\newcommand{\ssLlamaRStep}{$1.7$}
\newcommand{\ssLlamaRStepLo}{$-0.7$}
\newcommand{\ssLlamaRStepHi}{$3.7$}
\newcommand{\ssThirtyRho}{18}
\newcommand{\ssThirtyRhoLo}{11}
\newcommand{\ssThirtyRhoHi}{23}
\newcommand{\ssThirtyR}{9}
\newcommand{\ssThirtyRLo}{5}
\newcommand{\ssThirtyRHi}{11}
\newcommand{\ssThirtySigma}{5}
\newcommand{\ssThirtyTasks}{86}
\newcommand{\ssThirtySigmaLo}{3.5}
\newcommand{\ssThirtySigmaHi}{7.9}
\newcommand{\ssLamPythia}{3.40, 3.61 and 4.00}
\newcommand{\ssLamQwen}{3.15, 3.31 and 4.06}
\newcommand{\ssLamLlama}{3.46, 3.52 and 4.24}
\newcommand{\ssLamThirty}{4.68}

\newcommand{\rkCeilTests}{41{,}286}
\newcommand{\rkCapAbove}{101--195}
\newcommand{\rkCapTests}{426}
\newcommand{\rkCapAboveQ}{4}
\newcommand{\rkCapTestsQ}{132}
\newcommand{\rkAboveOne}{1096}
\newcommand{\rkRoomOne}{1124}
\newcommand{\rkAboveTwo}{1098}
\newcommand{\rkRoomTwo}{1121}
\newcommand{\rkAboveFour}{1097}
\newcommand{\rkRoomFour}{1121}
\newcommand{\rkAboveSixteen}{1127}
\newcommand{\rkRoomSixteen}{1148}
\newcommand{\rkAboveSixtyFour}{1127}
\newcommand{\rkRoomSixtyFour}{1148}
\newcommand{\rkAboveDm}{1129}
\newcommand{\rkRoomDm}{1149}
\newcommand{\rkCeSqOne}{1102}
\newcommand{\rkCeSqTwo}{1099}
\newcommand{\rkCeSqFour}{1105}
\newcommand{\rkCeSqSixteen}{1130}
\newcommand{\rkCeSqSixtyFour}{1125}
\newcommand{\rkLostRavel}{1.94\,$T$, lower bound 0.67}
\newcommand{\rkLostNpi}{0.99 and 1.00}
\newcommand{\rkGained}{5--9}
\newcommand{\rkTopRatio}{1.000--1.007}
\newcommand{\rkTopRatioCap}{1.032--1.038}
\newcommand{\rkKGain}{0.2--1.8}
\newcommand{\rkCeRatio}{0.97--1.01}
\newcommand{\rkCeRatioSixtyFour}{0.97--1.00}
\newcommand{\rkIoiCases}{27}
\newcommand{\rkIoiLp}{43--44}
\newcommand{\rkLanding}{0.63--0.92}
\newcommand{\rkLandingQ}{0.26--0.63}
\newcommand{\rkReach}{0.87--0.97}
\newcommand{\rkReachQ}{0.68--0.87}
\newcommand{\rkTenPass}{1--8}
\newcommand{\rkSqT}{0.94--1.00}
\newcommand{\rkAdequacy}{0.96--1.00}
\newcommand{\rkAboveTenOne}{1102}
\newcommand{\rkAboveTenTwo}{1102}
\newcommand{\rkAboveTenFour}{1102}
\newcommand{\rkAboveTenSixteen}{1131}
\newcommand{\rkAboveTenSixtyFour}{1127}
\newcommand{\rkTopFalls}{3}
\newcommand{\rkShareOne}{0.79--0.97}
\newcommand{\rkShareSixtyFour}{0.91--0.99}
\newcommand{\rkMultiN}{21}
\newcommand{\rkMultiGrow}{0.36--1.74}
\newcommand{\rkMultiOos}{$-0.02$ to $+0.05$}
\newcommand{\rkMultiTop}{$-0.02$ to $+0.13$}
\newcommand{\rkMultiDas}{$-0.02$ to $+0.02$}
\newcommand{\rkIoiCeOne}{0.15--0.23}
\newcommand{\rkIoiCeSixteen}{1.13--1.60}
\newcommand{\rkKfAbove}{10--27}
\newcommand{\rkKfN}{1097}
\newcommand{\rkKfMax}{1.4}
\newcommand{\ceSinCe}{0.15--0.44}
\newcommand{\gcgConstr}{25}
\newcommand{\gcgRankSeven}{$+0.088$}
\newcommand{\gcgRankSevenLo}{$+0.055$}
\newcommand{\gcgRankSevenHi}{$+0.123$}
\newcommand{\gcgThRho}{0.79}
\newcommand{\gcgBestSingleRho}{0.67}
\newcommand{\gcgBestSevenRho}{0.70}
\newcommand{\gcgDoorRho}{0.76}
\newcommand{\gridMaps}{460}
\newcommand{\gridConstr}{59}
\newcommand{\gridRankSeven}{$-0.065$}
\newcommand{\gridRankSevenLo}{$-0.136$}
\newcommand{\gridRankSevenHi}{$+0.002$}
\newcommand{\gridRankSevenMap}{$-0.065$}
\newcommand{\gridRankSevenMapLo}{$-0.091$}
\newcommand{\gridRankSevenMapHi}{$-0.040$}
\newcommand{\gridRankDoor}{$-0.240$}
\newcommand{\gridRankDoorLo}{$-0.363$}
\newcommand{\gridRankDoorHi}{$-0.130$}
\newcommand{\gridThRho}{0.62}
\newcommand{\gridBestSingleRho}{0.45}
\newcommand{\gridBestSingleName}{$\mathbb{E}\,|G_i|$}
\newcommand{\gridBestSevenRho}{0.68}
\newcommand{\gridDoorRho}{0.48}
\newcommand{\gridMinusSingleLo}{$+0.139$}
\newcommand{\gridAttnRho}{$+0.579$}
\newcommand{\gridMinusD}{$+0.112$}
\newcommand{\gridMinusDLo}{$+0.055$}
\newcommand{\gridMinusDHi}{$+0.172$}

\newcommand{\gotherRankSevenC}{$-0.156$}
\newcommand{\gotherRankSevenCLo}{$-0.245$}
\newcommand{\gotherRankSevenCHi}{$-0.065$}
\newcommand{\inMaps}{460}

\newcommand{\inTasks}{78}

\newcommand{\inDegenerate}{2065}
\newcommand{\inLive}{18{,}221}
\newcommand{\inCeAboveG}{16{,}489}
\newcommand{\inCeAboveGstar}{15{,}998}
\newcommand{\inCeAboveGstarTest}{15{,}954}
\newcommand{\inNpPos}{0.54--0.64}
\newcommand{\inNpSigned}{0.51--0.62}
\newcommand{\inNpCons}{0.54--0.63}
\newcommand{\inNpAll}{0.51--0.64}
\newcommand{\inNpBandLo}{0.50}
\newcommand{\inNpBandHi}{0.67}
\newcommand{\inThreeLive}{1218}
\newcommand{\inThreeAboveG}{1190}
\newcommand{\inThreeNotProduced}{0.56--0.65}
\newcommand{\inThreeThroughAttn}{0.57--0.72}
\newcommand{\inAboveSynRange}{94--97}
\newcommand{\inAboveOther}{53--81}
\newcommand{\inExcessMed}{0.14}
\newcommand{\inExcessQone}{0.04}
\newcommand{\inExcessQthree}{0.40}
\newcommand{\inCeilBH}{4894}
\newcommand{\inCeAboveGRaw}{16{,}560}
\newcommand{\inCeAboveGstarRaw}{16{,}099}
\newcommand{\inCeilRaw}{5872}
\newcommand{\inThroughAttnOfNotProduced}{0.47--0.56}
\newcommand{\inAttnRho}{$-0.55$ to $-0.37$}
\newcommand{\inBigLive}{1352}
\newcommand{\inBigCeAboveG}{1322}
\newcommand{\inBigNotProduced}{0.58--0.71}
\newcommand{\inBigThroughAttn}{0.61--0.71}
\newcommand{\inBigNotProducedUnc}{0.65--0.71}
\newcommand{\inBigGemmaNotProduced}{0.58}
\newcommand{\inBigGemmaThroughAttn}{0.61}
\newcommand{\inHalfExcessGstar}{0.27}
\newcommand{\inCeMinusG}{0.11}
\newcommand{\inCeSlope}{1.46}
\newcommand{\inCeIntercept}{0.22}
\newcommand{\inSqTMinusG}{0.09}
\newcommand{\inSqTSlope}{0.80}
\newcommand{\inSqTIntercept}{0.21}
\newcommand{\inSqGSlope}{0.91}
\newcommand{\inSqGIntercept}{0.00}
\newcommand{\inUhatSlope}{0.88}
\newcommand{\inUhatIntercept}{0.00}
\newcommand{\inSqTAttn}{0.55}
\newcommand{\inOwnKeep}{0.90}
\newcommand{\inCeAttnAll}{0.39}
\newcommand{\inExcessAttnAll}{0.58}
\newcommand{\inAttDirect}{0.56--0.66}
\newcommand{\inNpCtx}{0.52--0.62}
\newcommand{\inAboveGstarCtx}{16{,}110}
\newcommand{\inRowExcess}{0.43}
\newcommand{\inOffRowRemoved}{0.61}
\newcommand{\inCtxGainMed}{0.002}
\newcommand{\inCtxGainQnine}{0.05}
\newcommand{\inBisSum}{2.11}
\newcommand{\inUhatSum}{0.88}
\newcommand{\inBhatSum}{0.93}
\newcommand{\inOffRoute}{4300}
\newcommand{\inZero}{4630}
\newcommand{\inOffRouteIndet}{152}
\newcommand{\inOffRouteBH}{4270}
\newcommand{\inOffRouteGstarHi}{0.001}
\newcommand{\inOffRouteGstarHiNinety}{0.013}
\newcommand{\inOffRouteBelowZero}{1147}
\newcommand{\inOffRouteDasMed}{0.04}
\newcommand{\inOffRouteDasNinety}{0.43}
\newcommand{\inOffRouteAttn}{0.75}
\newcommand{\inZeroSqT}{0.04}
\newcommand{\inZeroJsd}{0.02}
\newcommand{\inZeroSqTAttn}{0.66}
\newcommand{\inZeroJsdAttn}{0.68}
\newcommand{\inZeroMax}{1.40}
\newcommand{\inZeroAtT}{36}
\newcommand{\inZeroSqTAboveGstar}{4368}
\newcommand{\inZeroJsdAboveGstar}{3791}
\newcommand{\inZeroSqTOff}{4354}
\newcommand{\inZeroJsdOff}{3720}
\newcommand{\inCeilGstarMed}{0.01}
\newcommand{\inCeilAttn}{0.85}
\newcommand{\inAccUpCe}{8800}
\newcommand{\inAccDownCe}{30}
\newcommand{\inAccSatCe}{322}
\newcommand{\inAccCeMed}{0.33}
\newcommand{\inAccOwnMed}{0.16}
\newcommand{\inTraceMid}{1.01--1.03}
\newcommand{\inTraceMidCloser}{79--86}
\newcommand{\inTraceOvershoot}{5--10}
\newcommand{\inFoUntrained}{0.96--1.03}
\newcommand{\inFoDas}{1.05--1.19}
\newcommand{\inSdasNpiFg}{0.57--0.72}
\newcommand{\inSdasAgr}{0.16--0.26}
\newcommand{\inSdasOther}{0.19--0.30}
\newcommand{\inDoorFirstOrder}{0.93--0.99}
\newcommand{\inDoorBlockLShare}{36--66}
\newcommand{\oeTasks}{29}

\newcommand{\oeCells}{114}
\newcommand{\oeEffects}{104{,}400}
\newcommand{\oeReadoutMax}{$2.3\times10^{-4}$}
\newcommand{\oeSpecMax}{$5.5\times10^{-6}$}
\newcommand{\oeAngleMax}{$1.8\times10^{-7}$}
\newcommand{\oeInsideMax}{$3.3\times10^{-4}$}
\newcommand{\oeInsideLam}{$6.6\times10^{-5}$}
\newcommand{\oeInsideWD}{$7.8\times10^{-4}$}
\newcommand{\oeOneQ}{26}
\newcommand{\srTestPrompts}{22--28}
\newcommand{\srBlocks}{11--14}
\newcommand{\snSentRange}{10--21}
\newcommand{\inMapHours}{0.01--5.4}
\newcommand{\inMapHoursMed}{0.6}
\newcommand{\cgrLdaRepro}{11--18}

\newcommand{\cgrMeanRel}{0.58--0.78}
\newcommand{\cgrMeanFixed}{0.96--0.98}
\newcommand{\cgrProbeFixed}{0.96--1.00}
\newcommand{\cgrPcaKmFixed}{0.12--0.36}
\newcommand{\cgrDasFixed}{1.9--3.7}

\newcommand{\tokNumLo}{4}
\newcommand{\tokNumHi}{28}
\newcommand{\tokGenLo}{62}
\newcommand{\tokGenHi}{66}

\newcommand{\srPubOurs}{114.7}
\newcommand{\srPubLo}{85.0}
\newcommand{\srPubHi}{137.4}
\newcommand{\srPythiaY}{198.8}
\newcommand{\srPythiaLo}{91.9}
\newcommand{\srPythiaHi}{271.0}
\newcommand{\srPythiaPlainLo}{104.8}
\newcommand{\srPythiaPlainHi}{262.3}
\newcommand{\srPythiaYLo}{198.5}
\newcommand{\srPythiaYHi}{199.0}

\newcommand{\srOwnMax}{121.7}
\newcommand{\srOwnSecond}{120.3}
\newcommand{\srOwnLo}{104}

\newcommand{\srOwnLayers}{14 and 15}
\newcommand{\srBLCosLo}{0.997}
\newcommand{\srBLDiffSel}{0.9}
\newcommand{\srBLDiffMax}{20}
\newcommand{\ruSmall}{1--3}
\newcommand{\ruThree}{4}
\newcommand{\ruLarge}{4--5}
\newcommand{\ruGemma}{14}
\newcommand{\ruThirtyTwo}{2}
\newcommand{\gridRavelRho}{0.32}
\newcommand{\gridSrcIoi}{$+0.144$ [$+0.092$, $+0.202$]}
\newcommand{\gridSrcNumOnly}{$+0.041$ [$+0.018$, $+0.061$]}
\newcommand{\gridSrcKumon}{$-0.002$ [$-0.075$, $+0.056$]}
\newcommand{\gridSrcRavel}{$-0.449$ [$-0.500$, $-0.400$]}
\newcommand{\gridSrcSent}{$-0.246$ [$-0.326$, $-0.172$]}
\newcommand{\gridSrcSst}{$-0.132$ [$-0.229$, $-0.054$]}
\newcommand{\snReuseWord}{six}
\newcommand{\roModelsWord}{fourteen}
\newcommand{\roModelsFOWord}{thirteen}
\newcommand{\roSourcesWord}{six}

\newcommand{\inExcludedMapsWord}{eight}
\newcommand{\gridModelsBelowWord}{four}
\newcommand{\kuUpWord}{two}
\newcommand{\oeMapsWord}{three}
\newcommand{\rkModelsWord}{fourteen}
\newcommand{\tokEmptyWord}{eight}
\newcommand{\tokTokenizersWord}{four}
\newcommand{\srModelsWord}{seven}

\newcommand{\srIndistPlainWord}{five}
\newcommand{\srBelowWord}{one}

\newcommand{\cgDasMedLo}{2.1}
\newcommand{\cgDasMedHi}{3.1}
\newcommand{\cgDasAbove}{251}
\newcommand{\cgEntries}{261}

\newcommand{\cgEntriesAligned}{90}
\newcommand{\cgOtherAbove}{112}
\newcommand{\cgOtherEntries}{1740}
\newcommand{\cgOtherAboveAligned}{76}
\newcommand{\cgOtherEntriesAligned}{600}

\newcommand{\cgOtherExceedMed}{1.03}

\newcommand{\cgFiltDasMedLo}{2.0}
\newcommand{\cgFiltDasMedHi}{2.9}

\newcommand{\cgFiltOtherAbove}{94}
\newcommand{\cgFiltOtherEntries}{1627}
\newcommand{\cgSelLo}{0.52}
\newcommand{\cgSelHi}{1.70}
\newcommand{\cgMeanAlLo}{0.96}
\newcommand{\cgMeanAlHi}{1.00}
\newcommand{\cgMeanOtLo}{0.54}
\newcommand{\cgMeanOtHi}{0.71}

\newcommand{\cgmMaps}{174}
\newcommand{\cgmRbarMed}{2.04}
\newcommand{\cgmRbarQone}{1.46}
\newcommand{\cgmRbarQthree}{2.71}
\newcommand{\cgmRbarAboveOne}{172}
\newcommand{\cgmRbarModelLo}{2.0}
\newcommand{\cgmRbarModelHi}{2.2}
\newcommand{\cgmSNlo}{1.02}
\newcommand{\cgmSNhi}{1.08}
\newcommand{\cgmSNtrlo}{1.01}
\newcommand{\cgmSNtrhi}{1.05}
\newcommand{\cgmSNsamelo}{1.01}
\newcommand{\cgmSNsamehi}{1.05}

\newcommand{\cgmSoffSmallMax}{0.02}

\newcommand{\cgmSoffQwen}{0.40}
\newcommand{\cgmSoffQwenStrict}{0.00}
\newcommand{\cgmSoffPooled}{0.11}
\newcommand{\cgmSoffPooledLo}{0.01}
\newcommand{\cgmSoffPooledHi}{0.21}
\newcommand{\cgmSoffPooledStrict}{0.03}
\newcommand{\cgmSoffPooledStrictLo}{0.01}
\newcommand{\cgmSoffPooledStrictHi}{0.05}
\newcommand{\cgmQwenFgT}{0.12}
\newcommand{\cgmPythiaFgT}{-0.10}
\newcommand{\cgmSoffQwenNoFg}{0.01}
\newcommand{\cgmSoffPooledNoFg}{0.03}
\newcommand{\cgmQwenFgPart}{0.39}
\newcommand{\cgmMapBeyondLo}{0.99}
\newcommand{\cgmMapBeyondHi}{1.01}

\newcommand{\cgmAttLo}{0.21}
\newcommand{\cgmAttHi}{0.40}
\newcommand{\cgmPubN}{30}
\newcommand{\cgmPubVanEq}{25}
\newcommand{\cgmPubDasIn}{28}
\newcommand{\cgmPubRho}{0.99}
\newcommand{\cgmPubRhoShift}{0.95}
\newcommand{\cgmNpiRbar}{2.36}

\newcommand{\cgmNpiPub}{2.38}
\newcommand{\cgmNpiLicLast}{8}
\newcommand{\cgmNpiAtt}{0.67}
\newcommand{\cgmNpiFrom}{6}
\newcommand{\cgmNpiDasLo}{1.1}
\newcommand{\cgmNpiDasHi}{1.4}
\newcommand{\cgmNpiGsMax}{0.06}
\newcommand{\cgmNpiFrozenRemoves}{97}
\newcommand{\cgmRouteLicLast}{3}
\newcommand{\cgmRouteThat}{4}
\newcommand{\cgmRouteHasLast}{8}
\newcommand{\cgmRouteCaughtFrom}{9}
\newcommand{\cgmGptNpiRbar}{2.43}

\newcommand{\computeplace}{ on UCloud (SDU eScience Center)}
\def\competing{}
\newcommand{\codeurl}{at \url{https://github.com/tanyelai/whose-effect}}

\ShortHeadings{Whose Effect Is It?}{Tanyel and Ayvaz}
\firstpageno{1}

\begin{document}

\title{Whose Effect Is It?\texorpdfstring{\\}{} How Objectives Inflate Learned Subspace Interventions}
\author{\name Toygar Tanyel\thanks{This author is a visiting researcher and the corresponding author.} \email contact.toygar@gmail.com \\
       \name Serkan Ayvaz \email seay@mmmi.sdu.dk \\
       \addr Centre for Industrial Software\\
       University of Southern Denmark}
\editor{}
\begin{NoHyper}\maketitle\end{NoHyper}   %
\thispagestyle{plain}

\begin{abstract}%
Interpretability often locates concepts by copying part of a language model's state from one input into another (activation
patching). A larger
output change counts as stronger evidence that this part carries the concept. Methods that learn which part to copy often
change the output more than the whole state does. By that reading, learned parts isolate the concept more cleanly. Yet the
whole contains the part. So whose effect is this excess, the network's or the method's? Copying the whole state learns
nothing, so it shows what the network produces. We prove that, to first order, the effects of orthogonal parts add up to that
of the whole. A part therefore exceeds the whole exactly when the rest moves the output back, which does not require
isolating the concept more cleanly. Objectives that reward a larger effect select such parts. Distributed alignment search
with cross-entropy, for example, exceeds the whole-state copy just before the output on almost all tasks, and inside the
network at most sites. In both, changing only its objective to target what the network produces removes most of the
excess. It therefore comes largely from the objective, so a larger learned effect alone is not stronger evidence.
\end{abstract}


\clearpage
\setcounter{tocdepth}{1}            %
\etocsettocstyle{\noindent{\large\bfseries Contents}\par\medskip}{}
\tableofcontents
\setcounter{tocdepth}{2}            %
\clearpage
  \phantomsection\addcontentsline{toc}{section}{Terms and Symbols}  %
  {\small\setlength{\tabcolsep}{5pt}   %
  \begin{longtable}{@{}>{\raggedright\arraybackslash}p{0.19\textwidth}>{\raggedright\arraybackslash}p{0.58\textwidth}>{\raggedright\arraybackslash}p{0.18\textwidth}@{}}
    \toprule
    term & meaning & defined \\
    \midrule
    \endhead
    \bottomrule
    \noalign{\vskip 4pt}
    \caption{\textbf{Terms and symbols}, in the order a reader meets them, with where each is defined.}\label{tab:glossary}\\
    \endlastfoot
    $x_b$, $x_s$; $y_b$, $y_s$ & a pair's base and source inputs; the labels, the next token each expects & \cref{sec:pairs} \\
    mirror; mirrored & a pair with base and source exchanged; a test set that holds every pair with its mirror &
      \cref{sec:pairs}; \cref{app:error} \\
    $z$ & the log odds of the source label over the base label, $o_{y_s} - o_{y_b}$ & \cref{sec:pairs} \\
    $T$ & the total effect, the change in $z$ when the whole input is replaced, as in causal mediation analysis & \cref{sec:pairs} \\
    site; cell; map & a set of (layer, position) locations in the residual stream; one region of the sentence at one layer; an
      effect at every cell of a model for one task & \cref{sec:sites} \\
    $h_b$, $h_s$, $\Delta$ & the base and source runs' states at a location, and $\Delta = h_s - h_b$ & \cref{sec:sites} \\
    whole-state patching & replacing the whole state at a site; its effect $G$ is the indirect effect through the site
      (CausalGym's \emph{vanilla}) &
      \cref{def:own} \\
    $e(d)$ & the effect of the rank-one interchange along a unit direction $d$ & \cref{def:own} \\
    inflated; excess & a learned effect above $G$, the effect of whole-state patching at the same site; the difference &
      \cref{sec:sites} \\
    effect-seeking, target-aimed, representation-only & objectives that reward a larger effect, aim at a target effect, or
      ignore the effect & \cref{sec:objectives} \\
    cut; complete; partial; readout & a site whose states, with the inputs both runs share, fix the output; complete: every
      position of one layer, or the readout, the last block's output at the last position; partial: a proper part of a layer,
      such as one position, that is not itself a cut & \cref{def:cut} \\
    $G^*$, best context & the effect of whole-state patching at a cell in its best context: the cell alone, or the effect it adds when
      swapped together with one other region of the same layer, whichever is larger on the training pairs &
      \cref{sec:offroute} \\
    to first order & terms of second and higher order in $\Delta$ are neglected & \cref{app:proofs-range} \\
    $w$, $M$ & the gradient of $z$ in the state at a location, at the midpoint between the base and source states (the readout gradient at the readout), and $M = \operatorname{sym}\E[w\Delta^\top]$ & \cref{app:proofs-range} \\
    the range; the trace & $[\lambda_D, \lambda_1]$, the first-order effects of the rank-one interchanges at a site, with upper
      end $\lambda_1$, the largest eigenvalue of $M$ and the predicted maximum; $\operatorname{tr}M$, to first order the effect of whole-state patching &
      \cref{thm:range} \\
    $\hat m$, $\hat w$; $\theta$ & the directions of the mean change of the state and of the readout gradient; the angle
      between them &
      \cref{prop:natural} \\
    maximum effect (out-of-sample maximum); in-sample maximum; predicted maximum & the largest mean effect of one direction,
found on training pairs and measured on test pairs; the largest exact effect of one direction, found and measured on test
pairs; the largest first-order effect of one direction, $\lambda_1$, the upper end of the range &
      \cref{sec:tasks,app:readout} \\
    case; margin above $T$ & one task in one model; the maximum effect minus $T$, and a case has a margin above $T$ when the
      interval of its maximum effect lies above $T$ & \cref{sec:tasks} \\
    $\hat u$, $\hat b$, $v_1$ & measured: the class-signed mean change of the state, the mean readout gradient, the top eigenvector of
      $M$ & \cref{app:quantities} \\
    through attention weights (through attention) & the part of an effect that freezing every downstream attention pattern
removes & \cref{app:evaluation} \\
    two-label task & a task whose pairs all contrast the same two answers (one label pair) & \cref{sec:inflation} \\
    licensor & the word that licenses the negative-polarity item (\emph{No} against \emph{The}) & \cref{sec:inflation,sec:published} \\
    route; route test; off-route; off-route effect & the cells of a map at which the interval of $G^*$ lies above zero; the
test of a cell against $G^*$; a cell where, for a
given direction (cross-entropy DAS unless stated otherwise), the interval of its effect $e$ lies above zero, that of $G^*$
does not, and $e$ lies significantly above $G^*$; $e$ at such a cell & \cref{sec:offroute,alg:route} \\
    live & of a cell: reached by the input's change, when some test pair changes the tokens up to it & \cref{app:fullgrid} \\
    
  \end{longtable}}

\clearpage

\begin{figure}[tbp]
  \centering
  \includegraphics[width=\textwidth]{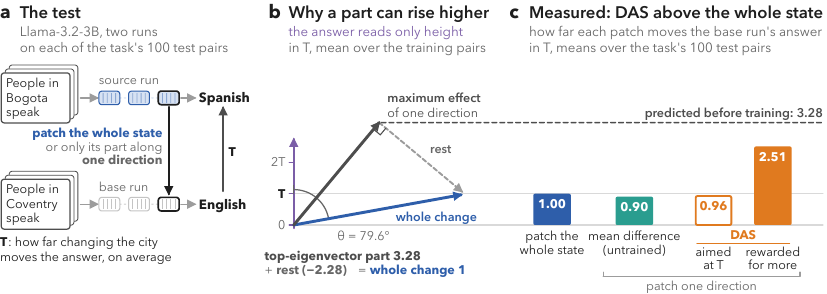}
  \caption{\textbf{A learned part of a state can change a model's output more than the whole state does.} Llama-3.2-3B
  \citep{dubey2025} on RAVEL's city-to-language task \citep{huang2024}, a typical case. (a)~The
  patched state is the readout, where whole-state patching gives exactly $T$. The direction is the
  same for every pair. (b)~The first-order picture (\cref{prop:natural}). With two answers and uncapped
  logits, the readout gradients of all pairs share almost exactly one direction, the readout direction, drawn as height.
  Its measured angle $\theta$ to the mean change of the state predicts the maximum effect before
  training. (c)~DAS is trained with cross-entropy (rewarded for more) or a per-pair squared error aimed at $T$. With
  cross-entropy, it lies above $T$ on \figOneAboveTText{} test pairs.}
  \label{fig:law}
\end{figure}
\suppressfloats[t]

\section{Introduction}
\label{sec:intro}

A standard way to study how a language model computes is to intervene on it. An internal state is changed, and the
resulting change in the output is measured. Much of our current understanding of the internal mechanisms of language models
rests on such evidence. This understanding is considered critical for their safety \citep{bereska2024}, and the same
evidence guides how models are edited and steered \citep{meng2022,rimsky2024}. A common form of such interventions is the interchange intervention, also known as activation patching
\citep{zhang2023,heimersheim2024}. The model runs on a base input, part or all of one internal state is replaced by its value
on a source input, and the output is observed \citep{geiger2021,geiger2023a}. Distributed alignment search (DAS) learns which
part of the state to replace. It trains a subspace so that the patched output follows a high-level causal model
\citep{geiger2023,wu2023}. Benchmarks rank such localisation methods by how strongly the intervention produces the
counterfactual output, or by how often it does so, a score called interchange intervention accuracy (IIA)
\citep{arora2024,huang2024,mueller2025}.

Published results contain a puzzle. A learned part of a state often changes the output more than the whole state does (\cref{fig:law}). For
example, on the CausalGym benchmark, the score of DAS is a median of two to three times that of patching the whole
state. \citet{arora2024} report that DAS ``consistently finds
the most causally-efficacious features''. In another case, \citet{hollinsworth2024} train a one-dimensional DAS direction
for sentiment, and their code patches it at every position of a layer. At its best layer, the direction scores 109.8\% on their logit-difference metric,
while patching the whole state scores exactly 100\%. If these scores are taken to measure what the network
itself computes, they imply that a part of a state contributes more than the whole state. Is this excess an effect of the network, or of the training objective?

Earlier work has reported such excesses and related problems, and has proposed causes for them. \citet{makelov2023} attribute one case to a pathway
that the network leaves dormant, an interpretation that \citet{wu2024} dispute. \citet[\S7]{sutter2025} note that a linear alignment map can ``sometimes achieve IIA higher than the DNN's accuracy'' and suggest that it ``may also `learn the
task'\,''. \citet{grant2025} show that interventions can push internal states off the network's distribution and activate
hidden pathways. \citet[\S1]{bao2026} argue that objectives that maximise probability make steering prone to overfitting, and
they propose CDAS, which trains the interchange of DAS with a distribution-matching objective instead. However, none of these works says in general when a learned
part exceeds the effect of patching the whole state, by how much, or where a given kind of training objective places it.

Our goal is to separate the part of a learned effect that the network itself produces from any excess beyond it, and to
answer these three questions. Such a separation requires a reference that involves no training. We use
whole-state patching, which copies the entire state of the source run at the site of the learned intervention and learns
nothing. CausalGym,
RAVEL and MIB \citep{arora2024,huang2024,mueller2025} already report it as a baseline; we make it the reference for every
method. At the readout, the last state before the output, whole-state patching has exactly the same effect as replacing the
whole input, that is, the total effect $T$ of the input.

Our analysis rests on a decomposition. To first order, when a state is split into orthogonal parts, the effects of
patching the parts add up to the effect of patching the whole state, whichever method chose the parts. A part therefore
exceeds the whole exactly when the other parts move the output back, which does not require that it isolate the high-level
variable more cleanly. The network fixes the range of effects that a part can have, and the upper end of this range can be computed
before any training. The training objective decides where in this range a learned part ends up.

Our contributions are as follows.
\begin{enumerate}
  \item We prove the decomposition at any rank, and that objectives rewarding a larger effect select the upper end of
        the range. Lean checks every first-order statement.
  \item At the readout, changing only the objective moves DAS from far above $T$ (cross-entropy) to close to $T$ (a squared
        error aimed at $T$), across \roModelsWord{} models.
  \item Inside the network, cross-entropy DAS exceeds whole-state patching at most sites, including sites where the
        latter shows no significant effect, mostly through attention.
  \item On CausalGym, DAS's lead over whole-state patching is an excess, not a better region. We propose a three-step
        check whose values our library OwnEffect computes.
\end{enumerate}

The paper is organised as follows. \Cref{sec:setup} defines whole-state patching and the other quantities we use, and
\cref{sec:theory} develops the theory. \Cref{sec:experiments} describes the experimental setup. We then test the theory in
three settings. At the readout, the effect of whole-state patching is known exactly (\cref{sec:inflation}), while inside the
network it has to be measured (\cref{sec:offroute}). The third setting is that of published benchmark results
(\cref{sec:published}). Alongside each result, we test the alternative explanations that a reader might raise.
\Cref{app:owneffect} turns the comparison into a reporting check. \Cref{sec:related} reviews related work, \cref{sec:discussion} discusses implications and limitations, and
\cref{sec:conclusion} concludes. The appendices give further details.

\section{Preliminaries}
\label{sec:setup}

To decide whether a learned effect comes from the network or from the training objective, we need a unit of measurement
and a reference. Our unit is the total effect of the input, and our reference for every learned effect is the effect of
whole-state patching at the same site. \Cref{sec:published} compares published scores with this reference in their own
units.

\subsection{Input Pairs and the Total Effect}
\label{sec:pairs}

A model maps a token sequence to next-token logits $o$ at the last position, and all runs are deterministic (evaluation
mode). A \emph{pair} consists of a base input $x_b$ and a source input $x_s$. Their expected next tokens are $y_b$ and $y_s$.
The \emph{mirror} of a pair is the same pair with the two inputs exchanged. We measure the output by the log odds of the source label over the
base label, $z = o_{y_s} - o_{y_b}$. The \emph{total effect} of the source input is $T = z(x_s) - z(x_b)$, the change in $z$
when the whole input is replaced. Because the counterfactual label is the source input's own next token, $T$ is the effect
that the counterfactual asks for. In the words of CausalGym, a successful intervention ``makes the model behave as if its
input was'' the source \citep{arora2024}. The total effect $T$ is the value that CausalGym's log odds ratio \citep{arora2024} takes when the whole input is replaced,
and it is the unit of logit-difference recovery \citep{zhang2023,heimersheim2024}. It is also the total effect of the
input on $z$ in causal mediation analysis, which \citet[Definition~52]{geiger2023a} express with interchange interventions. We report each mean
effect over test pairs divided by the mean $T$, that is, as a multiple of $T$. The symbols $T$, $G$ and $e$ (\cref{def:own})
denote means in log odds unless a statement holds for each pair, and $T_i$, $G_i$ and $e_i$ are the values for pair $i$.

Interchange intervention accuracy (IIA), the usual score in causal abstraction, is the fraction of pairs on which the patched
output agrees with the counterfactual output of a high-level causal model, usually judged by the most likely token. On our
tasks, this output is $y_s$, and we count a pair when the patched output prefers $y_s$ to $y_b$. This count applies a
threshold to the same quantity $z$, so it can also show an effect above that of whole-state patching. On pairs where whole-state patching already makes the output prefer $y_s$, however, it hides the excess
(\cref{sec:accuracy}).

\subsection{Sites and Interchange Interventions}
\label{sec:sites}

\begin{figure}[p]
  \centering
  \includegraphics[width=\textwidth,height=\dimexpr\textheight-66pt\relax,keepaspectratio]{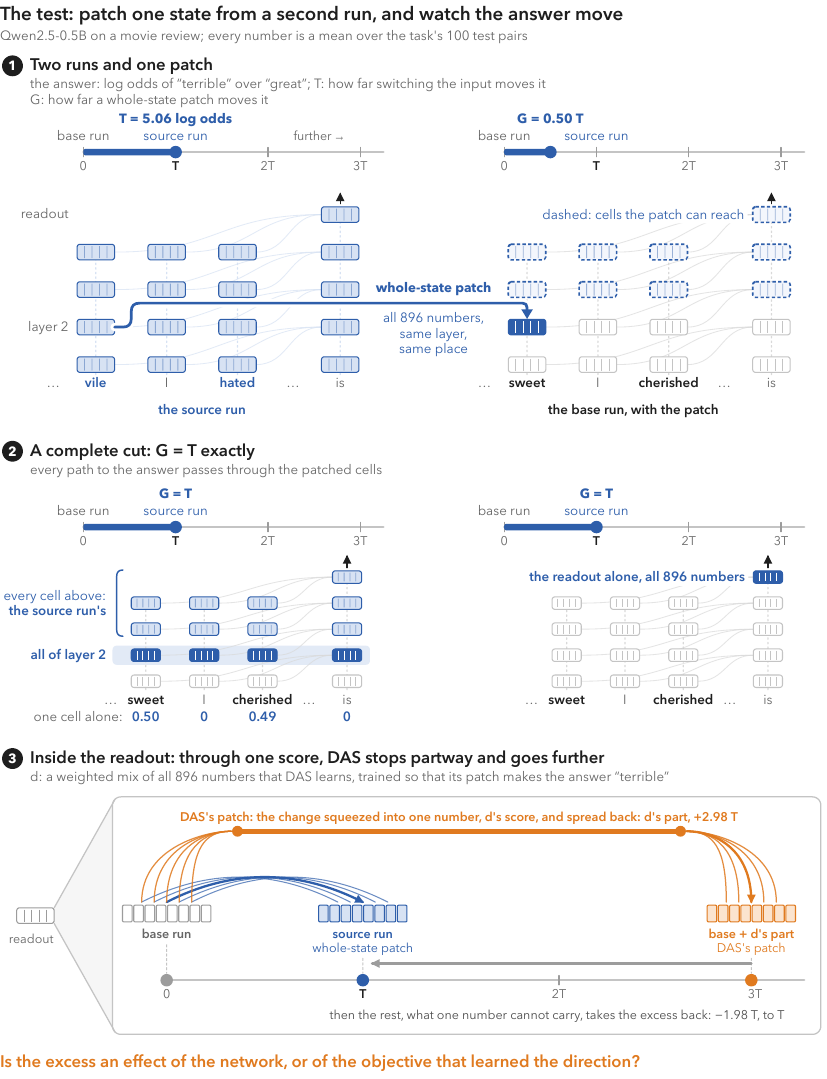}
  \caption{\textbf{Patching a complete cut moves the answer by exactly $T$; patching one learned direction of the readout
  moves it three times as far.} All of a layer and the readout are complete cuts (\cref{prop:cut}). The direction $d$ is
  that of DAS with cross-entropy, one for all test pairs; its effect is the mean of three training runs.
  \Cref{fig:guidec} shows why the rest pulls back.}
  \label{fig:guidea}
\end{figure}

A \emph{site} is a set $S$ of (layer, position) locations in the residual stream, where layer $\ell$ is the output of block
$\ell$, counting from 0. A \emph{cell} is one region of the input sentence, such as the
subject, at one layer, and a \emph{map} shows an effect at every cell of a model for one task. \Cref{tab:glossary} lists the terms we use, and a running example illustrates them (\cref{fig:guidea}). We write $h_b$ and $h_s$ for the states that the base and source runs compute at a location, and
$\Delta = h_s - h_b$.

\begin{definition}[whole-state patching]
\label{def:own}
\emph{Whole-state patching} at $S$ runs the base input with the states at $S$ replaced by those of the source run. Its
effect $G(S)$ is the resulting change in $z$. A \emph{subspace interchange} at a single location along a matrix $P$ with
orthonormal columns replaces $h_b$ by $h_b + PP^\top\Delta$. For rank one, $P=d$ with $\|d\|=1$, and we write $e(d)$ for its
effect.
\end{definition}

Whole-state patching learns nothing; it transfers whatever the network computes at $S$ and nothing else. CausalGym calls it
the vanilla intervention, and standard activation patching performs it \citep{arora2024,heimersheim2024}. Its effect
$G(S)$ is the indirect effect of the input through $S$ in causal mediation analysis \citep[Definition~52]{geiger2023a}, as
in causal tracing \citep{meng2022}. We use $G(S)$ as the reference for every method, because it is the effect that the
network itself produces at the site. We call a learned effect \emph{inflated} when it exceeds $G$, the effect of
whole-state patching at the same site, and we call the difference its \emph{excess}.

\subsection{Training Objectives}
\label{sec:objectives}

Every method that CausalGym ranks applies a subspace interchange, and the methods differ only in how they choose $P$.
DAS trains $P$ so that the patched output follows the source label. Its loss is cross-entropy
\citep{geiger2021a,geiger2023,arora2024}, a linear logit difference \citep{hollinsworth2024} or, in CDAS, a Jensen--Shannon
divergence to the next-token distribution of the source run \citep[\S3.3]{bao2026}. Other methods choose $P$ without running the intervention, from the states alone or together with their class labels, for
example a difference of means between classes, a linear probe, PCA, $k$-means or LDA
\citep{arora2024,hollinsworth2024}. A \emph{class} groups the inputs of a task by type, for example sentences with a singular
or with a plural subject.

The usual distinction between supervised and unsupervised methods \citep[Section~3.6]{geiger2023a} places a linear probe
and DAS in the same group. However, the probe is fitted to the states, while DAS is fitted to the effect of the
intervention. Since our results depend on the objective, we group methods by what their objective requires of that
effect. An
\emph{effect-seeking} objective keeps rewarding a larger effect, as DAS does with cross-entropy or a logit difference. A
\emph{target-aimed} objective is minimised at a target effect, for example a squared error to the total effect. A
\emph{representation-only} objective ignores the effect of the intervention altogether. Methods that choose $P$ without running the intervention have this kind of objective, and so does the readout gradient (the gradient of $z$ with
respect to the state at the readout), which depends on the output but not on the intervention. Our theory concerns these
kinds of objective rather than one particular method, and the effect-seeking method we measure is DAS.

\subsection{Complete and Partial Cuts}
\label{sec:cuts}

Whether whole-state patching gives exactly $T$ depends on the site.

\begin{definition}[cuts]
\label{def:cut}
A \emph{cut} for a pair is a site $C$ such that the output is a function of the states at $C$ and of side inputs (weights,
position indices, attention mask) that are the same in the base and the source run. A cut is \emph{complete} if it is of
one of the following two kinds.
\begin{enumerate}
  \item[(a)] All positions of one layer, in a model whose blocks read only the states of the previous block, for a base and a
             source of the same length with aligned positions and the same padding.
  \item[(b)] The \emph{readout}, the output of the last block at the last position. Only the final normalisation and the
             unembedding (and, in Gemma-2, its logit cap) follow it.
\end{enumerate}
A \emph{partial cut} is a proper subset of a kind-(a) cut that is not itself a cut, because the output also reads the other
positions. An example is the single position in one layer that most published maps patch.
\end{definition}

At a cut, the output depends on the input only through the states at the cut. Whole-state patching at a complete cut therefore gives exactly
the total effect $T$ (\cref{prop:cut}). At a partial cut it need not (\cref{rem:partial}). In the running example (\cref{fig:guidea}),
the two sentences differ in two words, the adjective and the verb. Patching either word alone at layer 2 moves the answer
about half as far as patching the whole layer, because the output still reads the other word as in the base run.

\subsection{A Running Example}
\label{sec:running}

Six figures illustrate the argument on one sentence, a movie review from the sentiment templates of
\citet{hollinsworth2024}, processed by Qwen2.5-0.5B. \Cref{fig:guidea} sets up the comparison, and \cref{fig:guidec} explains why a part can exceed the whole. \Cref{fig:guideb}
marks where whole-state patching shows an effect of the sentiment and what DAS reports there. \Cref{fig:guided} shows how a
learned direction can raise the attention to a site where whole-state patching leaves it as in the base run,
\cref{fig:guidef} that a learned effect follows the target of its objective, and \cref{fig:guidee} shows what to report, with one cell of a second template. Among the
six small models whose maps cover this task, Qwen2.5-0.5B lies in the middle at the readout, where DAS with cross-entropy
gives 3.0\,$T$, against 2.0 to 4.8\,$T$ across the six models.

In the figures, $G$ is the effect of whole-state patching at a cell or a set of cells, and $G^*$ is its effect in the cell's best context
(\cref{sec:offroute}).
Unless a figure states otherwise, every number in the figures is a mean over the task's 100 test pairs, that is, 50 pairs
and their mirrors.

\section{Theoretical Analysis}
\label{sec:theory}

This section answers the three questions of the introduction with four results, one exact and three to first order. We
first show that whole-state patching gives exactly $T$ at a complete cut. We then show when a part exceeds whole-state
patching and by how much, where a training objective places it, and how large the excess is when all pairs share one readout
direction. The rest of the paper builds on these results, and every first-order statement is machine-checked in Lean
(\cref{app:lean}).

\subsection{Whole-State Patching and the Total Effect}
\label{app:proofs-cut}

\begin{restatable}[complete-cut identity]{proposition}{propcut}
\label{prop:cut}
For each pair and complete cut $C$, $G(C) = T$ exactly.
\end{restatable}
\begin{proof}
Downstream of $C$, the output is computed from the states at $C$ and from side inputs that agree between the two runs. For
kind~(a), equal lengths, aligned positions and the same padding make the position indices and the attention mask agree. For
kind~(b), only the final normalisation, the unembedding and, in Gemma-2, the logit cap follow, and they read that one state. After the interchange the
states at $C$ are the source run's, so the patched output is the source run's output, and $G(C) = z(x_s) - z(x_b) = T$.
\end{proof}

This identity has precedents. In causal mediation analysis, it states that a complete cut fully mediates
the effect of the input, so that the indirect effect through the cut equals the total effect. \citet[Remark~53]{geiger2023a}
show that this equality is equivalent to an abstraction in which the input affects the output only through the mediator.
For the cuts of \cref{def:cut}, the equality holds for every pair, because they are defined through the computation itself. In activation patching, denoising metrics assign 100\% to the clean run, here the source run, and at a
kind-(a) cut this level equals $T$, as in the code of \citet{hollinsworth2024}. Beyond these precedents, our contribution is
to read an effect above $T$ at a complete cut as an effect that whole-state patching at that cut cannot produce.

\begin{remark}[partial cuts]
\label{rem:partial}
At a partial cut, $G(S)$ is not bounded by $T$. In Pythia-410m, on the sentiment task of \citet{hollinsworth2024}, whole-state patching at the adjective and
verb positions restores \srOwnSecond\% and \srOwnMax\% of $T$ at layers \srOwnLayers{}, with interval lower
bounds of at least \srOwnLo\%. A run that mixes base and source states therefore changes the output more than the source run
does. Among the \srModelsWord{} models of our re-run of the sentiment experiment (\cref{tab:sentiment}), whole-state patching
exceeds 100\% only in this one. Whether or not $G(S)$ exceeds $T$, at a partial cut the reference is $G(S)$ itself, not $T$.
\end{remark}

\subsection{The Range of Subspace Effects}
\label{app:proofs-range}

Fix one location whose hidden dimension is at least 2. For pair $i$, let $F_i(h)$ be the value of $z$ when the base run
carries the state $h$ at this location and everything after it is recomputed. Let $w_i = \nabla F_i(h_{b,i} + \tfrac12\Delta_i)$
be its gradient at the midpoint between the base and the source state, and define
\begin{equation}
  M \;=\; \operatorname{sym}\,\E\big[w\,\Delta^\top\big], \qquad \operatorname{sym}A = \tfrac12(A + A^\top),
  \label{eq:M}
\end{equation}
with eigenvalues $\lambda_1 \ge \dots \ge \lambda_D$, where $D$ is the hidden dimension. Here and below, \emph{to first order} means that terms of second and higher order in $\Delta$
are neglected. This assumes that $F_i$ is three times continuously differentiable on
the ball with diameter $[h_b, h_s]$, which contains every subspace-patched state (\cref{app:error}).

\begin{restatable}[the Rayleigh range]{theorem}{thmrange}
\label{thm:range}
To first order in $\Delta$:
\begin{enumerate}
  \item[(i)] a rank-one interchange along a unit $d$ has mean effect $\E[e(d)] = d^\top M d$, which takes every value in
             $[\lambda_D, \lambda_1]$ as $d$ ranges over unit vectors;
  \item[(ii)] whole-state patching at the location has mean effect $\operatorname{tr} M = \sum_j \lambda_j$, with
             an error of third order in $\Delta$;
  \item[(iii)] a rank-$k$ interchange along $P$ has mean effect $\operatorname{tr}(P^\top M P)$, which lies between the
             sum of the $k$ smallest and the sum of the $k$ largest eigenvalues;
  \item[(iv)] for a single pair, $\lambda_1 = \tfrac12\big(w^\top\Delta + \|w\|\|\Delta\|\big) \ge w^\top\Delta$, with equality
             if and only if $w$ and $\Delta$ are parallel and $w^\top\Delta \ge 0$.
\end{enumerate}
\end{restatable}

\begin{figure}[p]
  \centering
  \includegraphics[width=\textwidth,height=\dimexpr\textheight-80pt\relax,keepaspectratio]{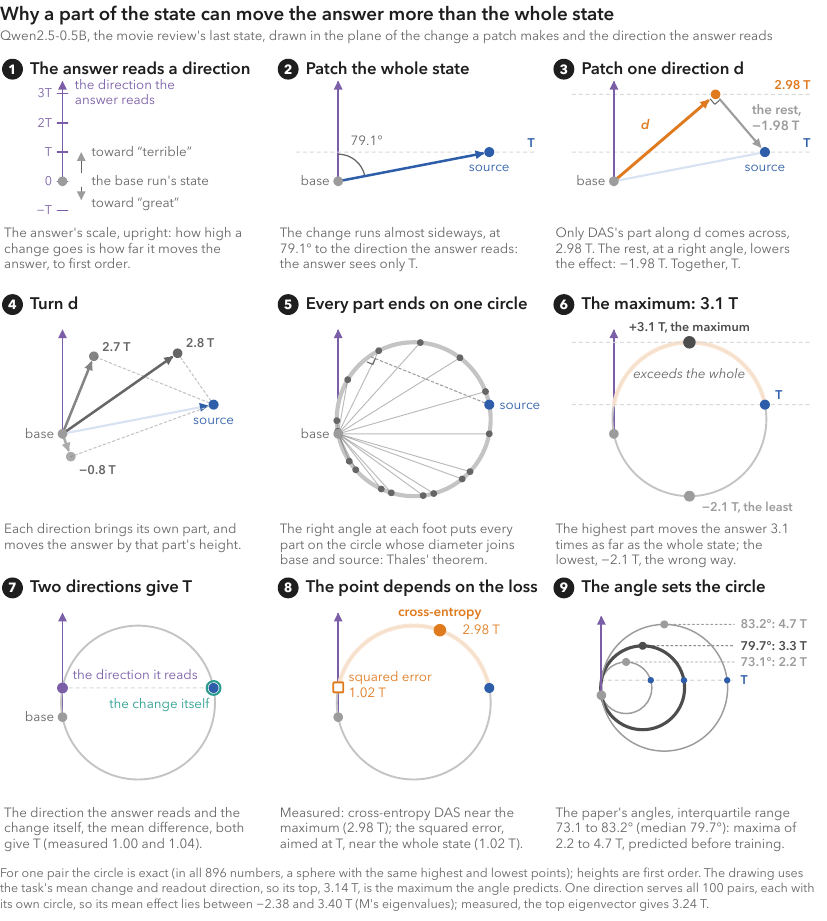}
  \caption{\textbf{A part of the state can change the answer more than the whole, because the rest of the change lowers the
  effect.} \Cref{fig:guidea}'s review at the readout, in the plane of the mean change and the readout direction (angle
  $\theta$, \cref{prop:natural}). (1)~\Cref{fig:guidea}'s scale of the answer, stood upright. In (3) and (8), measured values are drawn
  at their heights. (9)~The interquartile range of
  $\theta$ over the two-label cases of the \roModelsFOWord{} uncapped models.}
  \label{fig:guidec}
\end{figure}

\begin{proof}
For one pair, a subspace patch $h_b \mapsto h_b + v$, $v = PP^\top\Delta$, keeps the state in the ball with diameter
$[h_b, h_s]$, on its boundary since $v^\top(\Delta - v) = 0$ (Thales' theorem), and changes $z$ by $w^\top v + \tfrac12 v^\top H(v-\Delta)$ up to third order, with $w$ and the Hessian $H$ taken at the midpoint; to first order the change is $w^\top v$.

(i) The rank-one patch is $v = dd^\top\Delta$, so $e(d) = (w^\top d)(d^\top\Delta) = d^\top w\Delta^\top d$ to first order. A
quadratic form sees only the symmetric part of its matrix, so $e(d) = d^\top\operatorname{sym}(w\Delta^\top)\,d$, and
averaging over pairs gives $\E[e(d)] = d^\top M d$. On the unit sphere this Rayleigh quotient lies in $[\lambda_D, \lambda_1]$
and attains both ends at unit eigenvectors. The sphere is connected for $D \ge 2$ and the quotient is continuous, so it takes
every value in between.

(ii) Whole-state patching is the patch $v = \Delta$. Its first-order effect is
$w^\top\Delta = \operatorname{tr}(w\Delta^\top)$, whose mean is $\operatorname{tr}\E[w\Delta^\top] = \operatorname{tr}M$,
since a matrix and its transpose, and so its symmetric part, have the same trace. The second-order term
$\tfrac12 v^\top H(v-\Delta)$ vanishes at $v = \Delta$, so the error is of third order.

(iii) The patch is $v = PP^\top\Delta$, with first-order effect $(P^\top w)^\top(P^\top\Delta) =
\operatorname{tr}\big(P^\top\Delta w^\top P\big) = \operatorname{tr}\big(P^\top\operatorname{sym}(w\Delta^\top)P\big)$, whose
mean is $\operatorname{tr}(P^\top M P)$. For a symmetric $M$ and $P$ with $k$ orthonormal columns, Ky Fan's maximum principle
bounds $\operatorname{tr}(P^\top MP)$ above by the sum of the $k$ largest eigenvalues; applied to $-M$, it bounds it below by
the sum of the $k$ smallest.

(iv) For one pair, $M = \tfrac12(w\Delta^\top + \Delta w^\top)$ maps every vector into $\operatorname{span}(w, \Delta)$, so
all its eigenvalues but two, $\mu_+ \ge \mu_-$, are zero. Their sum is $\operatorname{tr}M = w^\top\Delta$, and the sum of their
squares is $\|M\|_F^2 = \tfrac12\big((w^\top\Delta)^2 + \|w\|^2\|\Delta\|^2\big)$, so their product is
$\tfrac14\big((w^\top\Delta)^2 - \|w\|^2\|\Delta\|^2\big)$ and $\mu_\pm = \tfrac12(w^\top\Delta \pm \|w\|\|\Delta\|)$. By
Cauchy--Schwarz $\|w\|\|\Delta\| \ge |w^\top\Delta|$, so $\mu_+ \ge 0$ is the largest eigenvalue, $\lambda_1$, and
$\lambda_1 - w^\top\Delta = \tfrac12(\|w\|\|\Delta\| - w^\top\Delta) \ge 0$, with equality if and only if $w$ and $\Delta$ are
parallel and $w^\top\Delta \ge 0$.
\end{proof}
\Cref{thm:range} places every rank-one method on a single scale. In any orthonormal basis, the rank-one effects sum to
$\operatorname{tr}M$, the effect of whole-state patching, so one of them exceeds it exactly when the others sum to a negative
number. For the top eigenvector of $M$, the eigenvector of $\lambda_1$, the others sum to $\sum_{j\ge2}\lambda_j$, which by (iv) is negative for a single pair
unless $w$ and $\Delta$ point in the same direction (\cref{fig:law}b and \cref{fig:guidec}).

An effect above that of whole-state patching does not mean that a direction isolates the variable more cleanly, as
the following two-dimensional example shows.

\begin{example}[two dimensions]
\label{ex:two}
Let the change of the state be $\Delta = (1, 0)$, along the variable, and let the gradient of $z$ at the site be $w = (1, 1)$, at $45^\circ$ to it. Then $M = \big(\begin{smallmatrix} 1 & 1/2 \\ 1/2 & 0 \end{smallmatrix}\big)$. Its trace, 1, is the
effect of whole-state patching, and its eigenvalues are $1.207$ and $-0.207$. The direction of the variable gives exactly 1.
The top eigenvector, which bisects $w$ and $\Delta$, gives more, $1.207$, because it leaves out the orthogonal direction,
whose effect is $-0.207$. The bisector moves the state by $0.354$ along the second coordinate, which the input does not
change at all, and the orthogonal direction moves it back by the same amount. The negative contribution of the orthogonal direction is thus not a separate mechanism, because it is negative only through
moving the state back along the second coordinate. If the gradient is turned to $w = (0, 1)$, whole-state patching gives 0, while the bisector of the new $w$ and $\Delta$ gives
$1/2$. This is the form of the illusion described by \citet{makelov2023}. In this illusion, a direction combines a
\emph{disconnected} part, which the input changes but the output does not read, with a \emph{dormant} part, which the output
reads but the input does not change.
\end{example}

With both gradients, the excess is an effect that whole-state patching does not produce, although it transfers the
entire change.
In general, whenever the effect of whole-state patching is positive, a part can exceed it only through its component orthogonal to
$\Delta$ (\cref{app:lean}).

When a site has several locations that are patched along the same $d$ (for example every position of a layer), the patch adds $dd^\top\Delta_t$ at each location $t$. To first order,
$e(d) \approx \sum_t (d^\top w_t)(d^\top\Delta_t)$, so (i)--(iii) and \cref{cor:select} hold with
$M = \operatorname{sym}\E\big[\sum_t w_t\Delta_t^\top\big]$, whose trace is the first-order effect of whole-state patching at
the site.

\subsection{Directions Selected by Training Objectives}
\label{app:proofs-select}

\begin{restatable}[selection; first order]{corollary}{corselect}
\label{cor:select}
\leavevmode   %
\begin{enumerate}
  \item[(i)] An effect-seeking objective in pure form, strictly decreasing in the mean effect, selects a unit vector of the top
             eigenspace of $M$ and reports $\lambda_1$. Its excess over whole-state patching is
             $-\sum_{j\ge 2}\lambda_j$. The excess is positive for a single pair unless $w$ and $\Delta$ are positively parallel
             (\cref{thm:range}(iv)), and, whenever $w_i = c_i\bar w$ for scalars $c_i$, unless $\bar w$ and $\E[c\Delta]$ are positively parallel.
  \item[(ii)] An objective that depends on $d$ only through the mean effect and is uniquely minimised at a target $\tau$
             (a target-aimed objective in pure form) selects the level set $\{d : d^\top M d = \tau\}$ when it is non-empty, which holds
             if and only if $\lambda_D \le \tau \le \lambda_1$.
  \item[(iii)] At the readout, $\operatorname{tr}M = T$ up to an error of third order (\cref{prop:cut}, \cref{thm:range}(ii)). An effect-seeking
             objective therefore departs from $T$ by $-\sum_{j\ge2}\lambda_j$, and a target-aimed one with $\tau = T$ reaches $T$
             when $T \in [\lambda_D, \lambda_1]$. This condition holds under one shared direction (\cref{prop:natural}), but
             not always (\cref{app:counter}).
\end{enumerate}
\end{restatable}

\begin{proof}
(i) By \cref{thm:range}(i) the mean effect of a unit $d$ is $d^\top Md \le \lambda_1$, with equality exactly on the unit vectors
of the top eigenspace. An objective strictly decreasing in the mean effect is therefore minimised exactly there, and the
direction it selects reports $\lambda_1$. Whole-state patching has mean effect $\operatorname{tr}M$
(\cref{thm:range}(ii)), so the excess is $\lambda_1 - \operatorname{tr}M = -\sum_{j\ge2}\lambda_j$. For a single pair this is
$\lambda_1 - w^\top\Delta$, positive unless $w$ and $\Delta$ are positively parallel (\cref{thm:range}(iv)). For $w_i = c_i\bar w$
and $m = \E[c\Delta]$, $\E[w\Delta^\top] = \bar w m^\top$, so $M = \operatorname{sym}(\bar w m^\top)$ is the single-pair matrix
with $\bar w$ and $m$ in place of $w$ and $\Delta$, and the same argument gives an excess of
$\tfrac12(\|\bar w\|\|m\| - \bar w^\top m)$, positive unless $\bar w$ and $m$ are positively parallel.

(ii) By \cref{thm:range}(i), unit vectors $d$ with $d^\top Md = \tau$ exist if and only if $\lambda_D \le \tau \le \lambda_1$,
and when they do, such an objective is minimised exactly at them.

(iii) The readout is a complete cut, so whole-state patching there gives $T$ exactly (\cref{prop:cut}) and
$\operatorname{tr}M$ up to an error of third order (\cref{thm:range}(ii)). Hence $\operatorname{tr}M = T$ up to that error,
and (i) and (ii) with $\tau = T$ give the rest. Under one shared direction, $M = \operatorname{sym}(\bar w m^\top)$ has, as in
the proof of \cref{thm:range}(iv), eigenvalues $\mu_+ \ge 0 \ge \mu_-$ besides zeros, so
$\lambda_D \le \mu_- \le \operatorname{tr}M \le \mu_+ = \lambda_1$: the trace, and with it $T$, lies in the range.
\end{proof}

Outside these cases, the excess in (i) can be negative (\cref{app:counter}). Our measurements at the readout include such a case, indirect object identification (IOI) with twenty names, where even the maximum effect lies below $T$ in every model
(\cref{app:rankkmeasured}).

Cross-entropy of the source label, unlike the objectives of \cref{cor:select}, is not a function of the mean effect alone.
However, until it saturates, it rewards a larger log probability of the source label, and with it larger log odds. At the readout it therefore behaves almost like an
effect-seeking objective in pure form, and the linear logit difference of \citet{hollinsworth2024} is exactly such an
objective. The behaviour of cross-entropy differs in one respect. Its
derivative with respect to the logit of the base label equals that label's probability $p_b$. When the base label is improbable, cross-entropy gives
little reward for lowering it, and the learned direction stays below the maximum effect (\cref{sec:shortfall}).

Part (ii) requires an objective that is a function of the mean effect alone. The CDAS divergence to the next-token distribution of the
source run \citep{bao2026} and a per-pair squared error on $z$ are not, and each can end up away from $T$
(\cref{app:counter}). When every pair shares one readout direction, however, the squared error provably reaches $T$ to
first order (\cref{app:lean}).

\subsection{One Shared Readout Direction}
\label{app:proofs-natural}

\begin{restatable}[one shared direction]{proposition}{propnatural}
\label{prop:natural}
Suppose $\E[w\Delta^\top] = \bar w m^\top$ for vectors $\bar w, m$: for instance a shared readout direction, $w_i = c_i\bar w$
with $m = \E[c\,\Delta]$, or a single code, in which every pair changes the state along one direction, $\Delta_i = s_i\bar\Delta$, with
$\bar w = \E[s\,w]$ and $m = \bar\Delta$. Let
$\theta = \angle(\bar w, m) \in [0^\circ, 90^\circ)$, $\hat w = \bar w/\|\bar w\|$, $\hat m = m/\|m\|$, and write $e(d)$ for the mean
effect. To first order:
\begin{enumerate}
  \item[(i)] $e(d) = (d^\top\bar w)(d^\top m)$ and $\operatorname{tr} M = \bar w^\top m$;
  \item[(ii)] the readout direction and the direction of the mean change of the state both give the trace:
             $e(\hat w) = e(\hat m) = \operatorname{tr}M$;
  \item[(iii)] for the unit $d$ at angle $\varphi$ from $\hat m$ toward $\hat w$ in their span,
  \begin{equation}
  \begin{aligned}
    \frac{e(d)}{\operatorname{tr}M} &= 1 + \frac{\sin\varphi\,\sin(\theta-\varphi)}{\cos\theta},\\
    \frac{\lambda_1}{\operatorname{tr}M} &= \frac12\Big(1 + \frac{1}{\cos\theta}\Big),
  \end{aligned}
    \label{eq:angle}
  \end{equation}
  so every direction strictly between $\hat m$ and $\hat w$ exceeds the trace, and the maximum is attained at the bisector
  $\varphi = \theta/2$, uniquely up to sign.
\end{enumerate}
\end{restatable}

\begin{proof}
(i) $M = \operatorname{sym}(\bar w m^\top)$, and a quadratic form sees only the symmetric part, so
$e(d) = d^\top\bar w m^\top d = (d^\top\bar w)(d^\top m)$; the trace of $\bar w m^\top$ is $\bar w^\top m$.

(ii) For $d = \hat w$, $e(\hat w) = \|\bar w\|\,\hat w^\top m = \bar w^\top m$, and likewise $e(\hat m) = \bar w^\top m$.

(iii) For $d$ in the span, $d^\top \bar w = \|\bar w\|\cos(\theta-\varphi)$ and $d^\top m = \|m\|\cos\varphi$. Since
$\theta < 90^\circ$, the trace $\operatorname{tr}M = \|\bar w\|\|m\|\cos\theta$ is positive, so
$e(d)/\operatorname{tr}M = \cos\varphi\cos(\theta-\varphi)/\cos\theta$, and expanding $\cos\theta = \cos(\varphi + (\theta-\varphi))$
gives $\cos\varphi\cos(\theta-\varphi) - \cos\theta = \sin\varphi\sin(\theta-\varphi)$, the first line of \eqref{eq:angle}.
For $0 < \varphi < \theta$ both sines are positive, so every direction strictly between $\hat m$ and $\hat w$ exceeds the trace.
Since $\cos\varphi\cos(\theta-\varphi) = \tfrac12[\cos\theta + \cos(2\varphi-\theta)]$, the in-span maximum is
$\tfrac12\|\bar w\|\|m\|(1+\cos\theta)$, attained only at $\varphi = \theta/2$ up to sign, and its ratio to the trace is the
second line. A unit $d = d_\parallel + d_\perp$ with $d_\perp$ orthogonal to the span has
$e(d) = (d_\parallel^\top\bar w)(d_\parallel^\top m) = \|d_\parallel\|^2\, e(d_\parallel/\|d_\parallel\|)$ (or $0$ if $d_\parallel = 0$),
which is below the in-span maximum unless $d_\perp = 0$, because that maximum is positive. So the maximum over all unit $d$
is the in-span one, which is $\lambda_1$.
\end{proof}

\Cref{fig:guidec} illustrates this result. Consider a mirrored test set, in which every pair appears with its mirror, and in which the class fixes each pair's label, as
in CausalGym. When the change of the state is a single code, a difference of means and a heavily penalised probe return
$\pm\hat m$, and the readout gradient returns $\hat w$. Both directions give the trace (\cref{prop:natural}(ii)), and a
direction above the trace must combine $\hat m$ and $\hat w$. Only an objective that
rewards an effect above the trace favours such a combination, and \cref{eq:angle} predicts how far a method with such an
objective goes. PCA and $k$-means use the states alone and have no such objective. A larger rank does not change this picture. Under one shared direction, every rank below the dimension has, to first order,
the same predicted maximum as rank one, so an effect-seeking objective in pure form reports the same excess at every such
rank (\cref{app:rankk}).

\section{Experimental Setup}
\label{sec:experiments}

We measure every method at the readout and at every cell of a map inside the network. This section describes the tasks,
models and objectives, the intervention sites, the measured quantities and the evaluation, and \cref{app:details} gives full
details of the data, the models, the training and the compute.

\subsection{Tasks, Models and Objectives}
\label{sec:tasks}

The \roDistinctTasks{} tasks come from \roSourcesWord{} published sources. They include the 29 tasks of CausalGym
\citep{arora2024} and re-paired versions of its six number-agreement tasks, in which only the number of the subject changes
(\emph{number-only agreement}). The other tasks are the constructions of \citet{kumon2026a}, sentiment
templates \citep{hollinsworth2024}, SST sentences \citep{socher2013}, RAVEL \citep{huang2024}, and indirect object
identification (IOI) \citep{wang2022}, both in its original templates and in a sentence frame of our own. The
\roModelsWord{} models range from 124M to 32B parameters and come from six families
\citep{radford2019,biderman2023,allal2025,qwen2024,dubey2025,team2024}. Pythia, Qwen2.5 and Llama are each included at three
or more sizes, so that size can be compared within a family. We call the six smallest models
(124M--1B) the \emph{small} models and the four models of 7--9B the \emph{large} models.

In each model and task, we train rank-one DAS at the readout with four objectives, each from the same three random
initialisations. The objectives are cross-entropy of the source label, a per-pair squared error to each pair's total effect,
the CDAS divergence \citep{bao2026}, and the mean effect itself, which we call the mean-effect objective. Training runs for ten passes over the training pairs,
and we keep each direction after the first and after the tenth pass. Counts refer to the directions after one pass unless stated
otherwise. We score each direction by its exact effect on the test pairs. Inside the network, we add a second per-pair squared error, aimed at the cell's $G$. There, this squared error is trained
from three initialisations, like cross-entropy, and the other objectives from one.

Each task in a model is a \emph{case}. The \emph{maximum effect} of a case is the largest mean effect that we find for a
single direction. We search for this direction by gradient ascent on the unit sphere, using the training pairs, and measure
its effect on the test pairs. The maximum effect is thus an out-of-sample estimate, the \emph{out-of-sample maximum}, because the search does not use the
test pairs. The largest exact (not
first-order) effect that a separate search finds on the test pairs themselves is the \emph{in-sample maximum}, the
counterpart of $\lambda_1$ on the same pairs. The \emph{predicted maximum} is $\lambda_1$ (the largest eigenvalue of $M$), the largest first-order effect of a single
direction and the upper end of the range. Where all pairs share one readout direction, the angle of \cref{prop:natural},
measured on the training pairs, estimates the same quantity. The
\emph{margin above $T$} is the maximum effect minus $T$, and a case has a margin above $T$ when the interval of its maximum
effect lies above $T$. We keep the \roTasks{} cases that have at least 20 test clusters (a cluster is a pair together with its
mirror) and whose interval for $T$ lies above zero. We call this the \emph{population rule}. A case counts as above or below a value only when its
95\% interval does, and intervals resample clusters.

\subsection{Intervention Sites}
\label{app:sites}

Patching a cell replaces the state at the last token of its region (CausalGym's
\emph{last} strategy), and the
region of the start-of-text token is never patched. The patch copies the source run's state at the last token of the same
region.

The maps of \cref{sec:offroute} cover every region at every second layer ($0, 2, 4, \dots$) and the readout cell, in each
of the six small models. For CausalGym's 29 tasks, the maps of \cref{sec:published} also cover the odd layers, and there
cross-entropy DAS is trained under CausalGym's protocol.

\subsection{Quantities Measured at the Readout}
\label{app:quantities}

Throughout, $M$, $\lambda_1$, $\operatorname{tr}M$, $\hat b$ and $v_1$, the top eigenvector of $M$, use gradients at the
midpoint of each pair's patch,
as in \eqref{eq:M}. Versions with gradients at the base point are used to check the trace (\cref{prop:trapezoid}). We compute $\lambda_1$ exactly in the span of the pairs' gradients and differences. $\lambda_1$ and
$\operatorname{tr}M$ come from the test pairs, and $v_1$ and the angle $\theta$ of \cref{fig:readout}b come from the training
pairs.

The angle $\theta$ lies between two directions. The first, $\hat u$, is the mean change of the state over the pairs.
Each pair's change is multiplied by $+1$ or $-1$ according to the pair's class, and the class fixes the pair's label. On IOI,
where the class does not fix the label, the sign follows the label instead. The second, $\hat b$, is the mean readout
gradient, signed in the same way. This is the angle of \cref{prop:natural} when the change of the state is a single code.

\subsection{Evaluation and Statistical Tests}
\label{app:evaluation}

An effect is the mean log odds ratio over a task's test pairs. Frozen
attention replaces the attention probabilities of every head and query row, in every block from the site's block onward, by those of the unpatched base run in the same batch. These probabilities are the attention weights. The part of an effect
that freezing removes is the part that runs through the attention weights, which we also call, for short, the part that runs
through attention. Intervals are 95\% bootstrap intervals that resample test clusters, unless stated otherwise.

\section{Results at the Readout}
\label{sec:inflation}

We begin where the reference is known exactly. At the readout, whole-state patching moves the output by exactly $T$, so
the position of every method in the range can be read directly against it. This section tests four predictions of the theory
there. Cross-entropy should exceed $T$ wherever some direction can, the errors of two first-order estimates of $T$ should
stand in a fixed relation, a single angle should predict the maximum effect before training, and cross-entropy should fall
short of that maximum where the base label is improbable. \Cref{app:extended} gives the details of the measurement and further tests.

\begin{figure}[t]
  \centering
  \includegraphics[width=\textwidth]{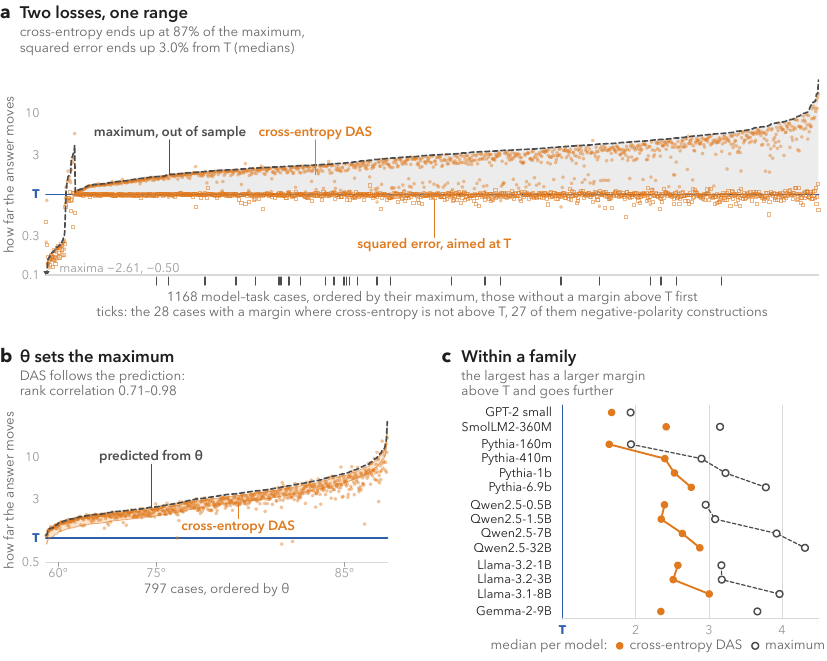}
  \caption{\textbf{At the readout, the loss determines where DAS ends up, and one angle predicts the maximum effect.} Exact effects
  on test pairs. (a)~Every case in all \roModelsWord{} models. Grey shows the margin above $T$, up to the maximum. A case has this margin when
  the interval of its maximum lies above $T$; the cases without it come first, and some of them have maxima above $T$ with
  intervals that reach $T$. The
  ticked cases are discussed in \cref{sec:shortfall}. (b)~The two-label cases of the \roModelsFOWord{} uncapped models,
  except one case whose DAS effect is not above zero. The prediction comes from \cref{prop:natural}, and the band shows
  the interquartile range of DAS as a share of it. The rank correlation is Spearman's, per model. (c)~Lines join a family's sizes and are not a test. The test of size compares ratios per task (\cref{tab:scale}).}
  \label{fig:readout}
\end{figure}

\subsection{The Effect of the Training Loss}
\label{sec:landing}

Changing only the loss moves DAS from far above $T$, with cross-entropy, to close to $T$, with the per-pair squared error or
the CDAS divergence. With cross-entropy, DAS ends up above $T$ in \roCeAbove{} of the
\roRoom{} cases with a margin above $T$ (\roCeAboveAll{} of all \roTasks{} cases; \cref{fig:readout}a), at a median of \roCeMedian\,$T$ per model, or a median of \roCeOfTop{} of the maximum effect per model. After ten passes,
it lies in much the same place, close to the optimum of its objective, so its position after one pass is mostly not due to
unfinished training (\cref{app:landing}). In
contrast, from the same initialisations, the per-pair squared error ends up a median of \roSqMedianDev\% from $T$ (farther on the sentiment
templates, which train for \roSentSteps{} steps per pass against \roOtherSteps{} elsewhere and come to \roSqSentTen\% after
ten passes, and where the maximum effect itself lies below $T$), and the CDAS divergence at a median of \roJsdMedian\,$T$ per model.
Interchange intervention accuracy follows the loss as well (\cref{sec:accuracy}). Thus, from the same initial directions,
the loss determines the effect of the trained direction.

\paragraph{A confound.} A second feature that changes together with the target does not explain the excess. Such a feature would move the output
toward the base label under whole-state patching and so lower $G$. CausalGym's six number-agreement tasks change the lemma of
the subject together with its number (\emph{senators} against \emph{minister}). On pairs built from the same rows that change
only the number, cross-entropy DAS ends up at a median of \noCe\,$T$ per model, against \noCeParent\,$T$ on CausalGym's pairs
of the same tasks, and its interval lies above $T$ in \noAbove{} of the \noTasks{} model--task cases. Changing only the lemma
moves the output by \noLemma\,$T$ on average. The excess of DAS therefore does not come from a lemma effect that lowers $G$.
These pairs, however, cannot separate number from a feature of the plural form itself.

\paragraph{Held-out evaluation.} Testing on held-out pairs removes the part of an effect that is specific to the training
entities, but not a shift that every entity shares, so it does not remove the excess either. On IOI's held-out templates, cross-entropy DAS lies
above $T$ in all \hoIoiN{} cases of the task with two names, and on RAVEL's test prompts in \hoRavelAbove{} of the
\hoRavelRoom{} cases with a margin above $T$. Where the test pairs differ from the training pairs in construction or
template, the margin above $T$ itself often disappears, and the excess with it (\cref{app:safeguards}).

\subsection{Accuracy of the First-Order Approximation}
\label{sec:tests}

\begin{figure}[tb]   %
  \centering
  \includegraphics[width=0.6\textwidth]{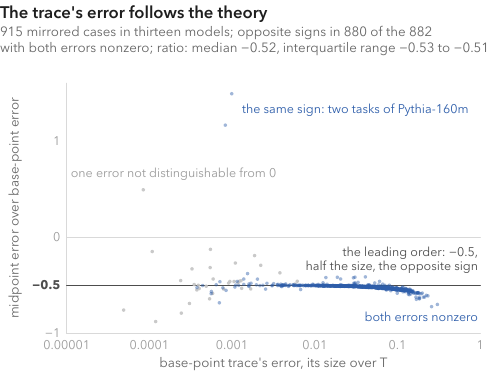}
  \caption{\textbf{The error of the trace has the form that the theory predicts.} Every mirrored task of the main population in the
  \roModelsFOWord{} uncapped models, with errors divided by $T$ on test pairs. The line shows the leading order
  (\cref{prop:trapezoid}), at which the midpoint error is $-\tfrac12$ times the base-point error. At the largest errors, the ratio falls a little
  below $-\tfrac12$, beyond the leading order. Least-squares slopes
  of the signed errors through 0 range from \roSlopeHigh{} to \roSlopeLow{} across models.}
  \label{fig:trace}
\end{figure}

The first-order approximation describes the readout closely. The trace of $M$, which is the first-order value of the effect of
whole-state patching, lies a median of \roTraceDevMid\% from $T$ per model. Its error also has the form that the theory predicts. On
mirrored test sets, computing $M$ at the midpoint of the patch rather than at the base point should, to leading order, halve
the error and reverse its sign (\cref{prop:trapezoid}). The two errors have opposite signs in \roOppositeNum{} of the
\roOppositeDen{} cases where both are nonzero (\cref{fig:trace}); the two exceptions are mirrored tasks of Pythia-160m. On
sets that are not mirrored, the midpoint trace should be the more accurate one, but in one task of each Llama model the
base-point trace is significantly closer to $T$. The main exception to the first-order approximation, left out
of these counts, is Gemma-2-9B, which passes its final logits through a soft cap $\kappa\tanh(o/\kappa)$. In this model the
approximation is less accurate, and the in-sample maximum lies above the predicted maximum in \gmTopAboveLam{} of its \gmTasks{} tasks, and in none once
the cap is removed. Removing the cap accounts for most, but not all, of the difference between Gemma-2 and the other
families (\cref{app:cap}). The checks of the trace and of the angle
therefore use the other \roModelsFOWord{} models. We made this split after seeing the results for Gemma-2.

\subsection{Predicting the Maximum Effect Before Training}
\label{sec:angle}

In a \emph{two-label} task, every pair contrasts the same two answers. Where the logits are not capped, the readout gradients
then share almost exactly one direction, because only the final normalisation, whose Jacobian depends on the state, makes
them differ. A single angle therefore predicts the maximum effect before training. The angle $\theta$ between the readout direction and the mean change of the
state, computed from the training pairs, gives the maximum effect within a median of \roAngleTopErr\% per model (\cref{app:landing}). Since the
median angle is \roThetaMed$^\circ$ (interquartile range \roThetaQone--\roThetaQthree$^\circ$), most of the change of the state is orthogonal to the readout direction. An effect-seeking objective chooses a direction
between the change and the readout direction. For the mean change, patching along this direction moves the state further in the
readout direction than the whole change does (\cref{prop:natural}). Across
cases, the angle also ranks the effects that cross-entropy DAS reaches (rank correlation \roAngleCeSpearman{} per model;
\cref{fig:readout}b). DAS ends up close to the direction that the theory predicts, the top eigenvector of $M$ (median $|\cos|$
of \roCosVOne{} per model). The readout direction itself, in contrast, gives $T$, as predicted. The angle tells how far an effect-seeking method exceeds $T$ at the readout, not where the network computes the variable,
because it is measured at the readout, through which every effect on the output passes.

\subsection{Higher Ranks, Shortfalls and Model Size}
\label{sec:further}

\paragraph{Higher ranks.}
\label{app:rankkmeasured}
We also trained DAS at the readout at ranks 2 to 64, at a rank equal to the number of labels minus one for tasks with more
than two labels (the \emph{label rank}), and on the orthogonal complement of one trained direction (\cref{app:rankkfull} gives
the protocol and all counts). The excess is not limited to rank one. Every case above $T$ at rank one stays above $T$ at the
higher ranks wherever both have a margin above $T$, except for three cases that lose significance at one rank. Outside
Gemma-2, no trained run of cross-entropy DAS lies significantly above the predicted maximum for its rank $k$, the sum of the
$k$ largest eigenvalues of $M$. With more than two labels, a larger rank can enlarge the margin above $T$. On IOI with twenty
names, the maximum effect lies below $T$ at rank one in every model, but from rank 16 onward DAS lies above $T$ in every
case, so in this task rank one understates the excess. On the other tasks with more than two labels, the growth of
the predicted maximum with the rank does not carry over from the training pairs to the test pairs.

\paragraph{Where cross-entropy does not exceed $T$.}
\label{sec:shortfall}
Of the \roCeMiss{} cases with a margin above $T$ in which cross-entropy is not above $T$, one is the sentiment classification
task of Pythia-160m, where after one pass the run from one initialisation lags far behind the other two
(\cref{app:shortfallfull}). The other \roCeMissNpi{} belong to the \npiCases{} cases of the negative-polarity constructions of
\citet{kumon2026a}, in which a word such as \emph{ever} needs a licensor such as \emph{no}. In the one-directional form that
the authors build, every base label is \emph{ever}, to which the models give a median probability of \npiPb{} per model. The
derivative of cross-entropy with respect to the base label's logit equals that label's probability, which is nearly zero here. Lowering \emph{ever} accounts for \npiTopShare{} of the maximum effect
in log odds, but its share of the gain of cross-entropy DAS is lower by \npiCeMinusTop{} in absolute terms
(\cref{fig:npidecomp}). Cross-entropy DAS therefore stays below the maximum effect on these constructions, and it does not
exceed $T$ where the margin above $T$ is smaller.

\paragraph{Model size.}
\label{app:scale}
Within a family, the largest model has a larger margin above $T$, and the excess of DAS grows with it. Measured as multiples
of $T$, the maximum effect grows by a median of \ssPythiaRho--\ssLlamaRho\% per task from the smallest to the largest of three
sizes in Pythia, Qwen2.5 and Llama, and cross-entropy DAS grows by \ssPythiaR--\ssLlamaR\%. From Qwen2.5-7B to Qwen2.5-32B,
both quantities still grow, by \ssThirtyRho\% and \ssThirtyR\% (\cref{tab:scale}). The predicted maximum, $\lambda_1/T$, which
is computed before training, grows with the maximum effect at every step in all three families, so the growth is a
property of the network, not of the search that finds it. These statements rest on a test within families, whose smallest step
holds only with CausalGym's tasks (\cref{app:scalefull} gives its design, intervals and limits).

\FloatBarrier   %

\section{Results Inside the Network}
\label{sec:offroute}

Inside the network, the reference is no longer known in advance. Most sites there are partial cuts, so whole-state patching
at a site moves the output by its own amount $G$, which has to be measured. This section compares learned effects with $G$
at every cell of a map. It shows that DAS exceeds whole-state patching at most cells and also changes the output where
whole-state patching shows no significant effect, that much of this excess runs through attention, and that, as at the readout, the
target of the loss decides where a learned effect ends up.

\begin{figure}[tbp]
  \centering
  \includegraphics[width=\textwidth]{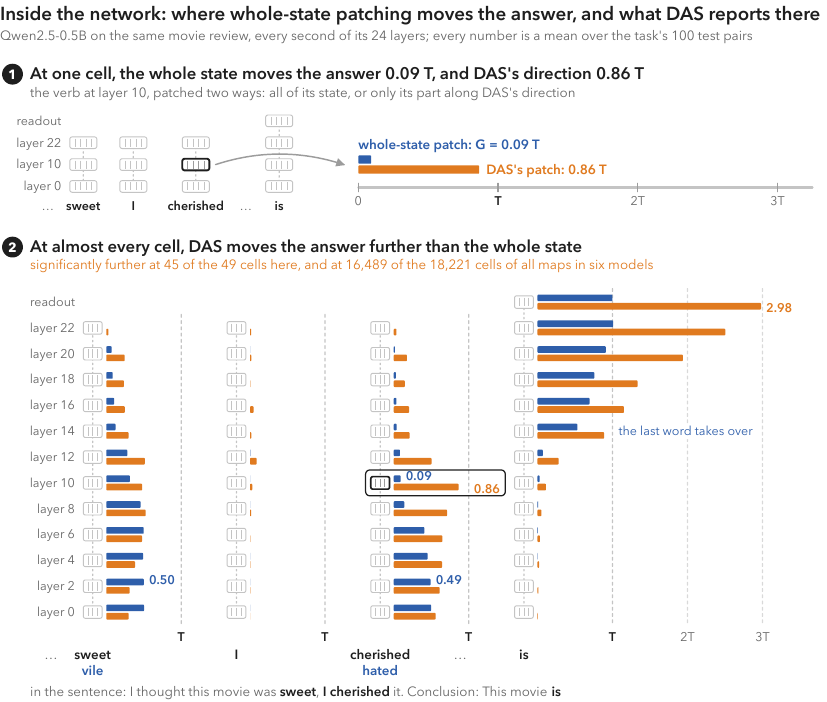}
  \caption{\textbf{The network moves the sentiment from two words to the last word, and at most cells cross-entropy DAS
  moves the answer further than whole-state patching of the cell.} $G$ is the effect of whole-state patching at the cell,
  and DAS is the mean of its three trained runs. The words before \emph{sweet} are not drawn, because the
  change of the input does not reach them. Counts use false-discovery control over the cells of each model (\cref{app:fullgrid}).}
  \label{fig:guideb}
\end{figure}

The maps cover \inTasks{} tasks (CausalGym's 29 and 49 from the other sources) in the six small models, at every region and
every second layer, with the effect of cross-entropy DAS averaged over its three trained runs; \inMaps{} of them meet the
population rule. \Cref{fig:guideb} shows the map of the running example. Unless stated otherwise, DAS in this section is
trained with cross-entropy.

For a reference that is harder to exceed, we also measure $G^*$, the effect of whole-state patching in the cell's \emph{best
context}. This is the effect of patching the cell alone, or the effect that the cell adds when it is patched together with
one other region of its layer (any region but the start-of-text region), whichever is larger on the training pairs. The cells
at which the interval of $G^*$ lies above zero form the \emph{route} of the map. \Cref{alg:route} gives the test of a cell
against $G^*$, which we call the \emph{route test}. Its branches show two ways in which a learned effect $e$ can exceed what
the network produces at a cell. Where the interval of $G^*$ lies above zero, $e$ can lie above $G^*$. Where it does not, $e$
can still change the output. A cell is \emph{off-route} for a direction when the interval of its effect $e$ lies above zero,
the interval of $G^*$ does not, and $e$ lies significantly above $G^*$. At such a cell, whole-state patching in its best context shows no significant positive effect. We call $e$ at such a cell an \emph{off-route effect}. In either case,
$e$ is inflated where it lies above $G$. Unless stated otherwise, off-route cells are those of
cross-entropy DAS.

\begin{algorithm}[tbp]
  \caption{Route test for one cell (layer $\ell$, region $r$).}
  \label{alg:route}
  \begin{algorithmic}[1]
    \STATE \textbf{Input:} model, training and test pairs, a direction $d$ (any method), $T$ from two plain forward passes per pair
    \STATE $e \gets$ effect of the rank-one interchange along $d$ at $(\ell, r)$
    \STATE $c \gets$ the context, none or one other region $r'$ of layer $\ell$, with the largest $G(\{(\ell,r)\})$ or
      $G(\{(\ell,r),(\ell,r')\}) - G(\{(\ell,r')\})$ on the training pairs
    \STATE $G^* \gets$ the same quantity in context $c$ on the test pairs, that is, the effect of whole-state patching in its best context
    \STATE $e_{\mathrm{frz}} \gets$ the effect along $d$ with every downstream attention pattern frozen to the base run's
    \STATE intervals for $e$, $G^*$, $e - G^*$, $e/G^*$ and $1 - e_{\mathrm{frz}}/e$, resampling test pairs with their mirrors
    \IF{$G^*$'s interval lies above 0}
      \STATE report $e/G^*$: an excess over $G^*$ where its interval lies above 1
    \ELSIF{$e$'s interval and that of $e - G^*$ lie above 0}
      \STATE \textbf{off-route}; report $e/T$, $G^*$'s upper bound and the share freezing removes, $1 - e_{\mathrm{frz}}/e$
    \ELSE
      \STATE no evidence that the network uses the cell; report $e$ and $G^*$ with their intervals
    \ENDIF
  \end{algorithmic}
\end{algorithm}

An off-route cell has the signature of the illusion described by \citet{makelov2023}, and its off-route effect has the
behavioural signature of what \citet{grant2025} call a pernicious divergence. A map can be read as showing where an
intervention succeeds or as showing the route of the network itself on these inputs. The route test separates these two
readings but does not choose between them.

\subsection{Learned Effects and Whole-State Patching Across Sites}
\label{sec:above}
\label{sec:large}

Cross-entropy DAS lies significantly above $G$ at \inCeAboveG{} of the \inLive{} cells that the change of the input reaches (the \emph{live} cells), by
a median of \inExcessMed\,$T$ at those cells. This holds for \inAboveSynRange\% of each source's cells on the syntactic tasks
(those of CausalGym, those of Kumon and Yanaka, and number-only agreement) and for \inAboveOther\% on RAVEL, IOI, SST and the
sentiment templates. Even against the best context, DAS lies significantly above $G^*$ at \inCeAboveGstar{} of the live
cells. Summed over the live cells, \inNpAll{} of the effect of DAS lies beyond $G^*$. This range covers every model and each of
three ways of summing (\cref{app:routefull}). The four large models, mapped on three of
these tasks, look much the same. We report this only as a description, because three maps per model are too few for a test
(\cref{app:largefull}). In the six small models, at the \inOffRoute{} off-route cells, where whole-state patching in its best context shows no significant positive effect, the effects of DAS are mostly small (a median of \inOffRouteDasMed\,$T$, and nine in ten below
\inOffRouteDasNinety\,$T$).

\paragraph{A stronger reference.} Choosing the second region of $G^*$ on the same test pairs on which $G^*$
is measured can only raise $G^*$. Even then, DAS lies above $G^*$ at \inCeAboveGstarTest{} of the live cells, against
\inCeAboveGstar{} with the region chosen on the training pairs, and where $G^*$ uses a second region, it exceeds $G$ by a
median of only \inCtxGainMed\,$T$.

\paragraph{A confound.} A second feature that cancels the target's effect within a cell
does not explain the off-route effects either, because the output receives only the net effect that whole-state patching
measures. At the off-route cells, the upper bound of the interval of $G^*$ has a median of \inOffRouteGstarHi\,$T$, and nine
in ten of these upper bounds lie below \inOffRouteGstarHiNinety\,$T$. The network itself therefore passes on to the output
little of what DAS isolates at these cells.

\paragraph{Prediction before training.}
\label{sec:ranked}
\label{sec:firstorder}
The route test does not rest on the first-order theory, because it compares two effects measured by forward passes at the same
cell. The theory explains why the first can exceed the second. The first-order excess at a cell, $(\lambda_1 -
\operatorname{tr}M)/T$, is the largest excess that a single direction can reach there to first order, and it is computed
from the gradients of the test pairs before any training. It ranks the cells by how much DAS exceeds $G$ only in part (mean
rank correlation \gridThRho{} over the maps; \cref{app:firstorderfull}). It does not bound the excess, because at \inCeilBH{} cells at least one of the three trained runs of DAS
exceeds the predicted maximum, mostly through attention. We report this as a description rather than a test.

\subsection{How Learned Directions Change Attention}
\label{sec:door}
\label{sec:route}

\begin{figure}[tbp]
  \centering
  \includegraphics[width=\textwidth]{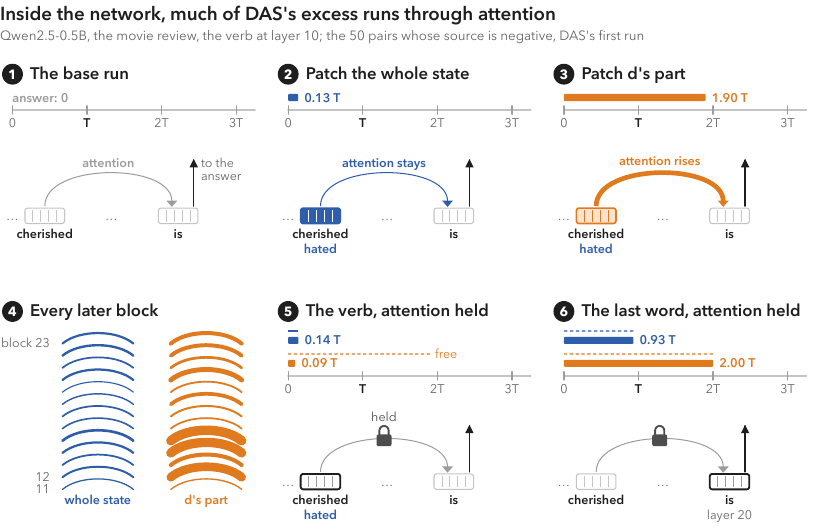}
  \caption{\textbf{Holding attention takes DAS's excess away at the verb, but not at the last word.} The beam in steps 1--3 is the
  attention of head 3 in block 12 from the last word to the verb; the arcs in step 4 are this attention summed over the 14 heads
  of each block that reads layer 10, and in the first of these blocks, which \cref{prop:door} describes, no head's attention rises
  by more than 0.02. All steps use the 50 test pairs whose source is negative and DAS's first run. On the other 50 pairs the same
  direction lowers this attention and moves the answer 0.03\,$T$; over all 100 pairs and its three runs, DAS gives the 0.86\,$T$
  of \cref{fig:guideb}. Held means that attention is frozen, with every attention pattern after the patched cell
  kept at the base run's; in steps 5 and 6 the beam and its lock stand for all of them. The dashed lines are the same patches with attention free. At the last word the patch is at the position the answer is read from, so
  holding attention leaves its excess, which comes from the geometry of $d$'s part, as at the readout (\cref{fig:guidec}). Over all
  \inLive{} live cells of six models, holding attention removes \inExcessAttnAll{} of DAS's summed excess over $G$ and keeps
  \inOwnKeep{} of the summed $G$.}
  \label{fig:guided}
\end{figure}

The state at a site also affects how much attention the last position pays to that site. To first order, in the first
block that reads the site, whole-state patching changes this attention only through the part of the change of the state that a
head reads. A learned direction, in contrast, can raise this attention even where the head reads none of the change, and
thus where whole-state patching leaves it as in the base run (\cref{prop:door}; \cref{fig:guided} shows such a rise in a
later block). Freezing the attention weights, as in path patching \citep{goldowskydill2023,hollinsworth2024},
separates every path through the attention weights, including the path through this rise, from the paths through the values
that the site passes on.

\begin{restatable}[raising attention that whole-state patching leaves low]{proposition}{propdoor}
\label{prop:door}
Take a site at position $p$ before the readout position $n$, and one head at $n$ in the first block that reads the
site's layer; its query $q$ and its scores on positions other than $p$ do not depend on the site. Its score on $p$ is $q^\top W_K\,\mathrm{LN}(h_p)$ (scaled). Let $a$ be the score's gradient in $h_p$ at $h_b$,
let $\alpha$ be the base run's attention weight on $p$, let $L = \log(1/\alpha)$, and write $\sigma$ for the logistic function.
To first order in the score:
\begin{enumerate}
  \item[(i)] whole-state patching changes the score by $a^\top\Delta$, and a rank-one interchange along $d$ by
             $(a^\top d)(d^\top\Delta)$, whose maximum over $d$ is $\tfrac12(a^\top\Delta + \|a\|\|\Delta\|)$, at the bisector of
             $\hat a$ and $\hat\Delta$;
  \item[(ii)] with the other scores and the values fixed, and for $t>0$: if $a^\top\Delta \le L - t$, whole-state
             patching leaves the weight on $p$ at most $\sigma\big(-t - \log(1-\alpha)\big)$; if
             $\tfrac12(a^\top\Delta + \|a\|\|\Delta\|) \ge L + t$, some rank-one interchange raises it to at least $\sigma(t)$. For a
             difference the head does not read ($a^\top\Delta = 0$) and $t \le L$, both conditions hold if and only if
             $\|a\|\|\Delta\| \ge 2(L + t)$.
\end{enumerate}
\end{restatable}
\Cref{app:door} gives the proof, the range in which the first-order statement holds and a test of where the attention to the
site rises.

Much of the excess of DAS above $G^*$ runs through attention, \inThroughAttnOfNotProduced{} per model (\cref{app:routefull}).
At the verb of the running example, holding attention takes nearly all of the excess away, and at the last word it takes
none of it away, because there the patch is at the position from which the answer is read (\cref{fig:guided}).
Whole-state patching, in contrast, keeps \inOwnKeep{} of its effect summed over the cells when attention is frozen, so
freezing does not remove every effect. Over the live cells, it removes \inCeAttnAll{} of the summed effect of DAS and
\inExcessAttnAll{} of its summed excess, $e - G$. Freezing only the attention of the readout position, from the block of the
site onward, removes \inRowExcess{} of that excess, so most of the excess that runs through attention runs through the
attention of the readout position. At cells with a smaller $G^*$, a larger share of the DAS effect tends to run
through attention. At the off-route cells, freezing removes \inOffRouteAttn{} of the summed effect of DAS. How much runs through attention depends on the task. Freezing removes more of the DAS effect on negative
polarity and filler--gap than on agreement or on the other sources.

\subsection{The Target of the Squared-Error Loss}
\label{sec:target}

\begin{figure}[tbp]
  \centering
  \includegraphics[width=\textwidth]{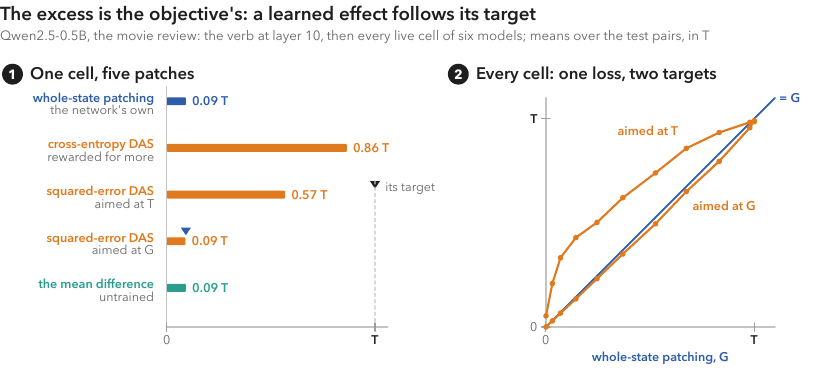}
  \caption{\textbf{The excess comes from the objective. Inside the network too, a learned effect follows the target of its
  objective.} In step 1, cross-entropy DAS and the squared error aimed at $G$ are means over three runs, and the squared error
  aimed at $T$ is one run; each triangle marks a target, $T$ or the cell's $G$. Step 2 sorts the \inLive{} live cells of six models
  into bins of $G$ and shows the median effect in each bin. Aimed at $T$, the squared error lies a median of \inSqTMinusG\,$T$
  above $G$ and meets $G$ as $G$ nears $T$; aimed at $G$, it lies near $G$ (slope \inSqGSlope{} and intercept \inSqGIntercept\,$T$, by least squares). Cross-entropy, another loss, is not drawn. It lies a median of
  \inCeMinusG\,$T$ above $G$, close to the squared error aimed at $T$ while $G$ is well below $T$, but past $T$ as $G$ nears
  $T$. The untrained mean difference is left out of step 2 because its comparison leaves out IOI, where a pair's class does not
  fix its answer.}
  \label{fig:guidef}
\end{figure}

At the readout, a per-pair squared error aimed at the total effect $T$ ends up close to $T$. Inside the network, where a cell produces
only part of $T$, the same loss ends up above the effect of whole-state patching at the cell (\cref{fig:guidef}). Over all the cells (trained from
one initialisation, and reported as a description), it lies a median of \inSqTMinusG\,$T$ above $G$, as cross-entropy
does (\inCeMinusG\,$T$), and every other objective that we trained to make the output follow the source also gives an excess.
Of its excess above $G^*$, \inSqTAttn{} runs through attention. DAS trained with this squared error or with the CDAS
divergence also produces off-route effects. Of the \inZero{} live cells where the interval of $G^*$ does not lie above zero,
it is off-route at \inZeroSqTOff{} with the squared error and at \inZeroJsdOff{} with the CDAS divergence, against
\inOffRoute{} with cross-entropy. At all but \inZeroAtT{} of these cells, both effects lie below $T$, so a target or bound at
$T$ rarely constrains them there, whereas whole-state patching, whose effect is near zero at every one of these cells, marks
them (\cref{app:targetfull}).

When the same loss is aimed instead at
the cell's $G$, it ends up near $G$ (\cref{fig:guidef}; slope \inSqGSlope{} and intercept \inSqGIntercept\,$T$ across the
cells, by least squares). Outside IOI, the untrained difference of means also lies near $G$ (slope \inUhatSlope{} and intercept
\inUhatIntercept\,$T$). The comparison leaves out IOI, because there a pair's class does not fix its answer. Summed over the
live cells outside IOI, the difference of means and the mean gradient of $z$ at the cell give \inUhatSum{} and \inBhatSum{}
times the effect of whole-state patching, but their bisector already gives \inBisSum{} times it. The excess therefore depends
on which direction is patched rather than on training as such, and to first order an effect-seeking objective selects a
direction at the upper end of the range (\cref{cor:select}).

At the readout and inside the network alike, the effect of a learned direction follows the target of its objective, so a
learned effect alone cannot measure the network itself. The effect of whole-state patching, $G$, does measure the network, because whole-state patching learns nothing and $G$ is the
indirect effect of the input through the site. It also takes only a few forward passes
per pair to compute (\cref{app:owneffect}).

\FloatBarrier   %

\section{Published Benchmark Results}
\label{sec:published}

Because part of a learned effect is chosen by the objective, published scores based on learned effects show something different once they are compared with whole-state patching. We re-examine the two results from the introduction, CausalGym's
ranking of localisation methods and a sentiment direction reported above 100\%, and we ask what RAVEL and MIB, which score
accuracy instead, can show.

\subsection{The CausalGym Ranking}
\label{app:causalgym}

\begin{figure}[tbp]
  \centering
  \includegraphics[width=\textwidth]{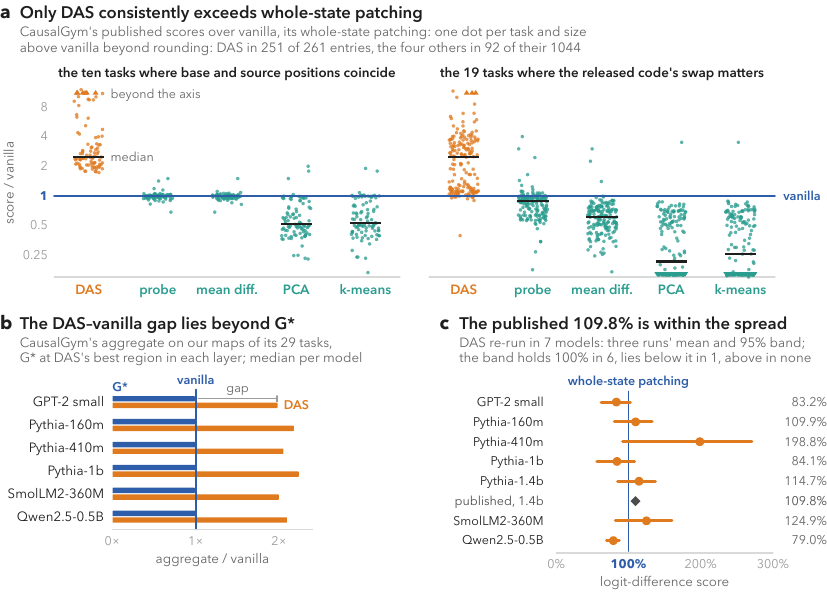}
  \caption{\textbf{What benchmark scores measure.} (a)~CausalGym's published scores. LDA and the random baseline are in
  \cref{tab:causalgym}, so the subtitle's counts cover the four methods drawn. The probe strip shows the higher-scoring of two probes, and \cref{app:indexswap} describes the
  swap. (b)~Each bar is aggregated over layers as CausalGym aggregates a method and divided by vanilla's aggregate; the context of $G^*$ is chosen on the test pairs. \Cref{tab:causalgymmap} gives the share of the gap beyond $G^*$. (c)~The sentiment direction of \citet{hollinsworth2024}, re-run with the authors' code (the
  mean of three trained runs) at a complete cut (\cref{tab:sentiment}). The 95\% bands account for the choice of layer; at Pythia-410m, the band holds 100\% only with its small-sample factor (\cref{app:sentiment}).}
  \label{fig:benchmarks}
\end{figure}

CausalGym ranks seven
localisation methods on 29 tasks by the log odds ratio of their interventions. It takes each method's best region at each
layer and averages over layers. Its vanilla intervention is whole-state patching scored in the same way, which the benchmark
calls ``a significantly more expressive intervention than any methods we tested'' \citep[App.~E]{arora2024}. Even so, the
score of DAS divided by that of vanilla is a median of \cgDasMedLo--\cgDasMedHi{} at every published size
(\cref{fig:benchmarks}a, \cref{tab:causalgym}). The released code passes the base and source token positions to its interventions in swapped
order, which matters on 19 of the 29 tasks (\cref{app:indexswap}). On the other ten, DAS lies above vanilla in all
\cgEntriesAligned{} entries (one entry per task and size), and the other methods in \cgOtherAboveAligned{} of
\cgOtherEntriesAligned{}. With the call corrected, the difference of means and the probes rise to near vanilla on the 19
tasks, while PCA and $k$-means stay low on most of them, and DAS, retrained in the same code, stays above vanilla and above
every other method on all 19 tasks at each of the five sizes that we re-ran.\footnote{We have informed the authors.} Consistent with
\cref{cor:select}, only the method whose objective rewards a larger effect consistently exceeds vanilla. CausalGym itself
suggests that much of the higher score of DAS comes from its access to the outputs, and it reports selectivity, the score
minus that on a control task with arbitrary labels \citep[\S\S5.2--5.3]{arora2024}. By its selectivity tables, the median ratio of the selectivity of DAS to that of vanilla
ranges from \cgSelLo{} to \cgSelHi{} over the nine sizes and lies above 1 only at 14m and 31m, so at every other size DAS
is no more selective than vanilla on the median task. Selectivity, however, controls for a whole task, whereas whole-state
patching gives a control at each cell.

To compare DAS with $G^*$, whole-state patching in its best context, we rebuilt CausalGym's maps in six models
(\cref{app:cgmapfull}). Here the context of $G^*$ is chosen on the test pairs, as CausalGym's aggregate chooses each method's
region on the pairs that it scores. On the rebuilt maps, we compare the excess of DAS over $G^*$ at the best region of DAS in each layer
with the gap between DAS and vanilla, both summed over layers. Per model, this excess is \cgmSNlo--\cgmSNhi{} times the gap,
or \cgmSNsamelo--\cgmSNsamehi{} times without the one task whose $T$ is near zero. At these regions,
$G^*$ thus gives about what the best region of vanilla gives (\cref{fig:benchmarks}b), and the gap between DAS and vanilla is an excess over what
whole-state patching produces at the same regions. The path that CausalGym reads from its negative-polarity map holds on the
rebuilt map of Pythia-1b and follows the cells where whole-state patching has its largest effect. DAS also stays on the
licensor into layers where whole-state patching there has almost no effect, a row that CausalGym's route analysis
\citep[\S6]{arora2024} does not discuss. Neither the excess of DAS over $G^*$ nor this row, by itself,
shows a better-localised variable.

\subsection{Accuracy-Based Benchmarks}
\label{sec:accuracy}

RAVEL and MIB rank methods by how often the intervened output matches the counterfactual label \citep{huang2024,mueller2025}.
The exception is MIB's IOI task, which MIB scores by a squared distance between logits. Where the label
is the source input's own next token, as on our tasks, the count thresholds the log odds of that label over its competitors,
of which $z$, the log odds over $y_b$, is one. Accuracy can therefore show the excess where whole-state patching still
gets some test pairs wrong, but not once it gets every pair right. \citet[\S7]{sutter2025} observed the excess as measured by
accuracy.

We compared accuracies wherever we trained DAS, and we report the comparison as a description. At the readout, the accuracy
of cross-entropy DAS is a median of \roAccCe{} per model, that of the squared error \roAccSq{} and that of whole-state
patching \roAccOwn. There, whole-state patching gives the accuracy of the network itself on the source inputs. In
\roAccUpCe{} of the \roCeAboveAll{} cases where the effect of cross-entropy DAS lies above $T$, its accuracy is significantly
higher than that of whole-state patching, and in \roAccDownCe{} it is significantly lower. Inside the network, at
\inAccUpCe{} of the \inCeAboveG{} live cells where DAS lies significantly above $G$, its accuracy is significantly higher than
that of whole-state patching at the same cell (medians \inAccCeMed{} against \inAccOwnMed), and at \inAccDownCe{} cells it is
significantly lower. Both accuracies are 1, with every test pair right, in \roAccSatCe{} of those readout cases and at
\inAccSatCe{} of those cells, where accuracy cannot show the excess. Where the two accuracies do not differ significantly,
accuracy does not show the excess that the effect shows. Two directions with equal accuracy can thus differ in how much of
their effect the network itself produces, and only the comparison with whole-state patching separates them. These counts
compare cross-entropy DAS with whole-state patching at the same site. They do not cover whole-state patching in its best
context ($G^*$), whose accuracy is not measured, or the size of the excess where the two accuracies agree.

\subsection{A Published Sentiment Direction}
\label{app:sentiment}

\citet{hollinsworth2024} report that a one-dimensional DAS direction scores 109.8\% on their logit-difference metric in
Pythia-1.4b (71.1\% for logistic regression) when their code patches it at every position of a layer. This site is a complete
cut, where whole-state patching scores exactly 100\% (\cref{prop:cut}), and their code trains the direction at the
adjective alone, so its objective does not act on this site (\cref{app:sentimentfull}). We re-ran the authors' code on \srModelsWord{}
models from three initialisations. At 1.4b, the re-run gives \srPubOurs\% [\srPubLo, \srPubHi], a band that accounts
for the choice of layer and contains both the published value and 100\%. The same kind of band includes 100\% in
\srIndistPlainWord{} models for every trained run and lies below 100\% in \srBelowWord, with or without its small-sample
factor (\cref{fig:benchmarks}c). The direction of Pythia-410m, at \srPythiaY\%, is the
exception. Its band includes 100\% only because of that factor, without which the direction lies above 100\% for every trained run. The band, with its small-sample factor, was chosen after the first trained run of Pythia-410m and of Pythia-160m had been seen. In any case, a score above 100\% at a complete cut exceeds the effect of whole-state patching there, which is
exactly 100\%.

\FloatBarrier   %

\section{A Reporting Check for Learned Effects}
\label{app:owneffect}

A learned effect needs a reference at its own site. Control tasks and random-network baselines adjust a comparison for a whole task or a whole network, not for a single site.
A control task asks whether the score of the same network drops when the labels are arbitrary, and a random-network
baseline whether a trained network gives a larger effect than an untrained one. Neither asks whether whole-state patching
in the trained network produces the effect at that cell, or how large this effect is. This question matters because, as
CausalGym notes, ``DAS finds causal effect at all timesteps, even when the model has just been initialised''
\citep[\S6]{arora2024}. The reference therefore has to involve no training, be measured at the same site and transfer
everything that the network computes there, as whole-state patching at that site does, and at a complete cut its effect is
known exactly.

\begin{figure}[tbp]
  \centering
  \includegraphics[width=\textwidth]{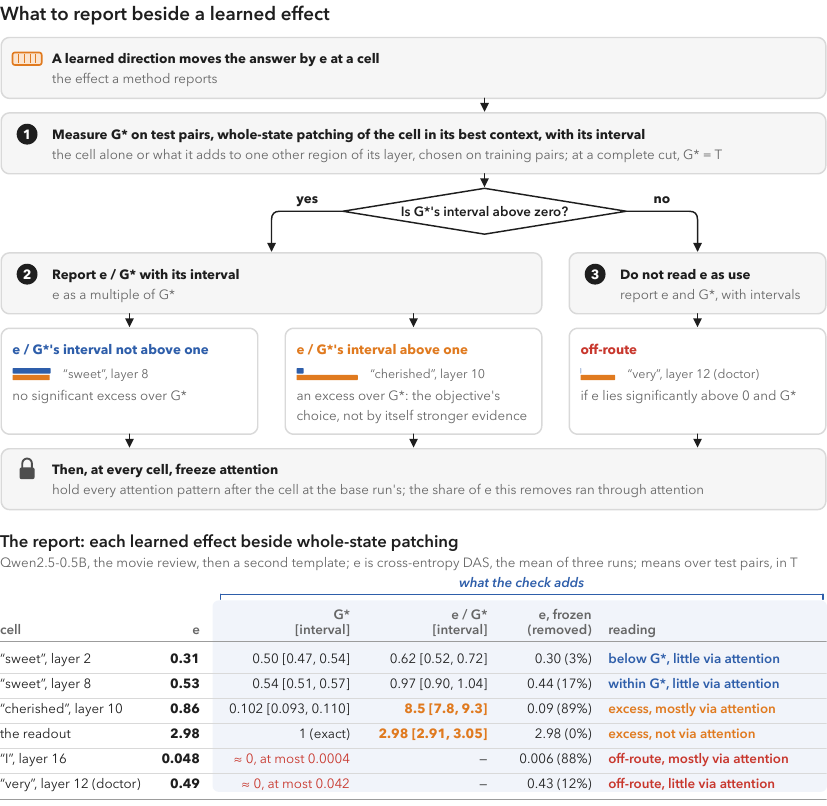}
  \caption{\textbf{What to report beside a learned effect.} The bars show $G^*$ in blue and $e$ in orange (blue is
  whole-state patching and orange DAS, as in \cref{fig:guideb}). Most off-route effects are smaller than the second template's (median \inOffRouteDasMed\,$T$, nine in
  ten below \inOffRouteDasNinety\,$T$). Intervals are 95\% and resample each pair with its mirror.
  OwnEffect computes $G^*$, $e$, $e/G^*$ and the share that freezing removes, with their intervals.}
  \label{fig:guidee}
\end{figure}

Wherever a learned effect $e$ is reported, we recommend comparing it with whole-state patching in three steps, and
\cref{fig:guidee} applies them to the movie review of the running example and to the last word of a second template,
``I heard the doctor use the word vaccination and I felt very''. The steps use the quantities of the route test (\cref{alg:route}), namely $G^*$, the learned effect $e$ and $e$ with
attention frozen. Each quantity that the steps report comes with an interval that resamples the test pairs, each pair
together with its mirror.
\begin{enumerate}
  \item Measure $G^*$, the effect of whole-state patching at the same site in its best context, with its interval. At a
        complete cut, $G^*$ equals $T$ and needs no separate measurement.
  \item Report the ratio $e/G^*$. Where its interval lies above one, attribute the excess $e - G^*$ to the direction that the training
        objective chose, and do not take it, by itself, as stronger evidence about the site.
  \item Where $G^*$ does not lie above zero but $e$ does, do not read $e$ as evidence that the network uses the site. If $e -
        G^*$ also lies above zero, the cell is off-route.
\end{enumerate}
In every case, also freeze attention to see whether $e$ runs through attention.

\paragraph{What the steps measure.} A ratio $e/G^*$ above one
marks an excess of the method above $G^*$. A ratio below one is no evidence against the method, because a direction
patched at the cell alone can fall short of the effect that the cell produces together with another region. The share of $e$ that freezing attention removes shows whether $e$ runs through attention. A benchmark can divide the score of each method by the score of
whole-state patching computed in the same way, as \cref{fig:benchmarks}a does with CausalGym's vanilla intervention.

\paragraph{Cost and library.} At a cell, the steps take at most five forward passes with a patch per test pair and no
gradient, and the direction may come from any method. Our library
OwnEffect measures the quantities of the three steps, with their intervals, for a direction or subspace from any method. At a
cell inside the network, it gives $G^*$, $e$, the excess $e - G^*$, the ratio $e/G^*$ and the share of $e$ that freezing
attention removes; at the readout, where $G^* = T$, it gives $e/T$. Before any training, it also computes the quantities of
the theory on the pairs at hand, among them the range and its upper end, the predicted maximum. It covers
the architectures of all the models that we measure and reproduces our measured values on GPT-2 (\cref{app:owneffectlib}).

\FloatBarrier   %

\section{Related Work}
\label{sec:related}

This section relates our reference and our analysis to four bodies of work. Benchmarks report whole-state patching as a
baseline, and our analysis extends critiques of subspace patching. We also discuss studies of training objectives
and of methods that can create the effects that they report. Finally, our work borrows from the practice of activation
patching.

\subsection{Interchange Interventions and Their Benchmarks}
\label{sec:relbench}

Causal abstraction gives interchange interventions their meaning \citep{geiger2021,geiger2023a}. DAS and its scaled variants
learn the intervened subspace \citep{geiger2023,wu2023}. CausalGym scores
localisation methods by the log odds ratio of their interventions, while RAVEL and MIB score them by how often the
intervened output matches a counterfactual label
\citep{arora2024,huang2024,mueller2025}. RAVEL's Disentangle score also rewards leaving other attributes unchanged. MIB
scores its IOI task by the squared distance between the logit of the network and that of a causal model. Each benchmark also
reports whole-state patching at the same site as a baseline (CausalGym's vanilla, RAVEL's Full Rep., MIB's Full Vector). On
RAVEL's pairs, this baseline is expected to fail, because the pairs also require other attributes to stay unchanged. MIB reports
that on these pairs ``the full vector baseline entirely fails completely'' \citep{mueller2025}. In the BlackboxNLP 2025
shared task, which extends MIB, featurizers that exceed DAS are reported as significant gains \citep{belinkov2025}. A
featurizer maps a hidden state to the features on which an intervention acts.

We use whole-state patching as the reference for the effect of every method, rather than as a baseline in a results
table. Benchmarks
with ground truth use models that are compiled or trained to implement a known algorithm \citep{sharkey2025}. In contrast,
whole-state patching gives a reference for the size of an effect in any trained model, and at a complete cut its value is
known exactly. We show that, to first order, the effect of whole-state patching is the trace of the
matrix whose range contains the effect of every rank-one method, and that the objective of a method determines
where in this range its effect ends up.

\subsection{Interpretability Illusions}
\label{sec:relillusions}

\citet{makelov2023} show that subspace patching can activate a dormant pathway (\cref{ex:two}): ``the variation in the
disconnected part activates the dormant part, which then achieves the causal effect''. Whole-state patching leaves the dormant
part unchanged, so under this reading as well, the effect is one that the network does not produce. \citet{wu2024} reply that such effects can
still be meaningful. Consider the case of one shared direction with
$\theta = 90^\circ$. Then $\operatorname{tr}M = 0$ (\cref{prop:natural}(i), whose proof does not use the angle), while
$\lambda_1 = \tfrac12\|\bar w\|\|m\|$ (\cref{thm:range}(iv) with $\bar w$ and $m$ in place of $w$ and $\Delta$). If, moreover, no pair
moves the state along $\hat w$, then $\hat w$ is dormant in the sense of \citet{makelov2023}. Since $\hat m$ is
orthogonal to the shared readout direction, it is
also disconnected to first order. Their definition is stronger, because it requires that no value along this direction
changes the output. The bisector gives the upper end of the range, and under these conditions it has the form of their
illusory direction. Their Lemma~A.1 finds the same $\pi/4$ angle optimal
under a stronger dormancy assumption. An angle near $90^\circ$ has been measured in practice. Moral framing in a pretrained model is linearly decodable, but it is
nearly orthogonal to the readout direction of the model's judgments and has no measurable
effect on its verdicts
\citep{konrad2026decoded}. In this case, whole-state patching at the readout has almost no effect, while to first order the
upper end of the range does not vanish. At smaller angles, the same geometry produces the inflation that we measure. However, the
illusion of \citet{makelov2023} is defined by its mechanism rather than by its size. In their expository toy model, a linear network writes its input into the state along $(1, 0, 1)$ and reads
its output along $(0, 2, 1)$, so $\Delta$ is a multiple of $(1, 0, 1)$ and $w = (0, 2, 1)$. Then $\cos\theta = 1/\sqrt{10}$,
$\theta \approx 72^\circ$, and their illusory direction $v = (1, 1, 0)/\sqrt{2}$ gives $(w^\top v)(v^\top\Delta) =
w^\top\Delta$, exactly the effect of whole-state patching.
Their IOI table shows both patterns that we measure against whole-state patching at the same site. A residual direction
decreases the logit difference by 140.7\% (interchange accuracy 74.8\%), whereas the whole residual vector decreases it by
123.6\% (54.8\%). Their illusory MLP direction decreases the logit difference by 46.7\%, whereas patching the whole MLP gives
$-8\%$ \citep[Table~1]{makelov2023}. They validate the residual direction as ``faithful to the model's computation to a
high degree'' and argue that patching subspaces of the residual stream is ``less prone to illusions''. The direction they validate also has a larger effect than whole-state patching. The check of \cref{app:owneffect}, however, measures the size of
an effect, not whether the direction of the intervention is the direction of the variable.

\citet{sutter2025} show that causal abstraction becomes vacuous when alignment maps are unconstrained. Our result, in contrast,
needs no non-linearity. A rank-one orthogonal map trained with an effect-seeking objective already reports an effect larger than that
of whole-state patching, and for a linear map their \S7 observes the same kind of excess in accuracy. The site where causal tracing localises a fact need not be the site where editing the fact works \citep{hase2023}. There too, the success of an intervention at a site
and the effect of patching that site come apart. \citet{grant2025} separate the divergent representations that causal interventions create into two kinds.
Harmless ones lie in the behavioural null space, and pernicious ones ``activate hidden network pathways and cause
dormant behavioral changes''. An off-route effect has the behavioural signature of what they call a pernicious divergence. We find such effects for
learned interchanges in unmodified models on natural tasks, and they run mostly through attention (\cref{sec:route}). A learned interchange can activate
such a pathway without leaving the scale of natural states, because its edit $dd^\top\Delta$ is no longer than $\Delta$.

\subsection{Training Objectives and Methods That Can Create the Effects They Report}
\label{sec:relobjectives}

DAS trains its rotation with the distributed form of the interchange intervention training objective of \citet{geiger2021a}.
This objective guarantees an abstraction when the loss is zero, though ``the reverse does not hold''. With cross-entropy, the
loss is zero only when the patched output gives the source label with certainty. This is more than whole-state patching
gives. Until it saturates, the loss therefore keeps rewarding a larger effect.
CDAS keeps the interchange of DAS. It trains this interchange with a Jensen--Shannon divergence to the next-token
distribution of the source run, ``a weaker but more faithful supervision signal'' \citep[\S1]{bao2026}. The divergence is not
a function of the mean effect alone, because it compares whole next-token distributions. At the readout, it ends up exactly at the effect of whole-state patching where one
direction reproduces the source run, and elsewhere it need not (\cref{app:counter}). Inside the network, its target is the
next-token distribution of the source run, not the output of whole-state patching at the site, so it can move the output
where the network itself produces no effect (\cref{sec:target}).

An intervention that is trained to produce an output can create that output. ReFT, for example, trains interventions
built on DAS as a fine-tuning method \citep{wu2024a}. Control tasks show that a probe can learn the task itself
\citep{hewitt2019}. Amnesic probing shows that decodable information need not be used \citep{elazar2021}, and randomisation
tests show that a saliency map can ignore the model \citep{adebayo2018}. Unlike these tests, our reference is
whole-state patching at the same site in the trained network. Apparent emergent abilities can come
from the metric that scores fixed model outputs, and a
continuous metric removes them \citep{schaeffer2023}. The excess that we measure differs in kind. The objective changes the intervention, and the output really
moves further. A change of metric can therefore hide the excess but cannot remove it. Accuracy can show the excess, but
saturated accuracy hides it (\cref{sec:accuracy}).

\subsection{Activation Patching Practice}

Logit-difference recovery and its normalisation are standard \citep{zhang2023,heimersheim2024}. Attribution patching estimates the effect of whole-state patching by a first-order value, the trace of $M$ built with
gradients at the base point (\cref{rem:atp}). The adequacy of its first-order approximation is listed as an open problem
\citep{sharkey2025}. Our test of the trace answers this question in part. At the readout, the trace built with midpoint
gradients lies close to $T$, and on mirrored sets the error of the base-point trace, which attribution patching uses, is to
leading order $-2$ times its error (\cref{prop:trapezoid}). Inside the network, the midpoint trace is the
closer of the two at most cells (\cref{app:firstorderfull}). AtP* notes that attribution patching misses
saturated attention \citep{kramar2024}. \Cref{prop:door} concerns the same saturation. To first order in the score, a small attention weight $\alpha$ becomes large only once the score changes by about $L = \log(1/\alpha)$, a change that a learned direction can reach
where whole-state patching does not.

We borrow attention freezing from path patching \citep{goldowskydill2023} and from \citet{hollinsworth2024}. Steering vectors
act mostly through attention values, whether they are added or trained. Freezing all attention scores during refusal steering
lowers its performance by only 8.83\% \citep{cheng2026}. Learned task vectors steer through the OV circuits of attention
heads \citep{yang2025}. In our measurements, the effect of whole-state patching runs mostly outside the attention weights (freezing
them keeps \inOwnKeep{} of it), whereas the off-route effect of a learned interchange runs largely through the attention weights of the readout position
(\cref{sec:route}).

Circuits can recover more than the logit difference of the full model, and the score changes with the ablation method
\citep{miller2024}. Whole-state patching involves no such choice, because it swaps the whole state exactly and learns
nothing.

\section{Discussion}
\label{sec:discussion}

A large learned effect suggests a simple reading, in which the site encodes the variable and the network uses it there.
Our results instead divide a learned effect into two parts. One part is the effect of whole-state patching at the site.
The
other part is the excess, which, to first order, the objective chooses within the range of effects that the network allows. This section
discusses what the excess measures and what it implies for interpretability and for safety and security. It ends with the
limitations of our account.

\subsection{Interpreting Effects Beyond Whole-State Patching}
\label{sec:excessmeans}

The excess is not an artefact of a metric, as some apparent effects are. The patched output
does move further than under whole-state patching, because some part of the network's state can move it further than the
whole state. Whether such a part exists depends on the network and the inputs, not on any method, while how far a learned method moves the output beyond whole-state patching depends on its objective. To first order, the theory says when such a part exists.
Under a
shared readout direction, the upper end of the range and the trace coincide only when the readout direction and the mean change of the
state point the same way (\cref{cor:select}). On the two-label tasks of our models whose logits are not capped, these
directions lie at a median angle of \roThetaMed$^\circ$, so the upper end and the trace are far
apart. The excess
of a learned method is the part of this gap that its objective selects.

Each earlier explanation of such excesses is consistent with this account (\cref{sec:relillusions,sec:relobjectives}).
Suppose that the readout
direction and the mean change of the state are at a right angle, and that no pair moves the state along the readout
direction. Then the direction that gives the upper end of the range has the form of the illusory direction of
\citet{makelov2023}. However, their illusion is defined by how it arises, not by how large it is. The excess does not need a non-linear alignment map of the kind that \citet{sutter2025} analyse, because
a single direction reaches it. A learned interchange can also activate a hidden pathway in the sense of \citet{grant2025}. \citet[\S1]{bao2026} argue that the training objective
and the intervention protocol determine whether a steering intervention is faithful. In our results, the objective controls the excess. The same squared error ends up above whole-state patching when it is aimed
at $T$, and near it when it is aimed at the cell's own $G$ (\cref{sec:target}).

\subsection{Implications for Interpretability}
\label{sec:impact}

For interpretability, this division bears on what a map or a ranking reports. A map of learned effects shows where an
objective can move the output. When the map is compared with the effects of whole-state patching, it also shows where the network
itself produces the effect and where a learned direction changes the output but whole-state patching does not. This comparison mostly separates
what holds in published readings from what does not, rather than overturning the readings. On the rebuilt map of Pythia-1b, the path that CausalGym infers from its negative-polarity map holds. By contrast, our re-run of the published sentiment direction is not reliably above whole-state patching at its site. The comparison does change what the ranking of
methods shows. A ranking by effect size rewards the objective that seeks the largest effect
(\cref{sec:published}). The same comparison is needed wherever a method is credited with a larger effect. For example, a featurizer that scores above
DAS \citep{belinkov2025} may localise better, or it may report a larger excess. Whole-state patching at the same site shows how much of
its higher score lies beyond what the network itself produces. This comparison does not depend on how a featurizer is
built. The decomposition of \cref{thm:range}, by contrast, covers interventions that patch an orthogonal subspace of a
state, as DAS, probes and the difference of means do. For a sparse dictionary or a non-linear
featurizer, the effects of its parts need not add up to that of the whole, and whole-state patching at the same site
remains the reference.

For the debate about interpretability illusions, the comparison gives a quantity to measure. Consider an effect that
whole-state patching does not produce. It is a question of interpretation whether such an effect is an illusion
\citep{makelov2023} or, as \citet[\S4]{wu2024} would read substantiated cases, ``evidence of latent causal structure rather
than a worrisome illusion''. Whole-state patching does not settle this
question. Instead, it shows at each cell how much of the effect the network itself produces, and
freezing attention then shows how much of the rest of the effect runs through attention.

The comparison also qualifies a common inference. When an intervention along a direction moves the output, this is often taken
as evidence of what the network represents. For instance, the success of activation steering ``can be thought of as
evidence'' for the linear representation hypothesis \citep[\S3.2.2]{sharkey2025}. Our evidence comes from interchanges, not
from activation steering. For interchanges, success alone does not support that inference, because a learned direction
can move the output at a site where whole-state patching does not. Success shows what a direction can make
the network do, whereas whole-state patching at the same site shows what the network itself does.

\subsection{Implications for Safety and Security}

Interventions are also used to change models, not only to explain them. They serve to choose where a model's knowledge
is edited \citep{meng2022}, to steer its behaviour \citep{rimsky2024}, to find the direction that mediates its refusals
\citep{arditi2024} and to audit where fine-tuning installs an evaluation-to-deployment mismatch
\citep{konrad2026routing}. Some of these applications aim to make a model safer, for example by removing a harmful
behaviour or by auditing what drives it. Doing so requires knowing what the effect of an intervention says about the
model. A learned direction that removes a behaviour shows that the behaviour can be changed. Whole-state patching at the
same site, in contrast,
shows whether the network itself produces the behaviour through that site, or whether only the learned direction changes
it there. The steering, refusal and audit directions above are differences of means, not trained directions. Suppose that the
pairs of a mirrored set change the state along one direction, up to sign, and that each pair's class fixes its label. Then, to first order, an interchange along a difference of means, such as these directions, has the same effect as
whole-state patching and so shows no excess, while a direction trained toward a larger effect exceeds it
(\cref{prop:natural,cor:select}). Here too, we
test interchanges, not activation steering, editing or ablation. For those uses, the same question
arises, but our numbers do not carry over to them.

Our results are dual-use, like interpretability in general \citep{bereska2024}. Consider a learned direction that raises the
attention to a site where whole-state patching leaves it as in the base run. Such a direction shows what a patched state at the site can make the network do, but it is also a means to change the model's behaviour. Anyone with a model's weights can already do so more directly. For
instance, editing one direction out of the weights reduces the model's refusals \citep{arditi2024}. We therefore expect our
results to add little to misuse. The larger risk is to read too much into the check. Reporting the check removes one source of overclaiming, but it does not show that a mechanism is complete or that a model is safe.

\subsection{Limitations and Future Work}
\label{sec:limitations}

Our account is first order. At the readout, where the final logits are not capped, it describes the maximum effect and
the effect of whole-state patching closely in all but a few tasks (\cref{sec:tests}). Exact effects go beyond it in two
places, and in both we find why. Under Gemma-2's soft cap, the in-sample maximum exceeds the predicted maximum in
\gmTopAboveLam{} of \gmTasks{} tasks and in none once the cap is removed (\cref{app:cap}). Inside the network, where DAS
exceeds the predicted maximum, freezing attention removes \inCeilAttn{} of its effect (\cref{app:firstorderfull}). There, the
first-order range ranks the cells by their excess only in part and does not bound it (\cref{sec:ranked}), and
\cref{prop:door}, which describes how a learned direction can raise the attention to a site, does not predict where the
effect runs through attention better than simple predictors do (\cref{app:door}).

Our measurements cover base checkpoints, not instruction-tuned ones, of 124M--32B parameters (\cref{tab:models}), at ranks 1
to 64 at the readout and at rank one inside the network, where the four models of 7--9B are described on three syntactic
tasks only. The reference has limits of its own. Inside the network, $G$ and $G^*$ patch the state at the
last token of a region, and $G^*$ adds at most one other region of the layer. An effect that the
network produces only through the earlier tokens of a region, or through three or more regions together, is therefore not
part of $G^*$. The intervals of the maps resample each pair with its mirror, but the maps lack the analysis at the level of
sentences that we report at the readout (\cref{app:readout}).

Because the account concerns objectives rather than DAS (\cref{cor:select}), it covers any method, including methods
yet to come, that trains an orthogonal subspace toward a larger effect. Four directions follow from our results.
\begin{itemize}
  \item \emph{What a localisation should require.} Causal abstraction counts a site as realising a variable when
        interventions there produce the counterfactual output. Constraining the alignment map, for example to be linear,
        does not rule out an excess, because a single orthogonal direction already reaches it (\cref{sec:relillusions}).
        Requiring that the effect be one that the network itself produces at the site would rule it out. How such a
        condition enters the theory of causal abstraction, and which published localisations meet it, is open.
  \item \emph{Where a model can be moved beyond its own effect.} At the readout, a single angle measured before training
        predicts how far a learned part can move the output beyond whole-state patching (\cref{sec:angle}). Inside the
        network, where much of the excess runs through attention, a predictor that includes attention would show before
        any training where a learned intervention can move a model most, which matters both for steering a model
        and for protecting it.
  \item \emph{Hidden routes.} Off-route effects run largely through attention, at sites through which the network itself
        carries no significant effect (\cref{sec:route}). Whether inputs, and not only edits, can open these routes would
        tell whether they belong to the intervention or are latent in the network. Part of this question is which property
        of a task decides where they arise, since more of the effect runs through attention on negative polarity and
        filler--gap than on agreement, and how the share through attention changes with model size.
  \item \emph{Edits made for safety.} An edit trained to remove a behaviour is trained toward an effect. It may therefore
        suppress the behaviour through a route that the network does not use for it, and leave the network's own route in
        place. Testing this needs a reference for edits other than interchanges, such as ablation and steering, and
        measurements on the instruction-tuned models on which such edits are made.
\end{itemize}

\section{Conclusion}
\label{sec:conclusion}

Patching a learned part of a model's state can change the model's output more than patching the whole state does. We asked
whose effect this excess is. To first order, the effects of orthogonal parts add up to that of the whole, so a part exceeds
the whole exactly when the rest moves the output back, and an objective that rewards a larger effect selects such a part.
The network fixes the range of effects that a part can have, and the objective chooses where in this range a learned
direction ends up. At the readout, where all pairs share one readout direction, a single angle measured before training
predicts the upper end of the range, and cross-entropy DAS goes most of the way to it. Changing only the objective of DAS so
that it aims at what whole-state patching produces removes most of the excess, at the readout and inside the network. Inside
the network, cross-entropy DAS exceeds whole-state patching at most sites, and it also changes the output, mostly by small
amounts and largely through attention weights, at sites where whole-state patching shows no significant positive effect. The
excess is therefore largely the objective's, not the network's.

A larger learned effect alone is thus not stronger evidence. The three steps of \cref{app:owneffect} compare it with
whole-state patching wherever it is reported, benchmarks can score methods in the same way, and OwnEffect computes their
quantities with intervals, in a few forward passes per pair, for a direction from any method. Which property of a task decides where
off-route effects arise, and how the share of the effect of DAS that runs through attention changes with model size, remain
open (\cref{sec:limitations}). Read beside whole-state patching at the same site, a learned effect shows how much of it the
network itself produces.

\FloatBarrier

\Needspace{6\baselineskip}   %
\acks{The computations and AI model evaluations for this work were performed on the Bitten AI Supercomputer through the UCloud platform provided by the eScience Center at SDU. The authors declare no competing interests. This work received no third-party funding. \ifx\competing\empty\else{} \competing\fi}

\appendix
\crefalias{section}{appendix}\crefalias{subsection}{subappendix}\crefalias{subsubsection}{subsubappendix}
\topfloatpages
\renewcommand{\floatpagefraction}{0.9}   %
\section{Error Terms, Extensions and Counterexamples}
\label{app:proofs}

This appendix bounds the error of the expansion on which the first-order statements of \cref{sec:theory} rest. It extends the range to rank $k$
and to a direction confined to a subspace, and it gives exact counterexamples that mark where the statements stop holding.
Every statement, proof and exact example is machine-checked in Lean~4 with Mathlib (\cref{app:lean}), except the expected
share of a random projection in \cref{app:confine}. The measurements on trained models are not.

\subsection{The Error of the Midpoint Expansion}
\label{app:error}

Fix a pair and a location. Let $F$ be $z$ as a function of the state at the location, with the rest of the base run
recomputed after the location. Assume that $F$ is three times continuously differentiable on a closed ball around
$h_b + \tfrac12\Delta$ that contains every patched state considered. For a subspace patch $v = PP^\top\Delta$, the ball with
diameter $[h_b, h_s]$ suffices, since $\|v - \tfrac12\Delta\| = \tfrac12\|\Delta\|$. The constant in $O(\|\Delta\|^3)$ below
is a bound on the third derivative of $F$ on this ball. Let $\bar h = h_b + \tfrac12\Delta$, $w = \nabla F(\bar h)$ and
$H = \nabla^2 F(\bar h)$. For a patch $h_b \mapsto h_b + v$, Taylor expansion about $\bar h$ gives
\begin{equation}
  F(h_b + v) - F(h_b) \;=\; w^\top v + \tfrac12 v^\top H (v - \Delta) + O(\|\Delta\|^3)
  \label{eq:taylor}
\end{equation}
for $\|v\| \le \|\Delta\|$, because $\tfrac12(v-\tfrac12\Delta)^\top H(v-\tfrac12\Delta) - \tfrac12(\tfrac12\Delta)^\top H(\tfrac12\Delta) =
\tfrac12 v^\top H(v-\Delta)$.
\begin{itemize}
  \item \textbf{Whole-state patching} ($v = \Delta$). The second-order term vanishes, so $G = w^\top\Delta + O(\|\Delta\|^3)$
        and \cref{thm:range}(ii) holds to third order.
  \item \textbf{Rank one} ($v = dd^\top\Delta$). The second-order term is $-\tfrac12(d^\top\Delta)\, d^\top H(I - dd^\top)\Delta$.
        It does not vanish in general, so \cref{thm:range}(i) and \cref{cor:select} carry a second-order error.
  \item \textbf{Mirrored test sets at the readout}. Suppose that the test set contains every pair together with its mirror, the same pair with base and source exchanged
(\cref{sec:pairs}). The mirror has the same midpoint, the opposite difference $-\Delta$ and the opposite
        measure $-z$ (so $-H$). Its second-order term is the negative of that of the pair, and the mean over the
        test set has no second-order term for any $d$. Every CausalGym test set is mirrored in this sense (all 100 pairs of
        each task).
\end{itemize}
Third order for every $d$ would require an expansion about the midpoint between the base state and the patched state,
$h_b + \tfrac12 dd^\top\Delta$. The gradient at that point depends on $d$, and we do not use this expansion.

\begin{proposition}[the trace's error on a mirrored set]
\label{prop:trapezoid}
At the readout, on a test set that holds every pair with its mirror, let $g(t) = z(h_b + t\Delta)$ on the segment from a
pair's base state to its source state, so that $T = g(1) - g(0) = \int_0^1 g'$. The pair and its mirror together contribute
$\tfrac12\big(g'(0) + g'(1)\big)$ to the trace built with base-point gradients and $g'(\tfrac12)$ to the trace built with
midpoint gradients. These are the trapezoid rule and the midpoint rule for $T$. To leading order their errors are $\operatorname{tr}M_{\mathrm{base}} - T \approx g^{(3)}/12$ and
$\operatorname{tr}M_{\mathrm{mid}} - T \approx -g^{(3)}/24$, so
\begin{equation*}
  \operatorname{tr}M_{\mathrm{mid}} - T \;\approx\; -\tfrac12\big(\operatorname{tr}M_{\mathrm{base}} - T\big),
\end{equation*}
and $\tfrac13\big(\operatorname{tr}M_{\mathrm{base}} + 2\operatorname{tr}M_{\mathrm{mid}}\big)$, Simpson's rule, gives $T$ to fifth order.
\end{proposition}

\begin{proof}
The mirror has base $h_s$, difference $-\Delta$ and measure $-z$, so its base-point term is $\nabla F(h_s)^\top\Delta = g'(1)$
and its midpoint term is $g'(\tfrac12)$, the same as that of the pair. The errors are those of the two quadrature rules for $\int_0^1 g'$.
\end{proof}

For a polynomial path of degree at most four, the relation and Simpson's rule hold exactly. For any path with $|g^{(5)}| \le
B$ on $[0,1]$, they hold up to $B/480$ and $B/720$ (both machine-checked, \cref{app:lean}). The difference
$\operatorname{tr}M_{\mathrm{base}} - \operatorname{tr}M_{\mathrm{mid}} \approx g^{(3)}/8$ estimates the third-order term
from the data before any direction is trained. It thus checks the first-order approximation in each model and task. At a
partial cut, the function $F$ of the mirrored pair is computed on the run of the other sentence, so it differs from that of
the pair, and the relation holds only at a complete cut.

\subsection{Attribution Patching}
\label{app:atp}

\begin{remark}[attribution patching estimates the trace]
\label{rem:atp}
Attribution patching (AtP) estimates the effect of whole-state patching at a site on one pair by $w_b^\top\Delta$, where
$w_b$ is the gradient at the base state \citep{nanda2023atp,kramar2024}. Its signed mean over pairs is the trace of $M$ built
with base-point gradients. This trace agrees with the midpoint trace to third order at a complete cut on a mirrored set, and
to second order elsewhere. AtP thus estimates the reference, which is the effect of whole-state patching, and not a
subspace effect. To first order, AtP reports $\operatorname{tr}M$, and a subspace method with an effect-seeking objective
(\cref{sec:objectives}) reports $\lambda_1$, the upper end of the range.
\end{remark}

\subsection{Rank \texorpdfstring{$k$}{k}}
\label{app:rankk}

For $P$ with $k$ orthonormal columns, \eqref{eq:taylor} with $v = PP^\top\Delta$ gives the first-order mean effect $\E[w^\top
PP^\top\Delta] = \operatorname{tr}(P^\top \E[\Delta w^\top] P) = \operatorname{tr}(P^\top M P)$. By Ky Fan's principle, as in \cref{thm:range}(iii), its
maximum over $P$ is $\sum_{j\le k}\lambda_j$ and its minimum is $\sum_{j > D-k}\lambda_j$. At $k = D$, it equals
$\operatorname{tr}M$.

Under one shared direction (\cref{prop:natural}), a larger rank does not raise the maximum. Take columns $q_1, \dots, q_k$
and let $v = \sum_l (q_l^\top \bar w)\, q_l$ be the projection of $\bar w$ on their span. Then $\operatorname{tr}(P^\top M P)
= \sum_l (q_l^\top \bar w)(q_l^\top m) = v^\top m$. The reflection $2v - \bar w$ has the same norm as $\bar w$, so
$2\operatorname{tr}(P^\top M P) - \bar w^\top m = (2v - \bar w)^\top m \le \|\bar w\|\|m\|$. Every rank-$k$ interchange
therefore stays at or below $\tfrac12(\bar w^\top m + \|\bar w\|\|m\|) = \lambda_1$, the first-order maximum at rank one. For
every $k < D$, some interchange reaches this value. One example is the top eigenvector of $M$ completed by $k - 1$
orthonormal directions that are orthogonal to it and to $\bar w$, which add nothing. An effect-seeking objective in pure
form, that is, one strictly decreasing in the mean effect (\cref{cor:select}(i)), therefore reports the same excess at every
rank below $D$. Unless $\bar w$ and $m$ point in the same direction, such an objective returns the trace only at $k = D$,
that is, for the whole state. Without one shared direction, $M$ can have several positive eigenvalues, and the maximum grows
with the rank. Where a task has more than two labels, as IOI with twenty names does, rank-one interchanges can understate
what a larger rank reports. \Cref{app:rankkmeasured} measures this.

\subsection{Confining the Direction to a Subspace}
\label{app:confine}

Let $U$ be a subspace with orthonormal basis $P\in\mathbb R^{D\times k}$, let $\Pi = PP^\top$, and suppose that $U$ contains
every difference, $\Pi\Delta_i = \Delta_i$. (i)~Pair by pair, the rank-$k$ interchange along $P$ is whole-state patching,
since $h_b + \Pi\Delta = h_s$. A special case is a single code, in which every difference is a multiple of one
vector, $\Delta_i = s_i\bar\Delta$. The rank-one interchange along $\pm\bar\Delta/\|\bar\Delta\|$ is then whole-state
patching, whatever the objective. Such an interchange can neither exceed the effect of whole-state patching nor change
attention in a way that whole-state patching does not. (ii)~To first order, a unit $d\in
U$ has $e(d) = d^\top \Pi M \Pi\, d$, so only $\Pi w$ enters. The compression $P^\top MP$ has the same trace as $M$, and its
range lies inside the range of $M$. Confinement thus shrinks the range around the same trace, and the confined excess never
exceeds the unconfined excess, although it need not vanish for $k \ge 2$. For example, take two pairs, one with $\Delta_1 = \mathbf{e}_1$ and $w_1 = \mathbf{e}_1 + \mathbf{e}_2$ and one with
$\Delta_2 = \mathbf{e}_2$ and $w_2 = 0$. They give a confined maximum of $1.207\,\operatorname{tr}M$ on $U =
\operatorname{span}(\mathbf{e}_1, \mathbf{e}_2)$. (iii)~For an attention score, the same argument replaces $a$ by
$\Pi a$. For $k\ge2$, the largest score change over unit $d\in U$ is $\tfrac12(a^\top\Delta + \|\Pi a\|\|\Delta\|)$. A
confined direction can therefore raise the attention to the site beyond whole-state patching (\cref{prop:door}) only to the
extent that $U$ contains the score's gradient.

\paragraph{The effect of subspace size alone.} Now drop the assumption that $U$ contains the differences, and assume instead the
hypothesis of \cref{prop:natural} with $m\in U$. The first-order confined maximum, $\max_{d\in U,\|d\|=1}(d^\top\bar
w)(d^\top m)$, is then $\tfrac12\big(1 + \|\Pi\bar w\|/(\|\bar w\|\cos\theta)\big)$ times the trace. Suppose that $U$ is
spanned by the code direction and $k-1$ random directions. In expectation, the projection on $U$ then keeps a $(k-1)/(D-1)$
share of the squared length of the part of $\bar w$ orthogonal to the code direction, and so about a
$\sqrt{(k-1)/(D-1)}$ share of its length.
For $k = 8$ and $D = 2048$, the confined maximum is about $1.003$, $1.03$ and $1.10$ times the trace at $\theta = 60^\circ$,
$80^\circ$ and $85^\circ$, against $1.50$, $3.38$ and $6.24$ for an unconfined direction.

\subsection{Counterexamples and Objectives Aimed at a Target}
\label{app:counter}

\paragraph{An indefinite $M$ need not make the maximum exceed the trace.} Two pairs with $w_1 = \mathbf{e}_1$, $\Delta_1 = \mathbf{e}_1 +
\tfrac12\mathbf{e}_3$ and $w_2 = \mathbf{e}_2$, $\Delta_2 = \mathbf{e}_2 - \tfrac12\mathbf{e}_3$ give
\[
  M = \begin{pmatrix} 0.5 & 0 & 0.125 \\ 0 & 0.5 & -0.125 \\ 0.125 & -0.125 & 0 \end{pmatrix},
\]
with $\operatorname{tr}M = 1$ and eigenvalues $0.5$ and $0.25 \pm \sqrt{0.09375} \approx 0.556, -0.056$. The maximum is below
the trace, although $M$ is indefinite.

\paragraph{The maximum can lie below $T$.} Take two pairs whose readout directions are orthogonal, $w_1 = \Delta_1 =
\mathbf{e}_1$ and $w_2 = \Delta_2 = \mathbf{e}_2$. They give $M = \tfrac12 I$ on their span, so $\operatorname{tr}M = 1$,
while every unit $d$ has $d^\top M d \le \tfrac12$. The best rank-one interchange reaches half the trace, and an objective
aimed at $T$ has no solution (\cref{cor:select}(iii) requires $T$ to lie inside the range). At real readouts, with LayerNorm
and unembedding, the out-of-sample maximum of IOI with twenty names (\roIoiTwentyLp{} label pairs) lies below $T$ in every model
(\cref{app:rankkfull}).

\paragraph{A per-pair squared error can miss $T$.} Take two pairs with $w_1 = \mathbf{e}_1$, $\Delta_1 = \mathbf{e}_1 +
3\mathbf{e}_2$ and $w_2 = \Delta_2 = \mathbf{e}_3$. They have per-pair effects $T_1 = T_2 = 1$, and $T = \operatorname{tr}M =
1$ lies inside the range, since the unit $(\mathbf{e}_1 + \mathbf{e}_2)/\sqrt2$ reaches it. Yet at every unit $d$ whose mean
effect is $T$, the per-pair squared error $\tfrac12\sum_i(e_i(d) - T_i)^2$ is at least $100/121$. At the unit $(\mathbf{e}_1
+ \mathbf{e}_3)/\sqrt2$, it is $1/4$, so its minimiser avoids $T$. Every minimiser lies at a mean effect of $(160 -
9\sqrt{10})/185 \approx 0.71\,T$.

\paragraph{Objectives aimed at a target (\cref{cor:select}(ii)).} To bound the measured effect, an objective has to state
its target in the unit of the measurement. The divergence of CDAS \citep{bao2026} targets the whole next-token distribution
of the source run, not the mean effect. When all pairs share a single code, the divergence reaches its target exactly,
because the patch along the code reproduces the source run. Variation between pairs that shows in the distribution over the
vocabulary can move its minimiser away from the code, in either direction. A per-pair squared error on $z$ is not a
target-aimed objective in pure form either, because it depends on more than the mean effect. It reaches $T$ when one readout
direction lets a single $d$ reproduce every pair, and every minimiser then has mean effect $T$ (\cref{app:lean}). Otherwise,
it can end up well below $T$, as in the example above. At the readout of our models, the divergence ends up at a median of
\roJsdMedian\,$T$ per model (\cref{sec:landing}), but it ranges from \roJsdMin{} to \roJsdMax\,$T$ across cases. The
squared error lies within \roSqDevQthree\% of $T$ in three quarters of the cases (\cref{app:landing}).

\subsection{Machine-Checked Statements}
\label{app:lean}

Every statement of \cref{sec:theory} except \cref{rem:partial}, which is a measurement, is checked in Lean~4 (v4.35.0-rc4)
with Mathlib, and so are \cref{prop:door} and every statement of this appendix except the expected share of a random
projection in \cref{app:confine}. In full, the checked statements are
\cref{prop:cut},
\cref{thm:range} with its rank-$k$ form under one shared direction and its sum over any orthonormal basis, and the statement
that a part exceeds the whole only through its component orthogonal to $\Delta$, with its example. They also include
\cref{cor:select}, the derivative of cross-entropy with respect to the base label's logit, \cref{prop:natural,prop:door}, and
the midpoint expansion of \cref{app:error} and \cref{prop:trapezoid} with their remainders. Finally, they include
\cref{rem:atp}, confinement (\cref{app:confine}), the exact examples and the minimum of the squared error in
\cref{app:counter}, and the bound of \cref{app:cap}. That bound shows that, where its two conditions hold and the two labels'
raw logits move toward the source's answer, the cap can only raise exact effects relative to first order. A first-order statement is checked as an exact statement about the first-order
effect that it defines. The step from the network to that quantity is the expansion of \cref{app:error}. This expansion is
checked for whole-state patching and for mirrored pairs. It is also checked for a general patch when $F$ is differentiable
on the whole space, which holds for the networks we measure. The
Lean sources are \codeurl. One command builds every theorem, with no unproved step and with only Lean's standard axioms, and
the notes of the sources map each statement to its theorems and files. The statements about trained models are measurements
and are not formalised. The Lean statement of \cref{prop:cut} takes as its hypothesis that the cuts of \cref{def:cut} factor
through their states. This hypothesis is exactly what that definition states.

\FloatBarrier
\section{Experimental Details}
\label{app:details}

This appendix describes the data, the models, the training of each direction and the compute, in enough detail to repeat the
experiments. \Cref{sec:experiments} explains where we intervene, how each direction is scored and the route test
(\cref{alg:route}).

\subsection{Data}
\label{app:data}

Every source uses CausalGym's format, in which a base sentence and a source sentence are split into regions, and each
sentence comes with the label that continues it. We build each source as its authors' code builds it, with the changes that
our method needs and a few stated choices. The changes are one-token labels, minimal pairs where the source has them, and
training and test rows from one task. The data notes of the repository list every change and its reason.

\suppressfloats[t]   %
\begin{table}[tbp]
  \caption{\textbf{One test pair from each data source.} Each row shows the first pair of one task's test file, split into
  its regions. Regions are separated by dots, a region where base and source differ is written base/source, as in the map
  figures, and the start-of-text token is left out. Labels: the next token of the base input, then that of the source
  input. Label pairs: the distinct pairs among the task's test pairs. Train/test: the pairs in the task's files; the readout
  keeps those whose labels are single tokens. Kumon \& Yanaka's, Tigges et al.'s and RAVEL's tasks are built per model and
  shown as built for GPT-2. \Cref{tab:readout} counts the tasks of each source. The maps of \cref{sec:offroute} cover
  \inTasks{} tasks of these sources in the six small models (\cref{app:fullgrid}), and in the four large models they cover the
  first task shown and two more of CausalGym's tasks.}
  \label{tab:tasks}
  \centering
  \footnotesize
  \setlength{\tabcolsep}{4pt}
  \hyphenpenalty=10000 \exhyphenpenalty=10000
  \begin{tabular}{@{}>{\raggedright\arraybackslash}p{0.25\textwidth}>{\raggedright\arraybackslash}p{0.4\textwidth}>{\raggedright\arraybackslash}p{0.12\textwidth}cc@{}}
    \toprule
    source: task & pair & labels & \shortstack{label\\pairs} & \shortstack{train/\\test} \\
    \midrule
    CausalGym: negative polarity\newline \texttt{npi\_\allowbreak{}any\_\allowbreak{}subj-\allowbreak{}relc} & \textbf{No/The} \textperiodcentered\ farmer \textperiodcentered\ that \textperiodcentered\ contacted \textperiodcentered\ no \textperiodcentered\ guard \textperiodcentered\ has \textperiodcentered\ caught & any / some & 1 & 400/100 \\
    CausalGym, number only: agreement\newline \texttt{agr\_\allowbreak{}sv\_\allowbreak{}num\_\allowbreak{}subj-\allowbreak{}relc\_\allowbreak{}\_\allowbreak{}num} & The \textperiodcentered\ \textbf{senators/senator} \textperiodcentered\ that \textperiodcentered\ hurt \textperiodcentered\ the \textperiodcentered\ athletes & have / was & 9 & 400/100 \\
    Kumon \& Yanaka: negative polarity\newline \texttt{NPIConditionals} & The actors \textperiodcentered\ riding \textperiodcentered\ a mountain \textperiodcentered\ \textbf{if/while} \textperiodcentered\ the cousins \textperiodcentered\ are & ever / still & 5 & 200/50 \\
    Tigges et al.: movie template\newline \texttt{sentiment\_\allowbreak{}movie} & I thought this movie was \textperiodcentered\ \textbf{sweet/vile} \textperiodcentered\ , I \textperiodcentered\ \textbf{cherished/hated} \textperiodcentered\ it. Conclusion: This movie is & great / terrible & 1 & 240/100 \\
    SST: continuation\newline \texttt{sst\_\allowbreak{}continuation} & \textbf{I like Frank the Pug, though./The charms of willful eccentricity, at least as evidenced by this latest cinematic essay, are beginning to wear a bit thin.} \textperiodcentered\ Overall the movie was just very & good / bad & 1 & 400/100 \\
    RAVEL: city to continent\newline \texttt{ravel\_\allowbreak{}city\_\allowbreak{}continent} & Rome is a city in the continent of Europe. \textperiodcentered\ \textbf{Yishui/Marbella} \textperiodcentered\ is a city in the continent of & Asia / Europe & 1 & 400/100 \\
    IOI, Wang et al.: twenty names\newline \texttt{ioi\_\allowbreak{}wang\_\allowbreak{}20} & While \textperiodcentered\ Jessica \textperiodcentered\ and \textperiodcentered\ Andrew \textperiodcentered\ were working at the store, \textperiodcentered\ \textbf{Andrew/Jessica} \textperiodcentered\ gave a ring to & Jessica / Andrew & 43 & 400/100 \\
    IOI, ours: one name pair\newline \texttt{ioi\_\allowbreak{}fixed} & Then, when \textperiodcentered\ Mary \textperiodcentered\ and \textperiodcentered\ John \textperiodcentered\ went to the school, \textperiodcentered\ \textbf{John/Mary} \textperiodcentered\ gave a phone to & Mary / John & 1 & 400/100 \\
    \bottomrule
  \end{tabular}
\end{table}

\paragraph{Sources of the main measurement.} These sources are used in \cref{sec:inflation,sec:offroute}. \Cref{tab:readout}
counts the tasks of each source, and \cref{tab:tasks} shows one test pair from each source.
\begin{itemize}
  \item \textbf{CausalGym} \citep{arora2024} (MIT licence). We use the 29 tasks as released, with the authors' mirrored
  splits of 400 training and 100 test pairs. We also re-pair its six number-agreement tasks so that base and source differ
  only in the number of the subject. These are the number-only agreement tasks. As a control, a second set of pairs differs
  only in the lemma of the subject. Both sets are balanced by class but not mirrored pair by pair.
  \item \textbf{Kumon \& Yanaka} \citep{kumon2026a} (MIT licence). We use seven filler--gap and eight negative-polarity
  constructions as the authors' loader builds them (the first half of the test rows, mirrored, with each label represented by
  its first token). The negative-polarity pairs come in two forms. One is the authors' one-directional form (every base
  label is \emph{ever}), and the other is mirrored. The authors' out-of-distribution test file and a test split without
  strings shared with training serve as held-out tests (\cref{app:safeguards}).
  \item \textbf{Sentiment} \citep{hollinsworth2024} (MIT licence). We use the English templates of the authors' code. Some
  templates differ only in punctuation, quotes or one word, and we keep one template from each such group. We also leave out
  the template whose variable is which review is asked about. We apply the authors' per-model filter to one-token words,
  combine every adjective with every verb, and use the authors' answer pair for each template (\emph{great}/\emph{terrible}
  on five templates; \emph{happy}/\emph{sad}, \emph{good}/\emph{bad}, \emph{Positive}/\emph{Negative} or
  \emph{pleased}/\emph{disappointed} on the others). The authors' transfer setting, in which a direction trained on the movie
  template is evaluated on another template, serves as a held-out test (\cref{app:safeguards}). For \emph{SST}
  \citep{socher2013}, we use sentences with the labels and split of the same repository (MIT licence), in its two prompt
  scaffolds. Positive and negative sentences are paired at random, so they are not minimal pairs. There are 200 training and
  50 test pairs, each with its mirror (400 and 100 in the task's files, as \cref{tab:tasks} counts them).
  \item \textbf{RAVEL} \citep{huang2024} (MIT licence). We build two-label tasks from two values of an attribute, with base
  and source sharing one prompt. An entity is kept for a model when the model's most likely next token is the label (RAVEL's
  criterion). Each prompt is used both in the form that RAVEL gives it and with a start-of-text token. A version rebuilt with
  test rows from RAVEL's test prompts and training rows from its other prompts serves as a held-out test
  (\cref{app:safeguards}). Many prompts open with one example, as the prompt of \cref{fig:law} does (\emph{People in Cape
  Town speak Afrikaans, English, Xhosa.}, which the figure leaves out).
  \item \textbf{IOI} \citep{wang2022} (MIT licence). We use the authors' templates with two or twenty one-token names, with
  and without a start-of-text token. We also use our own sentence frame with our own names, which carries a start-of-text token.
  Training on ten of the authors' fifteen templates and testing on the other five serves as a held-out test
  (\cref{app:safeguards}).
\end{itemize}

The sentiment direction of \cref{app:sentiment} follows the authors' own protocol, which has \srTestPrompts{} test
prompts per model.

\subsection{Models}
\label{app:models}

\Cref{tab:models} lists the models. All our own runs use float32 and eager attention, in evaluation mode. CausalGym's code
runs as its authors set it up (\cref{app:compute}). The readout measurement covers all fourteen models. The maps inside the
network cover the six small models (124M--1B) and, on three CausalGym tasks, the four large models (7--9B). The other four
models, Qwen2.5-1.5B and 32B and Llama-3.2-1B and 3B, are measured only at the readout (\cref{sec:limitations}). Gemma-2 caps
its final logits at 30, and \cref{app:cap} measures it with and without the cap.

\begin{table}[tbp]
  \caption{\textbf{Models.} Width is the width of the residual stream, and heads counts the query heads, with the key--value
  heads in parentheses where there are fewer. The last three columns show where each model is used. Readout refers to
  \cref{sec:inflation} and \cref{app:readout} (every readout analysis), and maps refers to \cref{sec:offroute}. Benchmarks
  refers to \cref{sec:published}, where c marks CausalGym's released code, s the code of the sentiment authors, and p our
  maps under CausalGym's protocol. All models are base checkpoints (not instruction-tuned) under the Hugging Face
  organisations of their authors (EleutherAI, \texttt{gpt2}, HuggingFaceTB, Qwen, meta-llama, google). Our runs read them at
  pinned revisions, and CausalGym's code reads them at their current address. Pythia is the non-deduplicated suite; at 14m
  and 31m, the weights at that address were replaced in February 2026 (\cref{app:indexswap}). Licences are as released by
  their authors. Sources: Pythia \citep{biderman2023}; GPT-2 \citep{radford2019}; SmolLM2 \citep{allal2025}; Qwen2.5 \citep{qwen2024}; Llama \citep{dubey2025}; Gemma-2 \citep{team2024}.}
  \label{tab:models}
  \centering
  \footnotesize
  \setlength{\tabcolsep}{4pt}
  \begin{tabular}{@{}lrrrlccl@{}}
    \toprule
    model & layers & width & heads & licence & readout & maps & benchmarks \\
    \midrule
    Pythia-14m & 6 & 128 & 4 & Apache-2.0 &  &  & c \\
    Pythia-31m & 6 & 256 & 8 & Apache-2.0 &  &  & c \\
    Pythia-70m & 6 & 512 & 8 & Apache-2.0 &  &  & c \\
    Pythia-160m & 12 & 768 & 12 & Apache-2.0 & $\checkmark$ & $\checkmark$ & c, s, p \\
    Pythia-410m & 24 & 1024 & 16 & Apache-2.0 & $\checkmark$ & $\checkmark$ & c, s, p \\
    Pythia-1b & 16 & 2048 & 8 & Apache-2.0 & $\checkmark$ & $\checkmark$ & s, p \\
    Pythia-1.4b & 24 & 2048 & 16 & Apache-2.0 &  &  & s \\
    Pythia-6.9b & 32 & 4096 & 32 & Apache-2.0 & $\checkmark$ & $\checkmark$ &  \\
    GPT-2 small & 12 & 768 & 12 & MIT & $\checkmark$ & $\checkmark$ & s, p \\
    SmolLM2-360M & 32 & 960 & 15 (5) & Apache-2.0 & $\checkmark$ & $\checkmark$ & s, p \\
    Qwen2.5-0.5B & 24 & 896 & 14 (2) & Apache-2.0 & $\checkmark$ & $\checkmark$ & s, p \\
    Qwen2.5-1.5B & 28 & 1536 & 12 (2) & Apache-2.0 & $\checkmark$ &  &  \\
    Qwen2.5-7B & 28 & 3584 & 28 (4) & Apache-2.0 & $\checkmark$ & $\checkmark$ &  \\
    Qwen2.5-32B & 64 & 5120 & 40 (8) & Apache-2.0 & $\checkmark$ &  &  \\
    Llama-3.2-1B & 16 & 2048 & 32 (8) & Llama 3.2 Community & $\checkmark$ &  &  \\
    Llama-3.2-3B & 28 & 3072 & 24 (8) & Llama 3.2 Community & $\checkmark$ &  &  \\
    Llama-3.1-8B & 32 & 4096 & 32 (8) & Llama 3.1 Community & $\checkmark$ & $\checkmark$ &  \\
    Gemma-2-9B & 42 & 3584 & 16 (8) & Gemma Terms of Use & $\checkmark$ & $\checkmark$ &  \\
    \bottomrule
  \end{tabular}
\end{table}

\paragraph{Layers and token positions.} In Hugging Face, layer $\ell$ (\cref{sec:sites}) is \texttt{hidden\_states[$\ell+1$]},
except that the library's last entry already includes the final normalisation. The readout (\cref{def:cut}) is the raw
output of the last block at the last position, before this normalisation. The patch at a cell copies the source run's state
at the last token of the same region (\cref{app:sites}). This token sits at a different absolute position whenever the two
sentences have different numbers of tokens before it. On CausalGym's test pairs, in the \tokTokenizersWord{} tokenisers of
the six small models, this never happens for the negative-polarity and filler--gap tasks. It happens for
\tokNumLo--\tokNumHi\% of the number-agreement pairs, for \tokGenLo--\tokGenHi\% of the gender-agreement pairs, and for every
pair of the \tokEmptyWord{} tasks whose sentences leave a region empty.

\subsection{Training the Directions}
\label{app:training}

DAS is rank one. We train a free vector $v$, initialised as a standard Gaussian draw scaled to unit length, and patch $h_b +
dd^\top\Delta$ along $d = v/\|v\|$. Training makes one pass over the task's training pairs in batches of 4 (100 steps on 400
pairs), with Adam at a learning rate of $5\cdot10^{-3}$. At the readout, we also train for ten passes, each pass in a new
order. The learning rate, batch size and single pass follow CausalGym, except that our rate is constant, whereas
CausalGym's follows a schedule. Cross-entropy and the Jensen--Shannon divergence of CDAS are computed over the full
vocabulary. The per-pair squared error is the batch mean of $(e_i - T_i)^2$, where $e_i$ is the effect of the patch on pair
$i$ and $T_i$ is the total effect of that pair. Inside the network, a second version uses $G_i$, the effect of whole-state
patching at the cell on pair $i$, in place of $T_i$. The mean-effect objective is minus the batch mean of $e_i$. One seeded
generator draws both the initial direction and the batch order. Its seed is $100(\ell+1) + r$ at layer $\ell$ and region $r$
of a map. At the readout, it is $2000$ plus the task's index in its readout task list (the repository gives each list). Each
further initialisation adds $100{,}000$ (maps) or $1000$ (readout). CausalGym's 29 tasks are in alphabetical order, and a
number-only task takes the position of the CausalGym task that it re-pairs.

The representation-only directions of \cref{app:indexswap} follow CausalGym's definitions and are fitted on base states with
the base class as the label. The probes use L2 penalties with the saga solver, and $k$-means uses ten initialisations. The
sentiment direction (\cref{app:sentiment}) is trained with the authors' code. This code uses an orthogonally parametrised
rotation and Adam with learning rate $10^{-3}$ for 64 epochs, trains at the adjective position and keeps the direction of the
last epoch.

\subsection{Evaluation}
\label{app:evaluationfull}

Effects are evaluated in batches of 25 with right padding on the maps, and in batches of equal length on the readout runs.
At the readout and on the maps, intervals use 2000 resamples of test clusters (percentile intervals, and basic intervals for
ratios involving $\lambda_1$). RAVEL's pairs share entities across clusters. At the readout we therefore use an analytic
two-way cluster interval by the entity of each value \citep{cameron2011} for four RAVEL quantities, namely $T$ in the
population rule, the margin above $T$, the excess and the comparison with the squared error (\cref{app:readout}). For the
sentiment direction, intervals use 10,000 circular-block resamples of neighbouring prompts (blocks of two), and a band that
is simultaneous over the layers accounts for the choice of layer (\cref{app:sentiment}). Because there are few blocks, the
band includes a \emph{small-sample factor}, which widens it (\cref{app:sentimentfull}).

\subsection{Compute and Availability}
\label{app:compute}

The measurements ran on NVIDIA B200 GPUs\computeplace, in float32 with TF32 off and with deterministic kernels. PyTorch
2.14.1 and Transformers 5.18.0 were pinned, and every model was loaded at a pinned revision. Every output was checked against
a run record that holds its sha256 and the fingerprint of the code that wrote it. With several runs sharing a GPU, the
readout measurement of one model takes \ruSmall{} hours at 124M--1.5B, \ruThree{} hours at 3B and \ruLarge{} hours at
7--8B. It takes \ruGemma{} hours for Gemma-2, which is measured with and without its cap, and \ruThirtyTwo{} hours for
Qwen2.5-32B, which runs alone on its GPU. A map inside the network takes \inMapHours{} hours at 124M--1B (median
\inMapHoursMed). CausalGym's own code also ran on the GPU, in its own environment (PyTorch 2.7.0, Transformers 4.46.2, pyvene
0.1.6) and with the models at their current address. The checks of its token positions (\cref{app:indexswap}) ran on the same
machine. The code, the scripts that build every data source and the Lean proofs are \codeurl. They come with OwnEffect, a small library that measures a learned effect beside whole-state patching at the same site, with
intervals, for a direction from any method (\cref{app:owneffectlib}). Its name refers to the effect that the network itself
produces at a site. One script per
experiment re-runs that experiment, and one script rebuilds each table and figure from
the outputs of the scoring scripts.

\FloatBarrier
\section{Details and Further Results at the Readout}
\label{app:extended}

This appendix describes the measurement behind \cref{sec:inflation}, with its population and checks. It also reports every
further test of the findings of that section, in the order of its subsections.

\subsection{Details of the Readout Measurement}
\label{app:readout}

\paragraph{The measurement.} Every number in \cref{sec:inflation} comes from one measurement. For each model and task, it
uses the training and test pairs whose labels are single tokens. At the readout cell, it measures the effect of whole-state
patching, $T$, and the exact effect of every direction on every test pair. It builds $M$ from midpoint gradients
(\cref{thm:range}) and computes its top eigenvalue exactly in the span of the pairs. Each initialisation has its own batch
order, and the four objectives (cross-entropy, the squared error, the divergence of CDAS and the mean-effect objective)
train together on this order. The effect of an objective is the mean of the per-pair effects of its three initialisations.
The maximum effect of \cref{sec:tasks}, the out-of-sample maximum, is found on the training pairs and measured on the
test pairs. The search uses Adam on an unnormalised vector $v$ with $d = v/\|v\|$. It takes 100
full-batch steps on the training pairs, with a step size of $10^{-3}$ that decays on a cosine schedule to $10^{-5}$, and it
keeps the best iterate. It starts from two directions. The first is the direction that the mean-effect objective reaches
after ten passes, from whichever of its three initialisations ends with the largest effect on the training pairs. The second
is the top eigenvector of $M$ built from midpoint gradients on the training pairs. The in-sample maximum (\cref{sec:tasks}) comes from a separate search on the test pairs and is measured on the same pairs. It is the exact counterpart of $\lambda_1$, which is also
computed on the test pairs, and it is never below the out-of-sample maximum. The margin
above $T$, the intervals (for RAVEL, two-way by entity) and the counting rule are those of \cref{sec:experiments}. After
false-discovery control over the tasks of a model (Benjamini--Hochberg, $q = 0.025$), cross-entropy DAS lies above $T$ in
\roCeAboveBH{} rather than \roCeAbove{} of the tasks with a margin above $T$. The mean-effect objective does so in
\roEffAboveBH{} rather than \roEffAbove{} of them. In Gemma-2, at least one trained run of cross-entropy DAS lies above
$\lambda_1$ in \gmCeilingTasksBH{} rather than \gmCeilingTasks{} tasks. \Cref{tab:readout} gives the results for every
source. RAVEL's plain and JSON prompts of one attribute (\roDupRavel{} model--task cases) and SST's two prompt scaffolds
(\roDupSst) count as separate tasks. With each construction counted once, cross-entropy DAS lies above $T$ in
\roCeAboveConstr{} of the \roRoomConstr{} cases with a margin above $T$, with the same \roCeMissConstr{} misses.

\paragraph{The population.} The main population consists of \roTasks{} cases in the \roModelsWord{} models, selected by the
population rule of \cref{sec:tasks}. The rule asks for at least 20 test clusters, which the bootstrap needs, and for an
interval for $T$ above zero, which is needed because effects are expressed in units of $T$. Both
conditions are fixed before any training and do not depend on any trained direction. Of the tasks with data, the
two conditions leave out \roExcl{} model--task cases. Of these, \roExclFew{} have \roExclFewClusters{} test clusters. Some of
them are sentiment templates, where the one-token word filter leaves fewer sentences. These are one template in each of Pythia-160m, 410m, 1b and 6.9b, four templates in SmolLM2-360M and the book-review template in GPT-2 and Gemma-2. The others are four
RAVEL attributes, in Pythia-1b, Pythia-6.9b and Llama-3.2-1B (two). In these cases, cross-entropy DAS lies at
\roExclFewCe\,$T$, as it does elsewhere. In the remaining \roExclFlat{} cases, the input does not change the output in the
way that the labels assume. Two of them are the \texttt{filler\_gap\_embed\_4} tasks of Qwen2.5-0.5B and Pythia-160m, and one
is Llama-3.2-3B's JSON prompt of a city's latitude. The intervals for $T$ of these three include zero (for RAVEL, by the
two-way interval). The other two are Gemma-2's two IOI tasks on the templates of \citet{wang2022}, with $T$ at \gmIoiRawT.
Gemma-2 expects a start-of-text token, which the prompts of RAVEL and IOI lack in the form that their sources use. Without
this token, Gemma-2 names the wrong person on those templates, and too few RAVEL prompts give both answers, so only
\gmRavelRaw{} of the population's \gmRavelPop{} RAVEL tasks are built. With the token, its IOI $T$ is \gmIoiSotT{}, and
\gmRavelSot{} RAVEL tasks are built. Cross-entropy DAS then lies above $T$ in \hoGemmaRavelAbove{} of the \hoGemmaRavel{}
RAVEL tasks, and the remaining one has no margin above $T$. It also lies above $T$ in the two-name IOI task
(\hoGemmaIoiCe\,$T$), and the twenty-name task has no margin above $T$. In Llama-3.1-8B, Qwen2.5-7B and Qwen2.5-32B, RAVEL's
object-size task is built from its JSON prompts only, because none of its plain test prompts has two known entities for each
value. Including the left-out tasks would add only tasks in which cross-entropy DAS lies above $T$, or ratios whose
denominator $T$ is near or below zero.

\paragraph{Numerical checks.} Every identity that the measurement relies on is checked by comparing two float32 computations
of the same quantity. The first two comparisons are between the logits computed from the readout and the model's logits, and
between $T$ from the readout and $T$ from the forward pass. The other two are between the effect of whole-state patching at
the readout cell and $T$, and between an unpatched run with attention frozen and the base run. In each comparison, the two
values must agree within the rounding error of a float32 dot product as long as the model's width, for results as large as
the model's logits. This tolerance is $2\gamma_D\,o_{\max}$ for a logit and $8\gamma_D\,o_{\max}$ for an effect, with
$\gamma_D = Du/(1-Du)$, $u = 2^{-24}$ and $D$ the hidden size. Here $o_{\max}$ is the largest $|$label logit$|$ over a sample
of test inputs (up to ten rows per task in the six small models and three in the others; for Qwen2.5-32B, the saved logits of
its readout runs). Because $o_{\max}$ is taken over a sample, it can only be smaller than its maximum over all inputs, so the
tolerance is, if anything, tighter. Pythia-160m's logits lie near 840, where one float32 step is $6\times10^{-5}$. Its
tolerance is therefore the widest, 0.31 log odds for an effect, about a fourteenth of its median $T$. Every deviation lies
far inside the tolerance of its model.

\begin{table}[tbp]
  \caption{\textbf{Results at the readout, per data source.} The main population over the fourteen models. tasks: distinct tasks; $n$: model--task cases; margin: cases with a margin above $T$, where the interval of the out-of-sample maximum lies above $T$; ce $>T$: cases among these where the interval of cross-entropy DAS lies above $T$; ce$/T$: the median over the $n$ cases of the effect of cross-entropy DAS divided by $T$; sq\%: the median distance of the per-pair squared error from $T$, in percent. The rows of IOI split its tasks by their label pairs. With one label pair, every pair contrasts the same two names. With many, the maximum lies below $T$ at rank one (\cref{app:counter}). The cases with a margin above $T$ where cross-entropy DAS is not above $T$ are one-directional negative-polarity constructions, to whose base label the models give little probability (\cref{sec:shortfall}), and Pythia-160m's sentiment classification. In that task, one initialisation lags behind after one pass, and cross-entropy reaches the maximum after ten passes (\cref{app:shortfallfull}).}
  \label{tab:readout}
  \centering
  \footnotesize
  \begin{tabular}{@{}lrrrrrr@{}}
    \toprule
    source & tasks & $n$ & margin & ce $>T$ & ce$/T$ & sq\% \\
    \midrule
    CausalGym & 29 & 404 & 404 & 404 & 2.79 & 2.4 \\
    CausalGym, number only & 6 & 84 & 84 & 84 & 1.56 & 1.1 \\
    Kumon \& Yanaka & 15 & 210 & 210 & 183 & 1.61 & 4.0 \\
    sentiment templates & 9 & 118 & 118 & 117 & 3.35 & 8.4 \\
    SST & 2 & 28 & 26 & 26 & 3.78 & 2.5 \\
    RAVEL & 25 & 270 & 255 & 255 & 2.35 & 3.7 \\
    IOI, one label pair & 1 & 13 & 13 & 13 & 1.80 & 0.5 \\
    IOI, 43 label pairs & 1 & 13 & 0 & 0 & 0.16 & 83.1 \\
    IOI, our frame, one label pair & 1 & 14 & 14 & 14 & 1.82 & 0.4 \\
    IOI, our frame, 44 label pairs & 1 & 14 & 0 & 0 & 0.20 & 79.2 \\
    \midrule
    all & 90 & 1168 & 1124 & 1096 & 2.34 & 3.0 \\
    \bottomrule
  \end{tabular}
\end{table}

\paragraph{Intervals when sentences repeat.} Some test sets reuse sentences across pairs, but mirror-pair resampling treats
these pairs as independent. In \snReuseWord{} of the sentiment templates, the pairs come from \snSentRange{} sentences, and
RAVEL's size tasks reuse bases. As a sensitivity analysis, we resample together every row that shares a base or a source
sentence. This leaves \sbDefined{} of \sbTasks{} cases with at least 20 clusters. The \snReuseWord{} templates are not among
them, so their intervals use mirror pairs only. The sensitivity analysis changes no classification of cross-entropy
against $T$, and no comparison of cross-entropy with the squared error, from above to not above. It changes \sbCeGained{}
classifications and \sbCeSqGained{} comparisons from not above to above, all in RAVEL tasks, where resampling by sentence is
less strict than RAVEL's two-way interval. In addition, \sbSqChangedText. RAVEL's two-way interval is analytic. The variance
of a ratio of sums is $V_A + V_B - V_{AB}$ over the clusters of the entities of the two values and of their pairs
\citep{cameron2011}. Each term is bias-reduced (CR2), and the sum is never below $\max(V_A, V_B)$. The interval uses
Student's $t$ with
the smaller effective number of clusters, less one, as its degrees of freedom.

\paragraph{Pairs shared by training and test.} Every effect that we report is measured on test pairs held out from training,
except in tasks with few templates, where some pairs occur in both sets. Three of Kumon and Yanaka's negative-polarity tasks
(\texttt{NPIDeterminermono}, \texttt{NPIOnlyScope}, \texttt{NPISimple}) also have \kuOverlapDet, \kuOverlapOnly{} and
\kuOverlapSimple{} of their \kuOverlapN{} test pairs in training (twice as many of 100 in their mirrored forms). In
\noOverlapTasks{} of the six number-only agreement tasks, \noOverlapLo{} to \noOverlapHi{} of the \noOverlapN{} test pairs
also occur in training. We rebuilt the two tasks with the most shared pairs, \texttt{NPIDeterminermono} and
\texttt{NPIOnlyScope}, without shared pairs. They give the same classification of cross-entropy in \kuSame{} of the
\kuCases{} model--task cases. In \kuUpWord{} other cases, the classification moves from including $T$ to above $T$, and in
Qwen2.5-1.5B's determiner construction it moves from above $T$ to including $T$.

\paragraph{The example in \cref{fig:law}.} \Cref{fig:law} shows a typical case. We chose its model and task together among
the two-label cases in the \roModelsFOWord{} models whose logits are not capped. We considered only cases whose question can
be followed without linguistics (RAVEL, sentiment, IOI), that are asked as a sentence with a minimal pair, and that have
enough distinct sentences to be scored when rows sharing a sentence are resampled together. Among these, we chose the case
whose cross-entropy DAS lies nearest the median of all \roOneQFO{} two-label cases (\figOneMedian\,$T$). Its
cross-entropy DAS lies at the \figOnePct{}th percentile of all two-label cases. The choice does not take into account how far
DAS falls below the predicted maximum. Here, DAS
reaches \figOneRatio{} of the predicted maximum, the \figOneRatioPct{}th percentile of the same cases (median
\figOneRatioMed, interquartile range \figOneRatioQone--\figOneRatioQthree), and \figOneTen\,$T$ after ten passes. The
out-of-sample maximum is \figOneTopOos\,$T$.

\subsection{Training Losses and the First-Order Approximation}
\label{app:landing}

\paragraph{Cross-entropy ends near its own optimum.} After ten passes, cross-entropy DAS lies above $T$ in \roCeAboveTen{}
of the \roRoom{} cases with a margin above $T$, at \roCeOfTopTen{} of the maximum effect (\roCeOfTop{} after one pass). Its
runs from the three initialisations reach the same direction (median $|\cos|$ \cvStartCosTen{} at ten passes and
\cvStartCosOne{} at one). We also optimised its objective exactly from the top eigenvector of $M$ (\cref{thm:range}), a
starting point unrelated to its training. In each of the \roModelsWord{} models, the median $|\cos|$ between this optimum and
DAS is \cvOptCos{}, and the optimum has a median of \cvOptGain\% more effect than DAS. In the 7--32B models, a tenth of the
tasks gain more than \cvOptGainNinetyLarge\% per model, and none gains more than \cvOptMaxLarge\%. Most of the distance
between DAS and the maximum effect is due to the objective, not to unfinished training. We also applied the first-order
approximation before training to the measure of cross-entropy itself, $\log p(y_s)$. Outside the negative-polarity tasks, it
does not predict the size of this distance. Whether it predicts the direction of this distance cannot be tested. In every
task, at least one of the two directions that such a test would compare, the first-order optimum and the mean-effect direction,
deviates from the top eigenvector of $M$ by an angle whose sine is below 0.05. A deviation this small does not define a
direction. On mirrored test sets, the first-order optimum coincides with this eigenvector. There, a pair and
its mirror cancel the softmax terms, so the first-order value of $\log p(y_s)$ is half that of the log odds, and the same
direction maximises both. The first-order approximation thus predicts no deviation from the top eigenvector, while
cross-entropy DAS deviates from it by a median sine of \ceSinCe{} per model.

\paragraph{The squared error ends near $T$.} With the squared error, three quarters of the \roTasks{} cases lie within
\roSqDevQthree\% of $T$. Cross-entropy DAS lies above the squared error in \roCeAboveSq{} of the \roRoom{} cases with a
margin above $T$. All the other cases are one-directional negative-polarity tasks, and in all but one of them (Llama-3.2-1B's
\texttt{NPIOnly}), cross-entropy DAS does not lie above $T$. The squared error lies farther from $T$ on the sentiment
templates (\cref{tab:readout}) and on IOI with twenty names, where the maximum effect itself lies below $T$. In all
\roNoRoomTasks{} cases whose maximum effect lies below $T$, cross-entropy DAS stays below $T$, as the second counterexample of \cref{app:counter}, in which the maximum lies below $T$, predicts. The data do not show that the squared error reaches the
first-order optimum of its loss instead of $T$.
Where the two differ, $T$ is significantly closer in \roSqTCloser{} tasks per model, although the optimum is closer in more
tasks, \roSqOptCloser{} per model (\roSqOptCloserSum{} against \roSqTCloserSum{} of the \roSqDiffer{} tasks, and
\roSqTCloserGemma{} of the tasks where $T$ is closer are Gemma-2's). The trained direction is far from that optimum (median
$|\cos|$ \roSqOptCos), and the two predicted values are a few percent apart. One pass leaves the squared error short of
convergence. After ten passes, it ends up a median of \roSqMedianDevTen\% from $T$ (three quarters within
\roSqDevQthreeTen\%). On the sentiment templates, which train for a median of \roSentSteps{} steps per pass against
\roOtherSteps{} elsewhere, it ends up \roSqSentTen\% from $T$ after ten passes, against \roSqSentOne\% after one. The divergence
of CDAS ends up at \roJsdMedianTen\,$T$ per model after ten passes.

\paragraph{Checks of the setup.} No direction can exceed the in-sample maximum, which is \roTopExactOfLam{} of $\lambda_1$
(per-model medians in the \roModelsFOWord{} models whose final logits are not capped). The test that no trained run of
cross-entropy DAS lies above $\lambda_1$ therefore asks only whether the first-order approximation describes the in-sample maximum. In Gemma-2, it does not (\cref{app:cap}). In a two-label task whose logits are not capped, the gradient of $z$ with respect to the normalised state is the same for
every pair, up to sign. The readout lies before the final normalisation, whose Jacobian depends on the state, so the readout
gradients share one direction only approximately. Let $s$
be the class sign of a pair. In the \roModelsFOWord{} uncapped models, the shared fraction $\|\E[s w]\| / \E\|w\|$ is at
least \roGradShareMin{} on the test pairs, with gradients at the midpoint between the base and source states. The readout
direction gives $T$ (median \roBhatMedian\,$T$,
interquartile range \roBhatQone--\roBhatQthree\,$T$ over the \roOneQFO{} such cases), as \cref{prop:natural}(ii) predicts.
The angle therefore determines $\lambda_1/\operatorname{tr}M$, so predicting this ratio from the angle tests only whether an
angle measured on training pairs carries over to test pairs. It does, since the angle from the training pairs gives the maximum
effect within a median of \roAngleTopErr\% per model. The margin above $T$ is defined from the maximum effect, and one
of the two starting points of its search is the direction of the mean-effect objective. Whether the mean-effect objective
lies above $T$ therefore mostly measures how well its initialisations agree.

\paragraph{The two forms of the angle.} The angle $\theta$ of \cref{app:quantities} is the angle of \cref{prop:natural} when
the change of the state is a single code. With one shared readout direction, the proposition instead weights each pair's
change by the size of its gradient. Over the \roOneQFO{} two-label cases of the \roModelsFOWord{} models whose logits are not
capped, the two angles differ by a median of \roAngleWeightMedian$^\circ$ and at most \roAngleWeightMax$^\circ$. This
comparison uses the test pairs, while the $\theta$ used for the prediction comes from the training pairs. The prediction from
the angle is undefined at $90^\circ$ or more, so the readout direction from the training pairs is signed toward the change,
which keeps its angle below $90^\circ$. When it is signed by class alone, on the test pairs, no two-label task has an angle
of $90^\circ$ or more (the largest is \roAngleSignedMax$^\circ$).

\paragraph{The trace.} On mirrored test sets, the error of the midpoint trace is $-\tfrac12$ times the error of the
base-point trace (\cref{prop:trapezoid}). \Cref{fig:trace} shows every mirrored task of the uncapped models. Over every mirrored task, Gemma-2's included, wherever both
errors are nonzero, Simpson's combination of the two traces is closer to $T$ than either trace in all but three tasks. The
exceptions are two tasks of Pythia-160m, its sentiment classification and the value form of RAVEL's gender of Nobel
laureates, and Gemma-2's sentiment completion. On sets that are not mirrored, the base-point trace is never significantly
closer to $T$ than the midpoint trace in the small models, and in the other models only in one task of each Llama model. In
Gemma-2, whose final logits are capped, the two errors have opposite signs in \gmTraceOpposite{} mirrored tasks where both
are nonzero (\cref{app:cap}).

\subsection{Held-Out Evaluation}
\label{app:safeguards}

DAS is evaluated on held-out pairs. This tests whether a direction generalises across inputs, not which path its effect
takes. Write the effect of a direction $d$ on the pairs of entity $\epsilon$ as $e_\epsilon(d) = s(d) + r_\epsilon(d)$, where
$s(d) = \E_\epsilon[e_\epsilon(d)]$ is the part shared across entities. Suppose that $\hat d$ is fitted on some entities and
evaluated on new ones drawn from the same distribution. Then $\E[r_\epsilon(\hat d)] = 0$, so the expected held-out effect is
$\E[s(\hat d)]$. Held-out evaluation thus removes the entity-specific part $r_\epsilon$, but not the shared part $s$, which
includes any change of the output that is common to all entities. If a direction raises the attention to the site on every
entity alike, the effect of this rise is part of $s$, and so is a shift toward the upper end of the range that every entity
shares. Both survive held-out evaluation as fully as the part of the effect that whole-state patching also produces. We test
held-out templates at the readout only. There, neither they nor held-out prompts remove this shift. On IOI's held-out
templates, cross-entropy DAS lies above $T$ in all \hoIoiN{} cases of the task with two names. On RAVEL's test prompts, it
lies above $T$ in \hoRavelAbove{} of the \hoRavelRoom{} cases with a margin above $T$, and on the split of Kumon and Yanaka
without shared strings in \hoDisjointAbove{} of \hoDisjointRoom{} such cases. When the test pairs differ from the training
pairs in construction or template, the margin above $T$ itself often disappears. It remains in \hoKumonRoom{} of
\hoKumonN{} cases on the out-of-distribution test set of Kumon and Yanaka, and in \hoSentRoom{} of \hoSentN{} cases on the
sentiment transfer test, which evaluates a direction trained on the movie template on another template. Cross-entropy DAS
ends up at a median of \hoKumonCe\,$T$ and \hoSentCe\,$T$ on these two tests. Where the margin remains, cross-entropy DAS
lies above $T$ in \hoKumonAbove{} of these cases on the first test and in \hoSentAbove{} on the second. Out-of-distribution
tests thus remove the excess mainly where they remove the margin above $T$ itself.

\subsection{Gemma-2's Logit Soft Cap}
\label{app:cap}

Gemma-2 passes its final logits through $\psi(o) = \kappa\tanh(o/\kappa)$ with $\kappa = 30$ \citep{team2024}. At the
midpoint between the base and source states, the raw logits $o$ of the two labels lie at a median depth $|o|$ of
\gmDepth$\kappa$ into the cap, beyond $0.66\kappa$, where $\psi'''$ changes sign. In the text of this subsection, medians are
taken over Gemma-2's \gmCapRankN{} fully mirrored tasks, and in \cref{tab:hcap} over all its tasks. Every result outside this
subsection uses the model as released. With the cap, the angle of \cref{prop:natural} predicts the out-of-sample maximum only
within a median of \gmAngleTopErr\% over Gemma-2's two-label tasks (\roAngleTopErr\% per model where the logits are not
capped). Removing the cap isolates its contribution (\cref{tab:hcap}). The weights, test pairs and hidden states stay the
same, so only the final logits differ.

\begin{table}[tbp]
  \caption{\textbf{Gemma-2 with and without its final logit cap.} The table gives five measures of how well the first-order
  approximation describes the readout, over all of Gemma-2's tasks. It gives each measure with the cap, without it, and as
  the range of the medians of the other three large families. The last column is the share of the gap to that range that
  removing the cap closes. With the cap, at least one trained run of cross-entropy DAS lies above $\lambda_1$ in
  \gmCeilAllAbove{} of the
  \gmCeilAllTasks{} tasks measured both ways. The counts \gmCeilingTasks{} and \gmCeilingTasksBH{} in \cref{app:readout}
  refer to the tasks of the main population, before and after false-discovery control.}
  \label{tab:hcap}
  \centering
  \footnotesize
  \begin{tabular}{@{}lrrrr@{}}
    \toprule
    median over tasks & with the cap & without & other families & gap closed \\
    \midrule
    distance of the midpoint trace from $T$ & \gmDevCap & \gmDevUnc & \gmDevOthers & \gmDevClosed\% \\
    share of tasks with a trained run above $\lambda_1$ & \gmCeilingCap & \gmCeilingUnc & \gmCeilingOthers & \gmCeilingClosed\% \\
    the angle's error on $\lambda_1/\operatorname{tr}M$ ($|\log|$) & \gmAngleCap & \gmAngleUnc & \gmAngleOthers & \gmAngleClosed\% \\
    readout gradients' shared fraction & \gmGradCap & \gmGradUnc & \gmGradOthers & \gmGradClosed\% \\
    readout direction, exact over first order, minus 1 & \gmBhatCap & \gmBhatUnc & \gmBhatOthers & \gmBhatClosed\% \\
    \bottomrule
  \end{tabular}
\end{table}

\paragraph{With and without the cap.} Without the cap, three measures fall inside the range of the other families, and the
other two close most of the gap. The cap therefore accounts for most of Gemma-2's difference, but not for all of it. In the main population, the interval of the in-sample maximum lies above $\lambda_1$ in \gmTopAboveLam{} of \gmTasks{} tasks with the
cap (a median of \gmTopOfLam{} times $\lambda_1$) and in \gmTopAboveLamUncapped{} without it. Without the cap, no trained
run of cross-entropy DAS lies above $\lambda_1$. In the other models, the in-sample maximum lies at $\lambda_1$ or just below, at
a median of \gmOtherTopOfLam{} times $\lambda_1$ per model and at most \gmOtherTopOfLamMax{} times $\lambda_1$.

\paragraph{Mechanism (exploratory).} The final normalisation bends the path of each pair in the space of raw logits. At the
midpoint between the base and source states, the label logits lie a median of \gmBend$\kappa$ further out than the midpoint
of the straight line between their base and source values, in every pair of the median task. They therefore lie where the
cap is flatter. First-order values use the gradient at this point, so they fall below the exact effects. The sign of this
difference follows from a theorem. Suppose that a label's raw logit
stays beyond the inflection point on one side and that the normalisation bends the midpoint outward. Then the capped exact
change is at least as large in size as the raw change times the cap's slope at the bent midpoint (by Hermite--Hadamard for
$\psi'$, then by the bend). Where the source label's raw logit rises and the base label's falls, the cap can therefore only
raise exact effects relative to first order (\cref{app:lean}). Both conditions hold in a
median of \gmCondMedian\% of the (pair, label) cases of a task. From the saved raw logits alone, the cap's share of the
first-order error of the trace is a median of \gmCapShareRaw{} ($\log$ of exact over first order). It would be
\gmCapShareStraight{} if the path were straight. Measured with the real gradients, the share is \gmCapShare{} (rank
correlation \gmCapShareRho{} over tasks). The tanh expansion itself is accurate here. For one label at the median depth that
moves by the median \gmMove$\kappa$, the exact error is \gmTanhExact\% and its leading term is \gmTanhLead\%. The size of the
error comes from the bend of the normalisation combined with the curvature of the cap. We write the log ratio of the in-sample maximum to $\lambda_1$ as $\log(\text{max}/\lambda_1)$. The cap shifts this log ratio
by a median of \gmDeltaMedian, more than it shifts the log ratio of $T$ to the trace (\gmCapShare).
We have not computed what produces the rest of this shift along the path of the maximising direction.

\paragraph{Ranking the cap's shift before training.} Over Gemma-2's \gmCapRankN{} fully mirrored tasks, we measure how far
whole-state patching moves the raw logits of the two labels relative to $\kappa$ (median over test pairs, before any
direction is trained). This quantity ranks the cap's shift of $\log(\text{max}/\lambda_1)$ on the same task (Spearman
\gmCapRank, one-sided permutation $p < 0.001$). The cap raises this log ratio in \gmDeltaPos{} of these tasks. The depth of the
midpoint into the cap, which enters $\lambda_1$ through the cap's slope, ranks the shift more closely (\gmDepthRho,
exploratory). Over the means of \gmFamilies{} label families, the rank correlation is \gmFamilyRho. Within label families,
the permutation test has little power (\gmFamilies{} families for \gmCapRankN{} tasks, $p = \gmWithinFamilyP$), so we cannot
say whether the ranking also holds there. By source, the rank correlation is \gmCGRho{} over the \gmCGN{} CausalGym tasks and
\gmNonCGRho{} over the other \gmNonCGN{} tasks. This difference is within chance (Fisher's $z$, $p = \gmSourceP$).

\subsection{Results for Higher Ranks}
\label{app:rankkfull}

We trained DAS at the readout of all \rkModelsWord{} models at ranks $k = 1, 2, 4, 16$ and $64$. Where a task has more than
two labels, we also trained it at the label rank, the number of labels minus one. Finally, we trained it on the complement of
one trained direction ($k = D-1$). These runs use the pairs, initialisations and intervals of \cref{app:readout}, its
training budgets of one and ten passes, three of its objectives (cross-entropy, the squared error aimed at $T$ and the
mean-effect objective) and $M$ at the midpoint. At $1 < k < D-1$, the parameter is a $D\times k$ matrix. Its first column is
the initial direction of the corresponding rank-one run, and its other columns are Gaussian draws scaled by $1/\sqrt D$.
Every forward pass patches along the QR basis $P$ of this matrix, and Adam updates the matrix. On the complement, the
parameter is one vector $u$, initialised at the initial direction of the corresponding rank-one run, and the patch is $h_b +
(I - \hat u\hat u^\top)\Delta$ with $\hat u = u/\|u\|$. The rank-$k$ maximum is found in the same parametrisation.

\paragraph{The first-order bound.} In the \roModelsFOWord{} models whose logits are not capped, no trained run of
cross-entropy DAS lies significantly above the first-order rank-$k$ maximum $K_k = \lambda_1 + \dots + \lambda_k$, at any
trained rank and with either budget (none of \rkCeilTests{} tests). Under Gemma-2's cap, some trained runs do so at every
rank (\rkCapAbove{} of \rkCapTests{} tests per rank from 1 to 64 and at $D-1$, and \rkCapAboveQ{} of \rkCapTestsQ{} at the
label rank), as at rank one (\cref{app:cap}). This bound holds only to first order. The exact effect of the Ky Fan subspace,
the span of the top $k$ eigenvectors of $M$, lies above $K_k$ in \rkKfAbove{} of \rkKfN{} cases per rank, by at most \rkKfMax\%.

\paragraph{The excess at every rank.} After one pass, cross-entropy DAS lies significantly above $T$ in \rkAboveOne{} of the
\rkRoomOne{} model--task cases with a margin above $T$ at rank one, and at ranks 2, 4, 16 and 64 in \rkAboveTwo{} of
\rkRoomTwo, \rkAboveFour{} of \rkRoomFour, \rkAboveSixteen{} of \rkRoomSixteen{} and \rkAboveSixtyFour{} of
\rkRoomSixtyFour{} cases. After ten passes, the counts at the same five ranks are \rkAboveTenOne, \rkAboveTenTwo,
\rkAboveTenFour, \rkAboveTenSixteen{} and \rkAboveTenSixtyFour. After one pass, cross-entropy DAS also lies above the squared
error aimed at $T$ in \rkCeSqOne, \rkCeSqTwo, \rkCeSqFour, \rkCeSqSixteen{} and \rkCeSqSixtyFour{} of the cases with a margin
above $T$, again in this order. On the complement of one
direction, where the first-order approximation puts the margin above $T$ at $|\lambda_D|$, it lies significantly above $T$ in
\rkAboveDm{} of \rkRoomDm{} cases. Among the cases with a margin above $T$ at both ranks, every case above $T$ at rank one
stays above $T$ at ranks 2 to 64 and at the label rank, with three exceptions. The first is Pythia-410m's JSON prompt
of RAVEL's object colours at rank 64, where the interval over its three entities widens (\rkLostRavel). The other two are
one-directional negative-polarity constructions of the Llama-3.2 models, whose lower bounds come to \rkLostNpi. In the other
direction, \rkGained{} cases per rank move from not above $T$ at rank one to above $T$. The other cases with a margin above $T$
in which DAS is not above $T$ at ranks 2 to 64 or at the label rank are all among the rank-one misses. These misses
are the one-directional negative-polarity tasks and Pythia-160m's sentiment classification (\cref{sec:shortfall}). The margin above $T$
can itself shrink. In \rkTopFalls{} agreement tasks of Pythia-160m, the out-of-sample maximum beyond rank one, which does not
start from the rank-one solution, falls below that at rank one.

\paragraph{Tasks with two labels.} Under one shared direction, a larger rank adds nothing to the maximum (\cref{app:rankk}).
In the two-label tasks of the \roModelsFOWord{} uncapped models, $K_k$ exceeds $\lambda_1$ by a median of \rkKGain\% of $T$
per model at ranks 2 to 64. Per model, the out-of-sample maximum at rank $k$ is a median of \rkTopRatio{} times that at
rank one (\rkTopRatioCap{} in Gemma-2, under its cap). The effect of cross-entropy DAS at rank $k$ is a median of
\rkCeRatio{} times its effect at rank one per model (\rkCeRatioSixtyFour{} at rank 64). The subspace of DAS contains most of the top eigenvector.
The share of the top eigenvector of the training pairs that lies in the subspace of DAS is a median of \rkShareOne{} at rank one and
\rkShareSixtyFour{} at rank 64, and the other columns add no effect.

\paragraph{Tasks with more than two labels.} Without one shared direction, $M$ can have several positive eigenvalues
(\cref{app:rankk}). On IOI
with twenty names (\rkIoiLp{} label pairs), the out-of-sample maximum lies below $T$ at rank one in every model, and
cross-entropy DAS lies at a median of \rkIoiCeOne\,$T$ per model. At ranks 16 and 64 and at the label rank, a margin above
$T$ appears in all \rkIoiCases{} model--task cases, and DAS lies above $T$ in all of them (\rkIoiCeSixteen\,$T$ at rank 16). There, rank one understates the excess.
In the \rkMultiN{} other tasks per model with more than two labels (Kumon and
Yanaka's constructions and agreement over several verbs), $K_k$ on the test pairs grows by a median of \rkMultiGrow\,$T$ from
rank one to rank 16. This growth does not transfer from the training pairs. Over the same step, the Ky Fan subspace of
the training pairs changes the effect on the test pairs by \rkMultiOos\,$T$, the out-of-sample maximum changes by
\rkMultiTop\,$T$, and DAS changes by \rkMultiDas\,$T$ (per-model medians). This holds although every test label pair of the
negative-polarity and agreement tasks also occurs in training.

\paragraph{Effects reached by the objectives.} In the \roModelsFOWord{} uncapped models (per-model medians), cross-entropy
reaches \rkLanding{} of $K_k$ at ranks 1 to 64, as at rank one. The mean-effect objective reaches \rkReach{} of $K_k$ after ten
passes. At the label rank, where $K_k$ includes growth specific to the test pairs, the two objectives reach \rkLandingQ{} and
\rkReachQ. Ten passes add \rkTenPass\% to cross-entropy. The squared error aimed at $T$ ends up at \rkSqT\,$T$ at every rank,
and the exact effect of the Ky Fan subspace is \rkAdequacy{} of its first-order value.

\subsection{Where Cross-Entropy Does Not Exceed \texorpdfstring{$T$}{T}}
\label{app:shortfallfull}

\begin{figure}[tbp]
  \centering
  \includegraphics[width=0.6\textwidth]{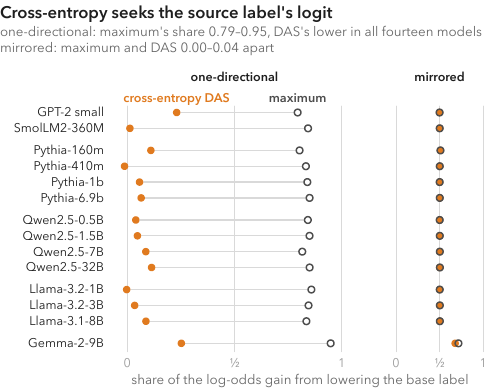}
  \caption{\textbf{Cross-entropy concentrates on the logit of the source label.} The eight negative-polarity constructions of
  \citet{kumon2026a}, the source of \roCeMissNpi{} of the ticks in \cref{fig:readout}a, with medians per model, ordered as in \cref{fig:readout}c. On the left, every base label is \emph{ever}. On the right, the same pairs are mirrored
  as a control, because a pair and its mirror give every direction a share of one half to first order.}
  \label{fig:npidecomp}
\end{figure}

In Pythia-160m's sentiment classification, the runs from the three initialisations reach \roClsSeedsOne\,$T$ after one
pass. After ten passes, cross-entropy reaches \roClsTenT\,$T$, the out-of-sample maximum. The test set is mirrored, so the
direction that is optimal for cross-entropy there also maximises the log odds. On the negative-polarity constructions,
whether cross-entropy lies above $T$ depends on the size of the margin above $T$. The misses have a median out-of-sample maximum of
\npiMissTop\,$T$, and the other cases have \npiHitTop\,$T$ (rank-sum $p = \npiRoomP$). The probability of \emph{ever} does
not separate them ($p = \npiPbP$). On the one-directional pairs, DAS takes a smaller share of its gain in log odds from lowering the logit of the base label
than the maximum does. On the same pairs in mirrored form, this gap closes, and every model's shares lie at \npiMirRest,
except those of Gemma-2-9B, which are \npiMirGemmaTop{} for the maximum effect and \npiMirGemmaCe{} for DAS
(\cref{sec:shortfall}, \cref{fig:npidecomp}). The effect of cross-entropy DAS on these tasks is about the same in the four large models,
\npiCeGainLarge{} log odds. It lies above $T$ in Llama-3.1-8B, Qwen2.5-7B and Pythia-6.9b, where $T$ is \npiTOtherLarge{} log
odds, and falls short of $T$ in Gemma-2, where $T$ is \npiTGemma. In Gemma-2, cross-entropy lies above $T$ on one of the
eight constructions and significantly below $T$ on four. Gemma-2's larger margin, a maximum of \npiTopAbsGemma{} log odds
against \npiTopAbsOtherLarge, comes from lowering \emph{ever} (\npiTopLowerGemma{} against \npiTopLowerOtherLarge), and
cross-entropy DAS does not gain this part of the effect in any model (exploratory).

\subsection{Results by Model Size}
\label{app:scalefull}
\begin{table}[tbp]
  \caption{\textbf{Model size within a family (the scale series).} For each family, the table gives the median over its common tasks of the per-task log ratio $x$, shown as the change in percent that it implies, $100(e^x - 1)$. The changes run from the smallest of its three sizes to the middle one and to the largest, and in Qwen2.5 from 7B to 32B, with 95\% task-bootstrap intervals. The columns give the out-of-sample maximum effect over $T$ (max.), the effect of cross-entropy DAS over $T$ (DAS) and the ratio of the two, the share of the maximum that cross-entropy reaches (share). The test asks that the maximum and DAS grow at both steps (both intervals above zero) and that the share fall to the largest size (its interval below zero). All three hold in every family and in the step to 32B, except for Llama's DAS, whose interval for the first step includes zero. In no family does the maximum or DAS fall from the middle size to the largest by its interval.}
  \label{tab:scale}
  \centering
  \footnotesize
  \begin{tabular}{@{}lrlrlrl@{}}
    \toprule
    & \multicolumn{2}{c}{max.} & \multicolumn{2}{c}{DAS} & \multicolumn{2}{c}{share} \\
    \cmidrule(lr){2-3}\cmidrule(lr){4-5}\cmidrule(l){6-7}
    step & change & interval & change & interval & change & interval \\
    \midrule
    \multicolumn{7}{@{}l}{\emph{Pythia} (84 tasks)} \\
    \quad 410m $\to$ 1b & $+8.3$ & $[+5.2, +11.3]$ & $+5.5$ & $[+2.7, +9.2]$ & $-1.6$ & $[-2.4, -0.5]$ \\
    \quad 410m $\to$ 6.9b & $+23.9$ & $[+20.3, +29.6]$ & $+15.8$ & $[+14.1, +18.0]$ & $-6.0$ & $[-7.8, -4.0]$ \\
    \addlinespace[3pt]
    \multicolumn{7}{@{}l}{\emph{Qwen2.5} (84 tasks; 86 for the step to 32B)} \\
    \quad 0.5B $\to$ 1.5B & $+4.8$ & $[+2.5, +6.9]$ & $+3.3$ & $[+0.6, +6.0]$ & $-1.5$ & $[-3.4, -0.7]$ \\
    \quad 0.5B $\to$ 7B & $+30.1$ & $[+25.1, +33.0]$ & $+18.1$ & $[+13.0, +21.6]$ & $-8.5$ & $[-10.5, -7.6]$ \\
    \quad 7B $\to$ 32B & $+18.0$ & $[+10.6, +22.7]$ & $+8.5$ & $[+5.2, +11.0]$ & $-5.4$ & $[-7.9, -3.5]$ \\
    \addlinespace[3pt]
    \multicolumn{7}{@{}l}{\emph{Llama} (84 tasks)} \\
    \quad 3.2-1B $\to$ 3.2-3B & $+2.4$ & $[+0.6, +4.3]$ & $+1.7$ & $[-0.7, +3.7]$ & $-0.9$ & $[-1.6, -0.3]$ \\
    \quad 3.2-1B $\to$ 3.1-8B & $+41.4$ & $[+34.9, +54.7]$ & $+28.9$ & $[+24.5, +35.4]$ & $-9.2$ & $[-11.5, -7.1]$ \\
    \bottomrule
  \end{tabular}
\end{table}

\begin{table}[tbp]
  \caption{\textbf{Where the growth with size comes from} (descriptive, no intervals). For each data source, the table gives the median per-task change from the smallest to the largest size of each family, in percent, for the maximum effect (max.) and for cross-entropy DAS. The number of tasks per source is given in parentheses. RAVEL's rows pass a filter in each model, so its test rows differ across sizes.}
  \label{tab:scalesources}
  \centering
  \footnotesize
  \setlength{\tabcolsep}{4pt}
  \begin{tabular}{@{}lrrrrrr@{}}
    \toprule
    & \multicolumn{2}{c}{Pythia} & \multicolumn{2}{c}{Qwen2.5} & \multicolumn{2}{c}{Llama} \\
    source & max. & DAS & max. & DAS & max. & DAS \\
    \midrule
    CausalGym & $+29$ (29) & $+20$ & $+39$ (28) & $+29$ & $+74$ (29) & $+52$ \\
    CausalGym number-only & $+18$ (6) & $+15$ & $+22$ (6) & $+17$ & $+42$ (6) & $+34$ \\
    IOI (Wang et al.) & $+24$ (2) & $+11$ & $+20$ (2) & $+3$ & $+17$ (2) & $+6$ \\
    IOI (ours) & $+19$ (2) & $+14$ & $+3$ (2) & $-11$ & $+8$ (2) & $+6$ \\
    Kumon \& Yanaka & $+23$ (15) & $+14$ & $+32$ (15) & $+17$ & $+55$ (15) & $+30$ \\
    RAVEL & $+25$ (20) & $+9$ & $+5$ (20) & $-1$ & $+4$ (19) & $-3$ \\
    SST & $+51$ (2) & $+55$ & $+65$ (2) & $+51$ & $+46$ (2) & $+30$ \\
    Tigges et al. & $+26$ (8) & $+22$ & $+60$ (9) & $+43$ & $+33$ (9) & $+29$ \\
    \bottomrule
  \end{tabular}
\end{table}

Comparing small and large models across families mixes size with training data, tokeniser and architecture. Our
statements about size therefore rest on a test within families. The test of \cref{app:scale} compares three sizes within each family, whose models share one tokeniser. These are Pythia 410m,
1b and 6.9b; Qwen2.5 0.5B, 1.5B and 7B; and Llama 3.2-1B, 3.2-3B and 3.1-8B. In Qwen2.5, the test takes one further step,
from 7B to 32B. Pythia-160m, which is smaller than the three sizes, is drawn on its family's line in \cref{fig:readout}c. On
each task of the common set of a family, we take the maximum effect (the out-of-sample maximum over $T$), cross-entropy DAS
(its effect over $T$) and the share of the maximum that cross-entropy reaches (the ratio of the two). We compare sizes by the
median per-task log ratio, with a task bootstrap. The test asks that a larger model have a larger margin above $T$ and
that cross-entropy DAS report a larger excess in it. For each of these two quantities, the interval of the median log ratio
must lie above zero at both steps up from the smallest size, to the middle size and to the largest. The test also asks that
cross-entropy take a smaller share of the maximum from the smallest to the largest of the three sizes. The step from
Qwen2.5-7B to 32B is held to the same three conditions.

In Pythia and Qwen2.5, all three conditions hold (\cref{tab:scale}). They hold again from Qwen2.5-7B to 32B, on the
\ssThirtyTasks{} tasks that the two models share. There, the maximum grows by a median of \ssThirtyRho\% per task (interval
\ssThirtyRhoLo\% to \ssThirtyRhoHi\%), and cross-entropy DAS grows by \ssThirtyR\% (\ssThirtyRLo\% to \ssThirtyRHi\%). The
share of the maximum that cross-entropy reaches falls by \ssThirtySigma\% (\ssThirtySigmaLo\% to \ssThirtySigmaHi\%). In
Llama, the maximum grows at both steps and the share falls. Cross-entropy DAS grows from the smallest to the largest size,
but from 1B to 3B it grows by only \ssLlamaRStep\% (interval \ssLlamaRStepLo\% to \ssLlamaRStepHi\%), an interval that
includes zero. From the smallest to the largest of the three sizes, the maximum grows by a median of \ssPythiaRho\% per task
in Pythia, \ssQwenRho\% in Qwen2.5 and \ssLlamaRho\% in Llama. Cross-entropy DAS grows by \ssPythiaR\%, \ssQwenR\% and
\ssLlamaR\%, and the share falls by \ssPythiaSigma\%, \ssQwenSigma\% and
\ssLlamaSigma\% (in \ssPythiaSigmaFalls{} of \ssPythiaTasks{}, \ssQwenSigmaFalls{} of \ssQwenTasks{} and \ssLlamaSigmaFalls{}
of \ssLlamaTasks{} tasks). \Cref{tab:scalesources} gives the same changes for each data source. The predicted maximum,
$\lambda_1/T$, which is computed before training, grows at every step, with a median over the common tasks of \ssLamPythia{}
in Pythia, \ssLamQwen{} in Qwen2.5 (\ssLamThirty{} at 32B) and \ssLamLlama{} in Llama.

\paragraph{Limits.} The test has three limits. First, the step between the two smaller sizes of Qwen2.5 and Llama, whose
embeddings are tied, is small and rests on CausalGym's tasks; without these tasks, the growth at that step does not hold.
Second, most of the growth comes with the largest of the three sizes, where the embeddings are untied and, in Llama, the
model is the one from which the two smaller models were pruned and distilled. These steps do not separate size from those
changes. However, Pythia, which never ties its embeddings, grows at about the same rate per doubling at both steps, and
Qwen2.5 still grows from 7B to 32B, with untied embeddings at both sizes. Third, the growth is uneven across sources
(\cref{tab:scalesources}), and RAVEL, whose test rows pass a filter in each model, shows almost no growth in Qwen2.5 and
Llama.

\FloatBarrier
\section{Details and Further Results Inside the Network}
\label{app:insidefull}

This appendix describes the maps behind \cref{sec:offroute}, with their cells, counts and controls. It also reports every
further test of the findings of that section.

\subsection{Results on the Maps of the Six Small Models}
\label{app:fullgrid}

\paragraph{Cells and quantities.} The full grid holds \inTasks{} tasks in the six small models. These are CausalGym's 29
tasks and 49 tasks from the other sources (Kumon and Yanaka's constructions, number-only agreement, RAVEL, IOI, SST and the
sentiment templates). The population rule of \cref{sec:tasks} leaves out \inExcludedMapsWord{} maps, and \inMaps{} remain. The maps left out are
those of sentiment templates with fewer than 20 test clusters and those of \texttt{filler\_gap\_embed\_4} in Qwen2.5-0.5B and
Pythia-160m, whose interval for $T$ does not lie above zero. Every
count excludes the \inDegenerate{} cells whose region is empty in some row, where the patch would take the token of another region. A cell is live when, in some
pair, base and source differ in at least one token up to that cell. At each live cell, we measure on test pairs the effect of
whole-state patching, $G$, and its value in the best context, $G^*$, with the context chosen on the training pairs. A cell is
off-route (\cref{alg:route}) when the interval of DAS lies above zero, the interval of $G^*$ does not, and DAS lies significantly above $G^*$.
There are \inOffRoute{} such cells. At \inOffRouteIndet{} more cells, the interval of DAS lies above zero and that of $G^*$
does not, but the difference between DAS and $G^*$ is not significant. We also measure cross-entropy DAS and the squared error aimed at $G$ from three initialisations, the mean-effect
objective, the squared error aimed at $T$ and the CDAS divergence from one initialisation, and the untrained directions. Each is
measured twice, once with attention free to change and once with every downstream attention pattern frozen to that of the
base run. The first-order range comes from the midpoint gradients of the test pairs, computed before any training. Intervals
resample mirrored pairs together. Correlations are averaged across maps, which serve as the units. Cell counts use per-cell
intervals after false-discovery control over the cells of each model (Benjamini--Hochberg, $q = 0.025$, with $p$-values from
the standard error of the resampling). The exception is the off-route cells, which are counted on raw intervals (\inOffRoute,
and \inOffRouteBH{} after the control). On raw intervals, the counts are \inCeAboveGRaw{} cells above $G$,
\inCeAboveGstarRaw{} above $G^*$ and \inCeilRaw{} above the predicted maximum, against \inCeAboveG, \inCeAboveGstar{} and
\inCeilBH{} after the control.

\paragraph{A stronger reference.} Where $G^*$ uses a second region, it exceeds $G$ by a median of \inCtxGainMed\,$T$ (90th
percentile \inCtxGainQnine\,$T$), so the second region adds little. With that region chosen on the test pairs, which can only
raise $G^*$, DAS lies above $G^*$ at \inCeAboveGstarTest{} of the \inLive{} live cells, against \inCeAboveGstar{} with the region
chosen on the training pairs. We also measured DAS with the whole state of the second region of $G^*$ patched at the same
time, and we subtracted the effect of patching that region alone, as for $G^*$. Where $G^*$ has no second region, DAS was
measured alone. In this measurement, DAS lies above $G^*$ at \inAboveGstarCtx{} cells, and the share of its effect beyond
$G^*$ is \inNpCtx{} per model. The doubt does not arise at a complete cut, where there is no choice of positions, because
inflation at the readout is measured against $T$ itself.

\paragraph{The off-route cells.} At the \inOffRoute{} off-route cells, the net effect that whole-state patching measures is
not above zero, even in the best context, and at \inOffRouteBelowZero{} of these cells the interval of $G^*$ lies below zero.
The upper bound of the interval of $G^*$ has a median of \inOffRouteGstarHi\,$T$, and nine in ten of these upper bounds lie
below \inOffRouteGstarHiNinety\,$T$. As the definition requires, DAS lies significantly above $G^*$ at these cells
(\inOffRouteBH{} of them after false-discovery control).

\subsection{The Large Models}
\label{app:largefull}

The four large models (Llama-3.1-8B, Gemma-2-9B, Qwen2.5-7B and Pythia-6.9b) were mapped on three of CausalGym's tasks,
negative polarity, filler--gap and subject--verb agreement, at every second layer and at the readout. Their results are
close to those in the small models. Three maps per model are too few for a test, because a bootstrap's 2.5\% quantile over
three maps would be the smallest map itself. We therefore describe these maps without a test across maps, and we compare
them with the small models on the same tasks. DAS lies significantly above $G$ at \inBigCeAboveG{} of the \inBigLive{} live
cells (with false-discovery control per model), against \inThreeAboveG{} of \inThreeLive{} in the small models. Per model,
Gemma-2 included, \inBigNotProduced{} of the effect of DAS summed over live cells lies beyond $G^*$, the effect of
whole-state patching in its best context, and \inBigNotProducedUnc{} in the three models whose logits are not capped. The
share in the small models is \inThreeNotProduced{} per model. Of the part beyond $G^*$, \inBigThroughAttn{} runs through attention in
the uncapped models (\inThreeThroughAttn{} in the small models). Gemma-2, whose logits are capped, gives \inBigGemmaNotProduced{} and
\inBigGemmaThroughAttn. It is left out of the shares of the uncapped models, but the count of cells above $G$ includes it.

\subsection{The First-Order Approximation Inside the Network}
\label{app:firstorderfull}

\paragraph{Where the first-order approximation holds.} The midpoint trace is a median of \inTraceMid{} times $G$ per model,
and it is closer to $G$ than the base-point trace at \inTraceMidCloser\% of cells. Its per-layer medians lie close to $G$ in the
first layers and near the readout and exceed it most in between, by at most \inTraceOvershoot\% per model. Untrained directions have exact
effects within a few percent of their first-order effects (per-model medians \inFoUntrained), whereas cross-entropy DAS has
\inFoDas{} times its first-order effect, so, inside the network, training reaches effects beyond first order (exploratory).
Cross-entropy DAS exceeds the predicted maximum at \inCeilBH{} cells (the run from any of its three initialisations, with a
Bonferroni correction over the three runs and then false-discovery control). At these cells, the median $G^*$ is
\inCeilGstarMed\,$T$, and freezing attention removes \inCeilAttn{} of the effect of DAS.

\paragraph{Ranking the excess before training.} Across the full grid, the first-order excess $(\lambda_1 -
\operatorname{tr}M)/T$ and the other predictors are computed on the resampled pairs, and the excess of DAS, $(e - G)/T$, on
the pairs left out. Each rank correlation is averaged over the resamples. The mean rank correlation of the first-order excess
is \gridThRho{} over the \gridMaps{} maps. Pooled over the maps, its rank correlation exceeds that of each of seven untrained predictors by at least \gridMinusSingleLo{} at the lower bound. The seven are the layer, $G$, $-G$, the mean size of the effect of whole-state
patching per pair, $\E|G_i|$, that size minus $G$, $G^* - G$, and the effect of the mean difference minus $G$. The best
single predictor, \gridBestSingleName, reaches \gridBestSingleRho. If the best of the seven is chosen afterwards on each map,
the chosen predictor differs from map to map, and its correlation is an upper bound for that of any single one of the seven.
This per-map best reaches \gridBestSevenRho, and the first-order excess does not exceed it. The difference is
\gridRankSeven{} [\gridRankSevenLo, \gridRankSevenHi] over the \gridConstr{} constructions and \gridRankSevenMap{}
[\gridRankSevenMapLo, \gridRankSevenMapHi] over maps, and per model the first-order excess falls significantly below this
bound in \gridModelsBelowWord{} of the six models. The sources differ. On CausalGym's 29 tasks, the first-order excess has a higher rank correlation than the per-map best (\gcgRankSeven{} [\gcgRankSevenLo, \gcgRankSevenHi] over \gcgConstr{} constructions). Its mean
correlation there is \gcgThRho, against \gcgBestSingleRho{} for the best single predictor and \gcgBestSevenRho{} for the
per-map best. On the other sources taken together, it falls below the per-map best (\gotherRankSevenC{} [\gotherRankSevenCLo,
\gotherRankSevenCHi]). For each source separately, the comparisons are over maps and descriptive. The first-order excess has a higher rank correlation than the per-map best on IOI (\gridSrcIoi) and number-only agreement (\gridSrcNumOnly), and it performs on par with it
on Kumon and Yanaka's constructions (\gridSrcKumon). It falls below the per-map best on RAVEL (\gridSrcRavel), the sentiment
templates (\gridSrcSent) and SST (\gridSrcSst). On RAVEL, where its correlation is \gridRavelRho, the layer alone ranks the
excess better. At the live cells below the readout that lie before the last position, the norm of the input change, alone and
divided by the layer, can be added to the predictors, which makes nine. There, the correlation of the first-order excess is \gridDoorRho{} (\gcgDoorRho{} on CausalGym's
tasks). It lies above the correlation of the norm by \gridMinusD{} [\gridMinusDLo, \gridMinusDHi], but below the best of nine
predictors chosen on each map (\gridRankDoor{} [\gridRankDoorLo, \gridRankDoorHi]). The ranking by the first-order excess
therefore fails at these cells. Attention does not explain this failure. The first-order excess ranks these cells better in
maps where freezing attention removes more of the effect of DAS at them (Spearman \gridAttnRho{} across maps). We include $G$
among the predictors instead of computing a rank correlation that controls for $G$. Such a correlation, computed on the same
pairs by residualising the ranks on those of $G$, is biased under its null hypothesis.

\subsection{The Attention to the Site}
\label{app:door}
\label{app:proofs-door}

\begin{proof}[Proof of \cref{prop:door}]
(i) To first order a patch $v$ at $p$ changes the score by $a^\top v$. Whole-state patching is $v = \Delta$, and the
rank-one interchange along $d$ is $v = dd^\top\Delta$, which gives $(a^\top d)(d^\top\Delta) = d^\top\operatorname{sym}(a\Delta^\top)\,d$.
Its maximum over unit $d$ is the top eigenvalue of $\operatorname{sym}(a\Delta^\top)$, which is
$\tfrac12(a^\top\Delta + \|a\|\|\Delta\|)$ by the proof of \cref{thm:range}(iv) with $a$ in place of $w$. Its eigenvector is
$\hat a + \hat\Delta$, since $\operatorname{sym}(a\Delta^\top)(\hat a + \hat\Delta) = \tfrac12\|a\|\|\Delta\|(1 + \hat a^\top\hat\Delta)(\hat a +
\hat\Delta)$, so the maximum is at the bisector (if $\hat a = -\hat\Delta$ the maximum is $0$ and there is no
bisector).

(ii) Let $s$ be the score on $p$ and $Z$ the sum of the exponentials of the other scores, so that $\alpha = e^s/(e^s + Z)$ and
$\log\tfrac{\alpha}{1-\alpha} = s - \log Z$. With $Z$ fixed, a score change $\delta$ gives the weight
$\sigma(\log\tfrac{\alpha}{1-\alpha} + \delta)$, which is increasing in $\delta$. Since $\log\alpha = -L$, $\delta \le L - t$
gives a weight of at most $\sigma(-t - \log(1-\alpha))$, and $\delta \ge L + t$ a weight of at least
$\sigma(t - \log(1-\alpha)) \ge \sigma(t)$. Whole-state patching has $\delta = a^\top\Delta$, and the rank-one interchange at the
maximum of (i) has $\delta = \tfrac12(a^\top\Delta + \|a\|\|\Delta\|)$. With $a^\top\Delta = 0$ the first condition holds
whenever $t \le L$, and the second reads $\tfrac12\|a\|\|\Delta\| \ge L + t$.
\end{proof}
The statement is first order through the LayerNorm of the site, so it is accurate only while $\|\Delta\|$ is small compared
with the centred state. The attention can rise within this range, because the condition of \cref{prop:door}(ii) compares the change of the score,
$\|a\|\|\Delta\|$, with the negative log weight $L$, not with the state. For larger differences, whole-state patching can also
change a score that it leaves unchanged to first order, and the largest first-order change of the score overstates how much
the attention weight rises.

\paragraph{Where the attention to the site rises.} In the first block that reads the site, the first-order description of \cref{prop:door}
holds (median correlations \inDoorFirstOrder{} per model), as expected for the first-order form of a single block. Outside IOI's
maps, this block contains the largest rise of attention to the site at \inDoorBlockLShare\% of the live cells before the last position,
per model; at the other cells, the largest rise comes in a later block. Before training, we compute for each cell a predictor of where this rise occurs, namely the largest first-order change in the
head's score (\cref{prop:door}(i)) minus $L$. Within each map, on the same cells and wherever the interval of DAS excludes
zero, we rank-correlate this predictor with the share of the effect of DAS that runs through attention, that is, the share
that freezing removes.
This correlation does not exceed the highest correlation of three simple predictors with the same share, namely the norm of
the change of the state, the attention weight and $-G$. The first-order description therefore predicts where the effect of
DAS runs through attention no more closely than these simple predictors do.

\subsection{The Summed Excess and Its Share Through Attention}
\label{app:routefull}

Where DAS lies significantly above $G$, it does so by a median of \inExcessMed\,$T$ (quartiles
\inExcessQone--\inExcessQthree\,$T$). We sum the excess of DAS over $G^*$ across the live cells where this excess is
positive. Per model, this sum is \inNpPos{} of the effect of DAS summed over live cells, and the network itself does not
produce this part. The share is \inNpSigned{}
when cells where DAS lies below $G^*$ count against it, and \inNpCons{} when we count only cells whose excess over $G^*$ is
significant. The per-model 95\% intervals of all three shares lie within \inNpBandLo--\inNpBandHi. Half of the excess summed
in the first way lies at cells whose $G^*$ is below \inHalfExcessGstar\,$T$. The smaller $G^*$ is at a cell, the larger the share
of that part that runs through attention (rank correlation of $G^*$ with the share that freezing removes, \inAttnRho{} per model).
The part of each cell is weighted by the share of the whole effect of DAS, not of its excess, that freezing removes. This
share is clipped to $[0, 1]$ and set to zero where the interval of DAS includes zero. This weighting gives the lower of two
estimates, by which \inThroughAttnOfNotProduced{} of the excess runs through attention per model. In the other estimate, we compare each cell's excess over $G^*$ with and without freezing, with $G^*$ held fixed and
the reduction clipped to $[0,\,$excess$]$. Freezing then removes \inAttDirect{} of the excess per model. How much of the
effect runs through attention depends on the task. Freezing removes a median of \inSdasNpiFg{} of the effect of DAS per model
on negative polarity and filler--gap. In every model, this is more than on agreement (\inSdasAgr), both among CausalGym's
tasks and when Kumon and Yanaka's constructions are compared with number-only agreement. On the other sources, freezing
removes \inSdasOther. At the off-route cells, freezing only the attention row of the readout position removes
\inOffRowRemoved{} of the summed effect of DAS.

\subsection{Details of the Training-Target Experiment}
\label{app:targetfull}

As in \cref{app:fullgrid}, cross-entropy and the squared error aimed at $G$ are trained from three initialisations, and the
squared error aimed at $T$ from one (descriptive; \cref{fig:guidef} shows them at one cell). Over the live cells, we fit each
effect against $G$ by least squares. Cross-entropy has a slope of \inCeSlope{} and an intercept of \inCeIntercept\,$T$. The
squared error aimed at $T$ has \inSqTSlope{} and \inSqTIntercept\,$T$, and the squared error aimed at $G$ has \inSqGSlope{}
and \inSqGIntercept\,$T$. The untrained mean difference, outside IOI, has \inUhatSlope{} and \inUhatIntercept\,$T$. Cross-entropy
lies above $G$ however large $G$ is; the squared error aimed at $T$ lies above $G$ while $G$ is below $T$ and meets it as $G$
nears $T$; the last two follow $G$.

\paragraph{The other objectives where whole-state patching shows no effect.} Inside the network, every objective that we
trained to make the output follow the source gives an excess (all but cross-entropy from one initialisation, reported as a
description), while the squared error aimed at the cell's own $G$ ends up near $G$. We consider the \inZero{} live cells where
the interval of whole-state patching in its best context does not lie above zero. At these cells, DAS trained with the
squared error aimed at $T$ reaches a median of \inZeroSqT\,$T$, lies significantly above $G^*$ at \inZeroSqTAboveGstar{} cells
and loses \inZeroSqTAttn{} of its effect when attention is frozen. With the CDAS divergence, DAS reaches a median of
\inZeroJsd\,$T$, lies significantly above $G^*$ at \inZeroJsdAboveGstar{} cells and loses \inZeroJsdAttn{} of its effect. The
route test (\cref{alg:route}) also requires the interval of the learned effect itself to lie above zero. By this test, DAS is
off-route at \inZeroSqTOff{} of these cells with the squared error, at \inZeroJsdOff{} with the CDAS divergence and at
\inOffRoute{} with cross-entropy. At all but \inZeroAtT{} of these cells, the effects of the squared error and of the CDAS
divergence both lie below $T$ (the largest is \inZeroMax\,$T$), so a target or bound at $T$ rarely constrains them there. In
contrast, whole-state patching has an effect near zero at every one of these cells, so the comparison with it marks them
where a bound at $T$ does not.

\FloatBarrier
\section{Re-examining the Published Results}
\label{app:publishedfull}

This appendix gives the details behind \cref{sec:published}. It covers the rebuilt CausalGym maps, the re-run of CausalGym's
corrected code and the re-run of the sentiment direction.

\subsection{Rebuilding CausalGym's Maps}
\label{app:cgmapfull}

\paragraph{The rebuilt maps.} We rebuilt CausalGym's maps for its 29 tasks in the six small models, at every layer
(\cgmMaps{} maps, \cref{app:sites}). We follow the benchmark's protocol, that is, the rows as published, the first token of
each label, one epoch, batch size 4, Adam at $5\cdot10^{-3}$ and cross-entropy of the source label. We use three
initialisations, a constant learning rate, float32 and a renormalised direction. CausalGym's paper instead uses one
initialisation, a linear schedule with warmup, bfloat16 at 1b (float32 up to 410m) and pyvene's orthogonally parametrised
rotation. In the three models with published values, we compare with these values on the ten tasks that the released code reads
correctly (\cref{app:indexswap}). Our vanilla score equals the published one to its two decimals in \cgmPubVanEq{} of \cgmPubN{} cases and differs
by 0.01 in the rest. In \cgmPubDasIn{} cases, the published DAS value lies within the 95\% resampling range of our aggregate
for a single initialisation, with the resamples of the three initialisations pooled.
Pythia-1b's negative-polarity DAS map matches Figure~5(b) of CausalGym's paper layer by layer (Spearman \cgmPubRho, against
\cgmPubRhoShift{} with the layers offset by one).

\suppressfloats[t]   %
\begin{table}[tbp]
  \caption{\textbf{CausalGym's maps rebuilt under its protocol, per model} (29 tasks, every layer). $\bar R$: CausalGym's
  aggregate, DAS divided by vanilla, as a median over tasks. The gap is the aggregate of DAS minus that of vanilla, summed
  over layers. Shares are ratios of sums over the tasks of a model, with intervals that resample constructions (a task
  together with its \texttt{\_mod} twin). Beyond $G^*$: the excess of DAS over $G^*$, the effect of whole-state patching in its best context (chosen on the test pairs), at the best region of DAS in each layer, as a share of the gap. $G^*$ not above 0: the
  share at cells where the interval of $G^*$ does not lie above zero. Off-route: the same share, counted only where DAS also
  lies significantly above $G^*$. Through attention: the median share of the excess of DAS that freezing attention removes.
  Published: of the ten tasks that the released code reads correctly, the tasks where vanilla equals the published value to
  its two decimals, and the tasks where the published DAS lies within the 95\% resampling range of our single-initialisation
  aggregates; -- where CausalGym publishes no values for the model. $^a$One task, \texttt{filler\_gap\_embed\_4}, whose $T$ is $\cgmPythiaFgT$, dominates both sums of
  Pythia-160m. The same task holds \cgmQwenFgPart{} of Qwen2.5-0.5B's \cgmSoffQwen{} (\cgmSoffQwenNoFg{} without it). Shares
  are sums in units of each map's $T$, so a map with a small $T$ weighs heavily. Per map, the median share beyond $G^*$ is
  \cgmMapBeyondLo--\cgmMapBeyondHi, and the share where $G^*$ is not above 0 has median 0 in every model.}
  \label{tab:causalgymmap}
  \centering
  \footnotesize
  \setlength{\tabcolsep}{4pt}
  \begin{tabular}{@{}lrrrrrr@{}}
    \toprule
    model & $\bar R$ & beyond $G^*$ & \begin{tabular}[b]{@{}c@{}}$G^*$ not\\above 0\end{tabular} & off-route & \begin{tabular}[b]{@{}c@{}}through\\attention\end{tabular} & published \\
    \midrule
    GPT-2 small & 1.98 & 1.03 [1.00, 1.06] & 0.01 [0.00, 0.05] & 0.01 [0.00, 0.05] & 0.40 & -- \\
    Pythia-160m & 2.18 & 1.08 [0.95, 1.09] & \multicolumn{2}{c}{not established$^a$} & 0.30 & 9/10, 9/10 \\
    Pythia-410m & 2.05 & 1.02 [0.99, 1.04] & 0.02 [0.00, 0.04] & 0.02 [0.00, 0.04] & 0.21 & 10/10, 10/10 \\
    Pythia-1b & 2.24 & 1.02 [0.99, 1.04] & 0.02 [0.00, 0.04] & 0.02 [0.00, 0.04] & 0.37 & 6/10, 9/10 \\
    SmolLM2-360M & 2.00 & 1.02 [1.01, 1.04] & 0.02 [0.00, 0.04] & 0.02 [0.00, 0.04] & 0.36 & -- \\
    Qwen2.5-0.5B & 2.09 & 1.03 [1.02, 1.07] & 0.40 [0.00, 0.52] & 0.00 [0.00, 0.01] & 0.33 & -- \\
    \bottomrule
  \end{tabular}
\end{table}

\begin{figure}[tbp]
  \centering
  \includegraphics[width=\textwidth]{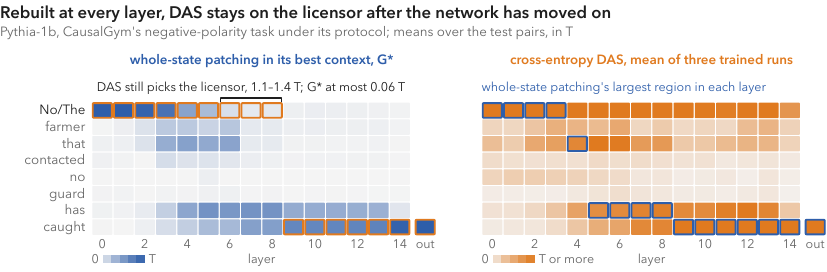}
  \caption{\textbf{On the map rebuilt at every layer, DAS stays on the licensor in layers where whole-state patching at the
  licensor has almost no effect.} Words run down and layers across, and each cell's shade gives its effect on one scale from 0
  to $T$. The outlines mark the best region of DAS in each layer on the left and whole-state patching's largest region on the right.}
  \label{fig:atlas}
\end{figure}

\paragraph{Where the gap between DAS and vanilla comes from.} CausalGym's aggregate puts DAS above vanilla in every rebuilt map, by a
median factor of \cgmRbarMed{} (quartiles \cgmRbarQone--\cgmRbarQthree). The interval of this factor lies above one in
\cgmRbarAboveOne{} of \cgmMaps{} maps, and per model the median factor is \cgmRbarModelLo--\cgmRbarModelHi{}
(\cref{tab:causalgymmap}). On the negative-polarity map, the factor is \cgmNpiRbar{} in Pythia-1b (\cgmNpiPub{} published)
and \cgmGptNpiRbar{} in GPT-2. We define the \emph{gap} between DAS and vanilla as the sum over layers of the effect of the best
region of DAS minus the best effect of vanilla. At the best region of DAS in each layer, $G^*$ gives about what the best region of vanilla
gives, so the gap lies beyond $G^*$. Per model, the excess of DAS over $G^*$ is \cgmSNlo--\cgmSNhi{} times the gap when the context of $G^*$
is chosen on the test pairs. We choose it there because CausalGym's aggregate picks the best region of each method on the pairs
that it scores. A context chosen on those pairs can only raise $G^*$, so this ratio is the lower one. On the maps
where the gap is established (all but Qwen2.5-0.5B's and Pythia-160m's \texttt{filler\_gap\_embed\_4}), the ratio is
\cgmSNsamelo--\cgmSNsamehi{} with that context and \cgmSNtrlo--\cgmSNtrhi{} with the context chosen on the training pairs, as
in \cref{sec:offroute}. Little of the gap lies at cells where the interval of $G^*$ does not lie above zero. In four models,
at most \cgmSoffSmallMax{} of it lies there. In Qwen2.5-0.5B, the share is \cgmSoffQwen, almost all of it from one task whose
$T$ is \cgmQwenFgT, and it is \cgmSoffQwenStrict{} at off-route cells, where DAS also lies significantly above $G^*$. In
Pythia-160m, one task with $T = \cgmPythiaFgT$ dominates both sums, and the share is not established. Pooled over the six
models, the share is \cgmSoffPooled{} [\cgmSoffPooledLo, \cgmSoffPooledHi]. It is \cgmSoffPooledNoFg{} without
\texttt{filler\_gap\_embed\_4}, whose small $T$ weighs heavily in the sums, and \cgmSoffPooledStrict{}
[\cgmSoffPooledStrictLo, \cgmSoffPooledStrictHi] where DAS also lies significantly above $G^*$. Freezing attention removes a
median of \cgmAttLo--\cgmAttHi{} of the excess of DAS per model, and \cgmNpiAtt{} on Pythia-1b's negative-polarity map.

\paragraph{How the route test reads CausalGym's map.} The path that CausalGym reads from its map lies on the route of the map, and it follows the cells where whole-state patching
has its largest effect. CausalGym writes that ``the negation feature is moved to the complementiser \emph{that} in the early layer, into the
auxiliary verb at middle layers, and the main verb in later layers'' \citep[\S6]{arora2024}. On the rebuilt Pythia-1b map,
whole-state patching has its largest effect at the licensor through layer \cgmRouteLicLast, at \emph{that} at layer
\cgmRouteThat, at the auxiliary \emph{has} to layer \cgmRouteHasLast{} and at the main verb from layer \cgmRouteCaughtFrom{}
onward (\cref{fig:atlas}, right). The DAS map also shows one row that this reading does not use, the licensor row (\emph{No}/\emph{The}). The best region of DAS stays on the licensor
to layer \cgmNpiLicLast, and from layer \cgmNpiFrom{} it reports \cgmNpiDasLo--\cgmNpiDasHi\,$T$ there, where whole-state
patching, even in its best context, gives at most \cgmNpiGsMax\,$T$ (\cref{fig:atlas}). Freezing attention removes
\cgmNpiFrozenRemoves\% or more of this effect. The route test does not mark these cells as off-route, because the interval of
$G^*$ there lies above zero, although $G^*$ is small. It classifies them instead as cells where the effect of DAS lies far
above $G^*$, so that almost all of this effect is excess. The row also appears in the
published map, where CausalGym's Figure~5(b) shades it through the late layers.

\subsection{Token Positions in the Released Code}
\label{app:indexswap}
\label{app:indexswapfull}

\begin{figure}[tbp]
  \centering
  \includegraphics{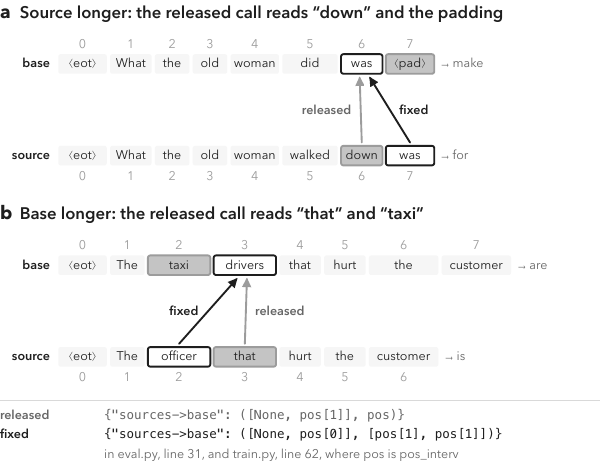}
  \caption{\textbf{What the released call reads, and the fix.} Two training pairs drawn by CausalGym's own sampler, in Pythia's
  tokeniser (\cref{app:indexswap}). Outlined boxes mark each sentence's last token of the intervened region, which the fixed call reads. Grey boxes
  mark what the released call reads instead, which is the state that it patches in for the source and, for the base, the
  state that it collects, on which the representation-only methods are fitted. After each sentence, its label. In the code, \texttt{pos[0]} holds the token positions of the source
  sentence and \texttt{pos[1]} those of the base sentence, one position per sentence, at the last token of the
  intervened region.}
  \label{fig:indexswap}
\end{figure}

CausalGym's released code passes the token positions of base and source to its interventions in swapped order
(\cref{fig:indexswap}). This has no effect where the two sentences have the same number of tokens up to the end of the
intervened region, which holds on all ten negative-polarity and filler--gap tasks. On the other 19 tasks, it fails for
12--100\% of the training pairs, and there an intervention reads a token other than the region's last. With the two positions
in order, the call reads each sentence at the last token of the region on every pair, and on the ten tasks it changes
nothing.

\Cref{tab:causalgym} divides every published score by vanilla, at each of the nine published sizes. Over all 29 tasks, the
score of DAS lies above vanilla, beyond the rounding of the tables, in \cgDasAbove{} of \cgEntries{} entries (one entry per
task and size), while the scores of the other methods lie above vanilla in \cgOtherAbove{} of their \cgOtherEntries{}
entries.

\begin{table}[tbp]
  \caption{\textbf{CausalGym's published scores divided by the score of its vanilla intervention} (whole-state patching scored in
  the same way, in the same units). The scores come from CausalGym's per-size tables \citep[App.~E.2]{arora2024}.
  The table gives medians over its 29 tasks at all nine published sizes. Of its seven methods, the probe has two settings,
  and the three smallest sizes publish only one of them (--). The last row gives the number of tasks per size in which DAS
  lies above vanilla beyond the rounding of the tables. Every value has two decimals, so a difference is known to
  $\pm 0.01$. The other six methods lie above vanilla in \cgOtherAbove{} of their \cgOtherEntries{} method--task--size
  entries, by a median of \cgOtherExceedMed$\times$. On the ten tasks that the released code reads correctly
  (\cref{app:indexswap}), they do so in \cgOtherAboveAligned{} of \cgOtherEntriesAligned{} entries. Over the entries with
  vanilla $\ge 0.5$ alone, the medians of DAS are \cgFiltDasMedLo--\cgFiltDasMedHi, and the other methods lie above vanilla
  in \cgFiltOtherAbove{} of \cgFiltOtherEntries{} entries.}
  \label{tab:causalgym}
  \centering
  \footnotesize
  \begin{tabular}{@{}lrrrrrrrrr@{}}
    \toprule
    method / vanilla & 14m & 31m & 70m & 160m & 410m & 1b & 1.4b & 2.8b & 6.9b \\
    \midrule
    DAS & 3.06 & 2.83 & 3.08 & 2.41 & 2.09 & 2.38 & 2.08 & 2.10 & 2.47 \\
    probe, first setting & 0.94 & 0.98 & 0.97 & 0.98 & 0.96 & 0.95 & 0.90 & 0.88 & 0.85 \\
    probe, second setting & -- & -- & -- & 0.78 & 0.95 & 0.89 & 0.74 & 0.88 & 0.65 \\
    mean difference & 0.87 & 0.72 & 0.78 & 0.65 & 0.84 & 0.78 & 0.77 & 0.70 & 0.71 \\
    PCA & 0.44 & 0.43 & 0.43 & 0.47 & 0.56 & 0.46 & 0.48 & 0.45 & 0.41 \\
    $k$-means & 0.46 & 0.42 & 0.43 & 0.47 & 0.57 & 0.48 & 0.50 & 0.47 & 0.41 \\
    LDA & 0.07 & 0.04 & 0.03 & 0.03 & 0.04 & 0.04 & 0.02 & 0.02 & 0.03 \\
    random & 0.02 & 0.01 & 0.01 & 0.00 & 0.01 & 0.00 & 0.00 & 0.00 & 0.00 \\
    \midrule
    DAS above vanilla & 26/29 & 28/29 & 29/29 & 28/29 & 26/29 & 29/29 & 29/29 & 29/29 & 27/29 \\
    \bottomrule
  \end{tabular}
\end{table}

In the published scores, the difference of means lies at the level of vanilla on the ten tasks whose positions coincide
(median \cgMeanAlLo--\cgMeanAlHi{} per size) and at \cgMeanOtLo--\cgMeanOtHi{} on the other 19 (\cref{fig:benchmarks}a). Since the
ten are exactly the negative-polarity and filler--gap tasks, this difference could also come from the kind of task. We
therefore re-ran the authors' code on the five Pythia sizes up to 410m, once as released and once with the two positions in
order, on the 19 tasks and on two of the ten as controls. As released, the code reproduces the published scores at 70m, 160m
and 410m to the second decimal on all 19 tasks (LDA on \cgrLdaRepro); at 14m and 31m, the weights at the published address
were replaced in February 2026, so the published values cannot serve as a reference there. On the controls, the
deterministic methods give identical rows in both conditions. With the positions in order, the difference of means rises
from a median of \cgrMeanRel{} of vanilla to \cgrMeanFixed{} and the probes reach \cgrProbeFixed{}, while PCA and $k$-means stay
at \cgrPcaKmFixed{}. Retrained DAS lies at a median of \cgrDasFixed{} times vanilla per size, above every other method on all 19
tasks at each of the five sizes. The order of the positions thus explains the lower published scores of the difference of
means and the probes on these tasks, but on most of them not those of PCA and $k$-means.

\subsection{Re-running the Sentiment Direction}
\label{app:sentimentfull}

\begin{table}[tbp]
  \caption{\textbf{The sentiment direction re-run on seven models.} We use the authors' code with three initialisations per model. The score is the effect of patching the direction at every position of a layer, a complete cut where whole-state patching gives exactly 100\%. It is given in \% on the authors' logit-difference metric (the share of the gap from the flipped run to the clean one), at the layer that the published rule selects (the largest over layers), averaged over the initialisations. Layers are numbered as in the rest of the paper, the output of block $\ell$ as layer $\ell$. Interval: a band that is simultaneous over the layers and so counts the choice of layer (blocks of two neighbouring prompts, 10{,}000 resamples), compared with 100\%. Each initialisation on its own gives the same result. Bonferroni-t: a conservative $t$ interval at the selected layer, corrected over the layers as if they were independent (neighbouring layers are strongly correlated). $^\dagger$The band was chosen after the first initialisation of this model had been seen, but before its other two initialisations were seen. Whole-state patching, partial cut: the effect of whole-state patching at the adjective and verb positions, as its largest value over the layers after the first block. At the embeddings, where only these two positions differ, this site is a complete cut, and the effect is exactly 100\%. Published, Pythia-1.4b: 109.8\%.}
  \label{tab:sentiment}
  \centering
  \footnotesize
  \setlength{\tabcolsep}{5pt}
  \begin{tabular}{@{}lrrrllr@{}}
    \toprule
    model & layers & selected & \begin{tabular}[b]{@{}c@{}}score\\(\%)\end{tabular} & \begin{tabular}[b]{@{}c@{}}band,\\choice counted\end{tabular} & Bonferroni-t & \begin{tabular}[b]{@{}c@{}}whole-state\\patching,\\partial cut (\%)\end{tabular} \\
    \midrule
    GPT-2 small & 12 & 7 & 83.2 & [62.1, 102.6] includes & [56.7, 109.8] includes & 98.3 \\
    Pythia-160m$^\dagger$ & 12 & 4 & 109.9 & [80.6, 132.8] includes & [74.9, 144.9] includes & 99.7 \\
    Pythia-410m$^\dagger$ & 24 & 14 & 198.8 & [91.9, 271.0] includes & [71.6, 326.0] includes & 121.7 \\
    Pythia-1b & 16 & 7 & 84.1 & [56.3, 108.5] includes & [49.7, 118.4] includes & 99.3 \\
    Pythia-1.4b & 24 & 10 & 114.7 & [85.0, 137.4] includes & [78.5, 150.9] includes & 99.5 \\
    SmolLM2-360M & 32 & 20 & 124.9 & [81.8, 160.3] includes & [70.5, 179.3] includes & 100.0 \\
    Qwen2.5-0.5B & 24 & 13 & 79.0 & [69.5, 87.3] below & [67.7, 90.4] below & 99.8 \\
    \bottomrule
  \end{tabular}
\end{table}

The re-run of \cref{app:sentiment} covers \srModelsWord{} models, each with three initialisations (\cref{tab:sentiment}). Three
details of the code of \citet{hollinsworth2024} matter for what the published number measures. First,
the authors' results script builds Table~1a from the all-position column, taking the maximum over layers. Their text
describes the adjective and verb positions, so only the script shows that the published number comes from a complete cut.
Second, the best-epoch reload restores the saved state dictionaries of the rotation module, which alias its live weights, so
it returns the last epoch. At the selected layer, the directions under the two rules have cosine \srBLCosLo{} or more and
values within \srBLDiffSel{} points. At other layers, the values differ by up to \srBLDiffMax{} points. Third, the code
trains the direction by patching it at the adjective alone, a partial cut, so its objective never acts on the complete cut
that the published number reads. We copy the behaviour
of the code, including the orthogonal parametrisation of the rotation, gradient clipping and the seed. Because the table
keeps the best of the layers, the interval at the selected layer accounts for that choice. It is a max-type studentised band,
simultaneous over the layers, on blocks of two neighbouring prompts (\srBlocks{} blocks per model). Each bound is at 2.5\%,
as for every interval in this paper. Under that band, our run at Pythia-1.4b is consistent with the published value. The band
does not lie above 100\% in any model, for any initialisation. The conservative Bonferroni interval does not lie above 100\%
in any model either (\cref{tab:sentiment}). The direction of Pythia-410m has the highest value, \srPythiaY\%, with a band
from \srPythiaLo{} to \srPythiaHi\%. This result depends on the small-sample correction, which widens both the band and the
Bonferroni interval when there are so few blocks. With $u$ the number of blocks, the band multiplies the standard
deviation of the resampling by $\sqrt{u/(u-1)}$ and its critical value by $t_{0.975,\,u-1}/z_{0.975}$. The Bonferroni interval
uses the same scaled standard deviation with Student's $t$ on $u-1$ degrees of freedom. Without the correction, the band runs
from \srPythiaPlainLo{} to \srPythiaPlainHi\%, above 100\% for every initialisation, and the Bonferroni interval lies above
100\% too. The band of every other model gives the same result without the correction. The directions from the
three initialisations lie within half a point of each other (\srPythiaYLo{} to \srPythiaYHi\%), so the two that we had not
seen when we chose the band cannot test that choice.

\FloatBarrier
\section{The OwnEffect Library}
\label{app:owneffectlib}

This appendix describes the library of \cref{app:owneffect} in full. Besides the quantities of the three steps of that
section, the library computes those of the theory on the given pairs. For the range of \cref{thm:range} at a cell or at the readout, it reports the trace, $\lambda_1$ (with
its interval at the readout), $\lambda_D$ and the position of a given direction in the range. The upper end $\lambda_1$, the
predicted maximum, is computed before any training and bounds the effect that a rank-one direction can reach to first order.
Inside the network, training exceeds it at some cells (\cref{sec:ranked}). The value that \cref{prop:natural} computes from a single
angle predicts $\lambda_1$ on a two-label set, and the trace test of \cref{prop:trapezoid} checks the first-order approximation on
a mirrored set. The library also checks \cref{prop:cut} on the given model. It works on Hugging Face models with the GPT-2, GPT-NeoX (Pythia), Llama, Qwen2 and Gemma-2 architectures (SmolLM2 uses the Llama architecture), loaded with eager
attention. It takes pairs in CausalGym's format, that
is, a base and a source split into regions, with two single-token labels. It follows the conventions of the paper. Layers and
cells are as in \cref{app:sites}, the best context is chosen on the training pairs or on the given pairs, effects are in
units of the mean $T$, and intervals resample a pair with its mirror 2000 times. The patches need only forward passes. The
range and the angle need one gradient per pair, at the midpoint between the base and source states, and the trace test
needs a second gradient, at the base point. The calls of the library are documented with the code (\cref{app:compute}).

On tiny random models of each family, the test of the library checks three statements exactly. Patching every position of a
layer gives $T$, a patch along a pair's own difference equals whole-state patching, and \cref{thm:range}(iv) holds for one
pair. On GPT-2, from our pairs and saved directions, the library reproduces the values of our measurement. At the readout of
\oeTasks{} CausalGym tasks, $T$ and the exact effect of every saved direction on each test pair (\oeEffects{} values) agree
within \oeReadoutMax{} log odds, and $\lambda_1$ and the trace agree within \oeSpecMax. On the \oeOneQ{} two-label tasks, the
cosine of the angle and the shared fraction of the readout gradients (\cref{app:landing}) agree within \oeAngleMax. Inside
the network, we compare at the \oeCells{} live cells (\cref{sec:above}) below the readout in
\oeMapsWord{} maps. At every cell, the context that the library chooses for $G^*$ on the training pairs is the same as ours.
The values of $G$, of $G^*$ (in that context and in the context chosen on the test pairs) and of five saved directions, each
measured with attention free and with attention frozen, agree per test pair within \oeInsideMax{} log odds. The value of
$w^\top\Delta$ agrees per pair within \oeInsideWD{}, and $\lambda_1$ agrees within \oeInsideLam. The library recomputes the
whole-state patches at every cell and the base run with every patch, whereas the scripts behind our maps share these passes.
It therefore needs about twice the forward passes counted below.

\paragraph{Cost.} Consider one cell of a template with $R$ regions besides the start-of-text region, which is neither mapped
nor used as a context. At this cell, \cref{alg:route} runs at most five forward passes with a patch per test pair. Three of
them are for $G^*$ (the cell alone, and its context alone and together with the cell), one is along the direction, and one is
along the direction with attention frozen. The $2R-1$ patches that choose the context of $G^*$ (the cell alone, each other
region alone, and each other region together with the cell) run on each training pair. The total effect $T$ and the base run
take forward passes without a patch, which every cell shares. Over a map, the patches of single regions and of pairs of
regions are shared by every
cell of their layer. This amounts to $R(R+1)/2$ whole-state patches per layer and training pair, at most as many on the test
pairs, and $2R$ patches along directions per layer and test pair. No gradient is computed, and the direction may come from
any method.

\FloatBarrier

\vskip 0.2in
\bibliography{refs}

\end{document}